\documentclass{article}
\usepackage{vade_preprint,times}
\usepackage{xcolor}
\usepackage{verbatim}

\usepackage{amsmath,amsfonts,bm}

\def\eqref#1{equation~\ref{#1}}

\def\1{\bm{1}}

\def\rh{{\textnormal{h}}}

\DeclareMathAlphabet{\mathsfit}{\encodingdefault}{\sfdefault}{m}{sl}
\SetMathAlphabet{\mathsfit}{bold}{\encodingdefault}{\sfdefault}{bx}{n}

\usepackage{hyperref}
\usepackage{url}
\usepackage{booktabs}
\usepackage{array}
\usepackage{multirow}
\usepackage{enumitem}
\usepackage{xcolor}
\usepackage{tikz}
\usetikzlibrary{arrows.meta, positioning, shapes.geometric, fit, backgrounds, calc, decorations.pathreplacing, shadows}
\usepackage{fontawesome5}

\usepackage{etoc}
\usepackage{titletoc}

\newcommand\DoToC{%
  \startcontents
  {\setlength{\parskip}{2pt}%
  \printcontents{}{1}{\noindent \textbf{\Large{Table of Contents in Appendix}}\vskip3pt\vskip5pt}}
  \vskip3pt\vskip5pt
}

\usepackage{pifont}

\newcommand{\vadeicon}[2]{\csname vadeicon@#1\endcsname{#2}}
\expandafter\newcommand\csname vadeicon@activity-deepseek\endcsname[1]{\includegraphics[height=#1]{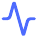}}
\expandafter\newcommand\csname vadeicon@bot-deepseek\endcsname[1]{\includegraphics[height=#1]{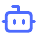}}
\expandafter\newcommand\csname vadeicon@calculator-gpt\endcsname[1]{\includegraphics[height=#1]{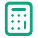}}
\expandafter\newcommand\csname vadeicon@calendar-gray\endcsname[1]{\includegraphics[height=#1]{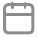}}
\expandafter\newcommand\csname vadeicon@circle-check-gpt\endcsname[1]{\includegraphics[height=#1]{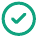}}
\expandafter\newcommand\csname vadeicon@clipboard-check-qwen\endcsname[1]{\includegraphics[height=#1]{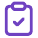}}
\expandafter\newcommand\csname vadeicon@clipboard-list-gray\endcsname[1]{\includegraphics[height=#1]{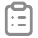}}
\expandafter\newcommand\csname vadeicon@compass-claude\endcsname[1]{\includegraphics[height=#1]{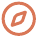}}
\expandafter\newcommand\csname vadeicon@eye-ink\endcsname[1]{\includegraphics[height=#1]{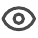}}
\expandafter\newcommand\csname vadeicon@eye-off-red\endcsname[1]{\includegraphics[height=#1]{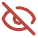}}
\expandafter\newcommand\csname vadeicon@file-check-gpt\endcsname[1]{\includegraphics[height=#1]{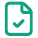}}
\expandafter\newcommand\csname vadeicon@file-text-gray\endcsname[1]{\includegraphics[height=#1]{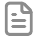}}
\expandafter\newcommand\csname vadeicon@gauge-claude\endcsname[1]{\includegraphics[height=#1]{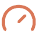}}
\expandafter\newcommand\csname vadeicon@gavel-deepseek\endcsname[1]{\includegraphics[height=#1]{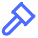}}
\expandafter\newcommand\csname vadeicon@globe-ink\endcsname[1]{\includegraphics[height=#1]{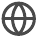}}
\expandafter\newcommand\csname vadeicon@inbox-claude\endcsname[1]{\includegraphics[height=#1]{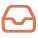}}
\expandafter\newcommand\csname vadeicon@layers-gpt\endcsname[1]{\includegraphics[height=#1]{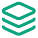}}
\expandafter\newcommand\csname vadeicon@list-checks-gray\endcsname[1]{\includegraphics[height=#1]{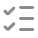}}
\expandafter\newcommand\csname vadeicon@lock-gray\endcsname[1]{\includegraphics[height=#1]{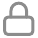}}
\expandafter\newcommand\csname vadeicon@package-check-qwen\endcsname[1]{\includegraphics[height=#1]{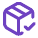}}
\expandafter\newcommand\csname vadeicon@pen-line-qwen\endcsname[1]{\includegraphics[height=#1]{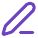}}
\expandafter\newcommand\csname vadeicon@pencil-ink\endcsname[1]{\includegraphics[height=#1]{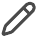}}
\expandafter\newcommand\csname vadeicon@quote-claude\endcsname[1]{\includegraphics[height=#1]{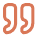}}
\expandafter\newcommand\csname vadeicon@quote-gray\endcsname[1]{\includegraphics[height=#1]{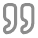}}
\expandafter\newcommand\csname vadeicon@rotate-ccw-red\endcsname[1]{\includegraphics[height=#1]{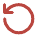}}
\expandafter\newcommand\csname vadeicon@route-gray\endcsname[1]{\includegraphics[height=#1]{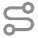}}
\expandafter\newcommand\csname vadeicon@scale-gpt\endcsname[1]{\includegraphics[height=#1]{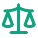}}
\expandafter\newcommand\csname vadeicon@search-claude\endcsname[1]{\includegraphics[height=#1]{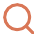}}
\expandafter\newcommand\csname vadeicon@shield-check-qwen\endcsname[1]{\includegraphics[height=#1]{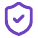}}
\expandafter\newcommand\csname vadeicon@snowflake-qwen\endcsname[1]{\includegraphics[height=#1]{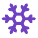}}
\expandafter\newcommand\csname vadeicon@table-claude\endcsname[1]{\includegraphics[height=#1]{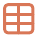}}
\expandafter\newcommand\csname vadeicon@target-gray\endcsname[1]{\includegraphics[height=#1]{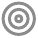}}
\expandafter\newcommand\csname vadeicon@workflow-gray\endcsname[1]{\includegraphics[height=#1]{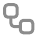}}

\definecolor{discblue}{HTML}{4F7CAC}
\definecolor{discbluelt}{HTML}{E6EDF6}
\definecolor{buildgreen}{HTML}{4E8F6E}
\definecolor{buildgreenlt}{HTML}{E4F0E9}
\definecolor{compileamber}{HTML}{BF8F3A}
\definecolor{compileamberlt}{HTML}{F7EFDD}
\definecolor{evalpurple}{HTML}{7D5BA6}
\definecolor{evalpurplelt}{HTML}{ECE6F3}
\definecolor{archgray}{HTML}{8A8F98}
\definecolor{feedred}{HTML}{B9463F}
\definecolor{inkblack}{HTML}{2B2D33}
\definecolor{gridyes}{HTML}{4E8F6E}
\definecolor{gridpart}{HTML}{BF8F3A}
\definecolor{gridno}{HTML}{D9DCE1}
\definecolor{gridtool}{HTML}{4F7CAC}

\newcommand{\pipename}{{VADE}}
\newcommand{\dsname}{{VADE-Finance}}
\newcommand{\lawname}{{VADE-Law}}
\newcommand{\energyname}{{VADE-Energy-Preview}}   %
\newcommand{\medname}{{VADE-Medicine-Preview}}   %
\newcommand{\energyfull}{{VADE-Energy}}     %
\newcommand{\medfull}{{VADE-Medicine}}      %
\newcommand{\nenergytasks}{85}      %
\newcommand{\nmedtasks}{85}         %
\newcommand{\dstest}{{VADE-Finance-Test}}   %

\newcommand{\ntasks}{1000}                  %
\newcommand{\nindustries}{25}                        %
\newcommand{\nreviewed}{186}               %

\newcommand{\ntraintasks}{400}   %
\newcommand{\ntesttasks}{75}   %

\newcommand{\nfiles}{96{,}524}        %
\newcommand{\nlawfiles}{53{,}498}     %
\newcommand{\nlawareas}{12}           %
\newcommand{\nlawtasks}{372}          %

\newcommand{\figfont}{\fontfamily{qpl}\selectfont}

\makeatletter
\def\paragraph{\@startsection{paragraph}{4}{\z@}{0.8ex plus 0.3ex minus .2ex}{-1em}{\normalsize\bf}}
\makeatother
\setlist{itemsep=1pt,topsep=3pt}

\hypersetup{pdftitle={VADE: A Verifiable Agent Data Engine from Professional Sources},
  pdfauthor={Xiangchao Chen, Zhiheng Lyu, Kai Zhang, Xueguang Ma, Ping Nie, Wenhu Chen}}

\title{VADE: A Verifiable Agent Data Engine \\ from Professional Sources}
\runningtitle{VADE: A Verifiable Agent Data Engine from Professional Sources}

\author{
{\bfseries
Xiangchao Chen\textsuperscript{1}\enspace
Zhiheng Lyu\textsuperscript{2}\enspace
Kai Zhang\textsuperscript{3}\enspace
Xueguang Ma\textsuperscript{1}\enspace
Ping Nie\textsuperscript{1}\enspace
Wenhu Chen\textsuperscript{1}}\\[3pt]
\textsuperscript{1}University of Waterloo\enspace
\textsuperscript{2}University of Illinois Urbana-Champaign\enspace
\textsuperscript{3}The Ohio State University\\[2pt]
\texttt{chasechen98@gmail.com}\\[2pt]
\faGlobe~\url{https://tiger-ai-lab.github.io/VADE/}
}

\begin{document}
\maketitle

\begin{abstract}
Language agents have demonstrated their practical value, particularly in software engineering, where abundant public code and executable tests support training and evaluation at scale.
Extending this success to other professional domains, such as law and finance, is not straightforward: many agent tasks involve open-ended analysis, making high-quality outcomes difficult to verify and, in turn, difficult to use as scalable supervision.
As a result, high-quality tasks in these domains are still largely expert-authored, making them expensive to scale and limiting their use for training.
To address this bottleneck, we introduce \pipename{}, a data engine that automatically constructs verifiable analysis tasks from publicly available sources used in professional practice.
\pipename{} exploits a simple asymmetry: professional analyses are difficult to solve because the relevant evidence must first be found and interpreted, but much easier to verify once that evidence is fixed.
During task synthesis, the engine first constructs a reference analysis from source evidence and asks an independent agent to reproduce its conclusions from the same evidence without seeing the reference analysis.
Only after the analysis is verified is it compiled into an open-ended task with an item-level scoring contract, allowing the same task to support both evaluation and training.
We apply \pipename{} across four professional domains; in an independent expert review of \nreviewed{} finance tasks, 93.5\% meet a high bar for source fidelity, answer correctness, and evaluation fairness.
In finance, these tasks expose substantial remaining model errors, particularly unsupported facts and inferences, while training on verified solver trajectories improves held-out performance and yields gains that transfer to public finance benchmarks such as Finance Agent Benchmark~v2.
Together, these results suggest a practical path toward scalable, verifiable supervision for open-ended professional tasks.
\end{abstract}

\section{Introduction}
\label{sec:intro}

Language agents, which follow natural-language instructions to act in real or simulated environments~\citep{su-etal-2024-language,sumers2024cognitive}, have become practically useful for a growing range of complex tasks.
Much of this progress comes from agent data, the tasks and trajectories from which agents learn to act~\citep{agenttuning2023,apigenmt2025,Zhang2026EarlyExperience}.
Tasks with verifiable outcomes are especially valuable: they evaluate agents directly and, once solution trajectories are filtered by their outcomes, also train them.
Coding agents show this most clearly~\citep{sweagent2024,openhands2025}: public repositories supply realistic tasks at scale, and executable tests make every outcome cheap to verify, so the same tasks serve as benchmarks~\citep{swebench2024} and as training environments~\citep{swegym2024,swesmith2025,r2egym2025}.

However, other professional domains, such as
law~\citep{legalbench2023} and finance~\citep{financebench2023}, have not
yet seen the same progress, largely because their tasks are hard to verify.
Many agent tasks in these domains involve open-ended analysis: the agent must find evidence scattered across the sources used in practice, interpret it under the right basis, period, and scope, and deliver a conclusion that holds under stated conditions.
The sources provide evidence rather than answers, and no test can be executed against the conclusion.
In a filing, for example, what drives a change in profit depends on which basis is used and how each adjustment is treated.
Without an executable check, reference answers need expert review, and solver trajectories cannot be filtered at scale.
As a result, tasks in these domains are still written mostly by experts.
GDPval~\citep{gdpval2025}, APEX~\citep{apexagents2026,apex2026}, Finance Agent
Benchmark~\citep{fabv1_2025,fabv2_2026},
BigFinanceBench~\citep{bigfinancebench2026}, and
the Legal Agent Benchmark (LAB)~\citep{harvey_lab2026} evaluate agents on realistic work tasks against expert-written criteria.

Automated task synthesis scales, but existing methods obtain verifiability
by restricting the task.
TaskCraft~\citep{taskcraft2026} uses tool-executable facts as reference
answers, FinForge~\citep{finforge2026} reduces tasks to multiple choice,
and AGORA~\citep{agora2026} limits the answer to a single number.
QUEST~\citep{quest2026} keeps research tasks open-ended, but its reference answers and rubrics are produced by the same process that writes the tasks, with no independent check.
What is missing is a way to verify the reference answer of an
open-ended analysis task without relying on experts or narrowing the task.

We introduce \pipename{}, a data engine that automatically constructs
verifiable analysis tasks from public sources used in professional
practice (Figure~\ref{fig:pipeline}).
\pipename{} is built on an asymmetry between solving and verifying a professional analysis: a solver must first find and interpret the relevant evidence, whereas a verifier given that evidence need only derive the conclusions from it.
The engine first finds candidate analyses in the sources through ten discovery routes (Table~\ref{tab:routes}) and builds a reference analysis for each from quoted evidence; code checks every quotation against the sources and recomputes every arithmetic step.
The reference analysis is then verified by independent reproduction: a second agent derives its conclusions from the same evidence without seeing them, and a third agent compares the two analyses and attributes every difference either to
the task, which is repaired, or to the reproduction, which is rerun.
Only after the reference analysis is verified is the question written, together with an item-level \emph{scoring contract} that specifies what a correct answer must contain and what fails it outright.
Because every answer can be judged item by item against the scoring
contract, the same task supports both evaluation and training.

We apply \pipename{} across four professional domains: finance, law,
medicine, and energy.
In finance, it delivers \ntasks{} tasks across \nindustries{} industry
groups, and in law, \nlawtasks{} tasks across \nlawareas{} practice
areas (\S\ref{sec:research-set}).
In an independent review of \nreviewed{} finance tasks, experts judged 93.5\% to meet a high bar for source fidelity, answer correctness, and evaluation fairness (\S\ref{sec:human-review}).
In finance, the tasks remain far from solved: six models reach pass rates of 19.4\% to 66.4\%
(\S\ref{sec:model-perf}), and unsupported facts and inferences, the most common error, appear in 85.9\% of failed answers
(\S\ref{sec:failure}).
The same tasks also serve as supervision.
Training Qwen3.5-27B on verified solver trajectories from
\ntraintasks{} tasks raises its pass rate from 6.7\% to 21.3\% on
\ntesttasks{} held-out tasks, and the gains transfer to four public
finance benchmarks; on Finance Agent Benchmark~v2, it improves from 29.6\% to 51.9\% (\S\ref{sec:results-learn}).

Taken together, these results show that open-ended professional tasks
need not trade verifiability for scale.
Verifying each reference analysis by independent reproduction removes
the need for expert authoring without narrowing the task, and the
resulting tasks are hard enough to expose where current models fail and
reliable enough to train on.
We see \pipename{} as a step toward agent data for professional work
that grows with the available sources rather than with expert time.

\section{The \pipename{} Engine}
\label{sec:method}
\providecommand{\figystretch}{1.10}
\begin{figure}[t]
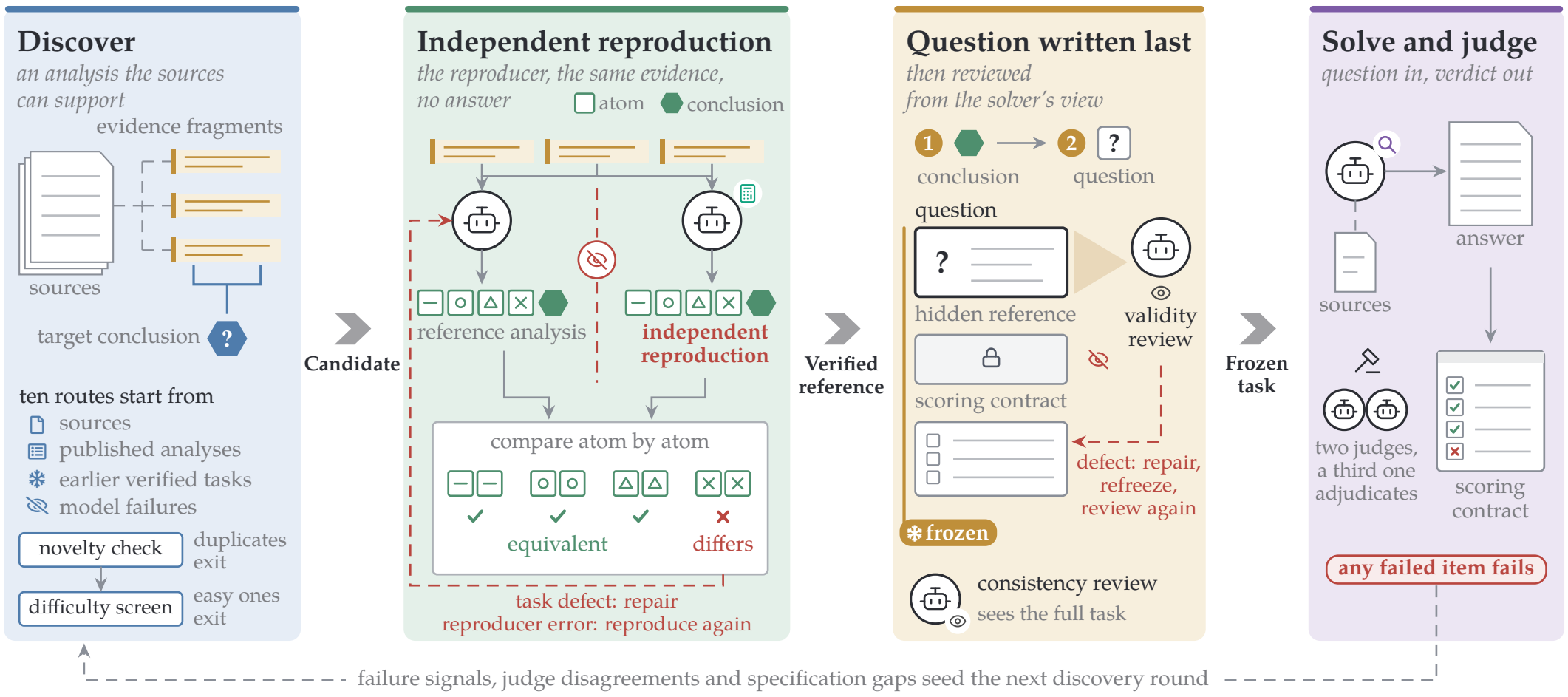

\centering
\begin{tikzpicture}[x=1cm,y=\figystretch cm,
  every node/.style={inner sep=0pt, outer sep=0pt, font=\figfont},
  ttl/.style={anchor=base west, font=\figfont\fontsize{7.5}{8}\selectfont\bfseries, text=inkblack},
  sub/.style={anchor=base west, font=\figfont\fontsize{5.8}{6.4}\selectfont\itshape, text=inkblack!62, align=left},
  lb/.style={font=\figfont\fontsize{5.5}{6.1}\selectfont, text=inkblack, align=center},
  lg/.style={font=\figfont\fontsize{5.5}{6.1}\selectfont, text=inkblack!62, align=center},
  lgs/.style={font=\figfont\fontsize{5.2}{5.8}\selectfont, text=inkblack!62, align=center},
  ic/.style={line width=0.5pt, line cap=round, line join=round},
  boticon/.pic={
    \draw[ic, rounded corners=1.1pt] (-0.115cm,-0.105cm) rectangle (0.115cm,0.055cm);
    \draw[ic] (0cm,0.055cm) -- (0cm,0.115cm);
    \draw[ic] (-0.045cm,0.115cm) -- (0.045cm,0.115cm);
    \draw[ic] (-0.045cm,-0.005cm) -- (-0.045cm,-0.045cm);
    \draw[ic] (0.045cm,-0.005cm) -- (0.045cm,-0.045cm);
    \draw[ic] (-0.115cm,-0.025cm) -- (-0.155cm,-0.025cm);
    \draw[ic] (0.115cm,-0.025cm) -- (0.155cm,-0.025cm);},
  usericon/.pic={
    \draw[ic] (0cm,0.045cm) circle[radius=0.05cm];
    \draw[ic] (-0.095cm,-0.105cm) .. controls (-0.095cm,-0.02cm) and (0.095cm,-0.02cm) .. (0.095cm,-0.105cm);},
  lockicon/.pic={
    \draw[ic, rounded corners=0.6pt] (-0.07cm,-0.075cm) rectangle (0.07cm,0.015cm);
    \draw[ic] (-0.04cm,0.015cm) -- (-0.04cm,0.045cm) arc[start angle=180,end angle=0,radius=0.04cm] -- (0.04cm,0.015cm);},
  searchicon/.pic={
    \draw[ic] (-0.012cm,0.012cm) circle[radius=0.045cm];
    \draw[ic] (0.022cm,-0.022cm) -- (0.065cm,-0.065cm);},
  gavelicon/.pic={
    \draw[ic] (0.012cm,0.104cm) -- (-0.034cm,0.058cm) -- (0.058cm,-0.034cm) -- (0.104cm,0.012cm) -- cycle;
    \draw[ic] (0.012cm,0.012cm) -- (-0.095cm,-0.095cm);
    \draw[ic] (0.0cm,-0.105cm) -- (0.115cm,-0.105cm);},
  g1/.pic={\draw[ic, buildgreen] (-0.055cm,0cm) -- (0.055cm,0cm);},
  g2/.pic={\draw[ic, buildgreen] (0cm,0cm) circle[radius=0.04cm];},
  g3/.pic={\draw[ic, buildgreen] (0cm,0.05cm) -- (-0.055cm,-0.04cm) -- (0.055cm,-0.04cm) -- cycle;},
  g4/.pic={\draw[ic, buildgreen] (-0.045cm,-0.045cm) -- (0.045cm,0.045cm) (-0.045cm,0.045cm) -- (0.045cm,-0.045cm);},
  okmark/.pic={\draw[line width=0.8pt, line cap=round, line join=round] (-0.055cm,0cm) -- (-0.018cm,-0.04cm) -- (0.06cm,0.05cm);},
  nomark/.pic={\draw[line width=0.8pt, line cap=round] (-0.042cm,-0.042cm) -- (0.042cm,0.042cm) (-0.042cm,0.042cm) -- (0.042cm,-0.042cm);},
  oksmall/.pic={\draw[line width=0.6pt, line cap=round, line join=round] (-0.036cm,-0.002cm) -- (-0.012cm,-0.028cm) -- (0.036cm,0.03cm);},
  nosmall/.pic={\draw[line width=0.6pt, line cap=round] (-0.028cm,-0.028cm) -- (0.028cm,0.028cm) (-0.028cm,0.028cm) -- (0.028cm,-0.028cm);},
  lbs/.style={font=\figfont\fontsize{5.2}{5.8}\selectfont, text=inkblack, align=center},
  red/.style={font=\figfont\fontsize{5.2}{5.8}\selectfont, text=feedred, align=center},
  hand/.style={font=\figfont\fontsize{5.2}{5.8}\selectfont\bfseries, text=inkblack, align=center},
  flow/.style={-{Stealth[length=1.3mm,width=1.1mm]}, line width=0.5pt, archgray},
  rflow/.style={-{Stealth[length=1.3mm,width=1.1mm]}, line width=0.5pt, feedred, densely dashed},
  rline/.style={line width=0.5pt, feedred, densely dashed},
  bot/.style={circle, draw=inkblack, line width=0.6pt, fill=white, minimum size=0.52cm,
              font=\fontsize{7}{7}\selectfont, text=inkblack},
  atom/.style={rectangle, rounded corners=0.6pt, draw=buildgreen, line width=0.5pt, fill=white,
               minimum size=0.22cm},
  hex/.style={regular polygon, regular polygon sides=6, fill=#1, minimum size=0.27cm},
  card/.style={rectangle, rounded corners=1.2pt, draw=archgray!70, line width=0.5pt, fill=white},
  qbar/.style={rectangle, fill=compileamberlt, minimum height=0.2cm},
]
\def\PT{5.66}\def\PB{0.52}
\def\Aa{0}\def\Ab{2.66}
\def\Ba{3.58}\def\Bb{7.04}
\def\Ca{7.96}\def\Cb{10.76}
\def\Da{11.68}\def\Db{13.97}
\foreach \a/\b/\c/\d in {\Aa/\Ab/discblue/discbluelt, \Ba/\Bb/buildgreen/buildgreenlt, \Ca/\Cb/compileamber/compileamberlt, \Da/\Db/evalpurple/evalpurplelt}{
  \fill[\d, rounded corners=3pt] (\a,\PB) rectangle (\b,\PT);
  \draw[\c, line width=1.6pt, line cap=round] (\a+0.04,\PT) -- (\b-0.04,\PT);
}
\node[ttl] at (\Aa+0.12,\PT-0.34) {Discover};
\node[sub] at (\Aa+0.12,\PT-0.58) {an analysis the sources};
\node[sub] at (\Aa+0.12,\PT-0.80) {can support};
\node[ttl] at (\Ba+0.12,\PT-0.34) {Independent reproduction};
\node[sub] at (\Ba+0.12,\PT-0.58) {the reproducer, the same evidence,};
\node[sub] at (\Ba+0.12,\PT-0.80) {no answer};
\node[atom, minimum size=0.17cm] at (\Ba+1.62,\PT-0.765) {};
\node[lgs, anchor=west] at (\Ba+1.75,\PT-0.765) {atom};
\node[hex=buildgreen, minimum size=0.21cm] at (\Ba+2.40,\PT-0.765) {};
\node[lgs, anchor=west] at (\Ba+2.54,\PT-0.765) {conclusion};
\node[ttl] at (\Ca+0.12,\PT-0.34) {Question written last};
\node[sub] at (\Ca+0.12,\PT-0.58) {then reviewed};
\node[sub] at (\Ca+0.12,\PT-0.80) {from the solver's view};
\node[ttl] at (\Da+0.12,\PT-0.34) {Solve and judge};
\node[sub] at (\Da+0.12,\PT-0.58) {question in, verdict out};

\foreach \x/\t in {3.12/{Candidate}, 7.50/{Verified\\reference}, 11.22/{Frozen\\task}}{
  \fill[inkblack!45] (\x-0.16,3.19) -- (\x+0.02,3.19) -- (\x+0.17,3.06) -- (\x+0.02,2.93) -- (\x-0.16,2.93) -- (\x-0.01,3.06) -- cycle;
  \node[hand, anchor=north] at (\x,2.84) {\t};
}

\foreach \o in {0.10,0.05,0}{
  \draw[archgray, line width=0.4pt, fill=white] (\Aa+0.24-\o,3.60-\o) -- (\Aa+0.24-\o,4.50-\o) -- (\Aa+0.86-\o,4.50-\o) -- (\Aa+0.98-\o,4.38-\o) -- (\Aa+0.98-\o,3.60-\o) -- cycle;
}
\foreach \y in {4.30,4.14,3.98,3.82}{\draw[archgray!60, line width=0.5pt] (\Aa+0.34,\y) -- (\Aa+0.86,\y);}
\node[lg, anchor=north] at (\Aa+0.55,3.42) {sources};
\node[lg, anchor=south east] at (\Aa+2.50,4.60) {evidence fragments};
\foreach \y/\n in {4.42/q1,4.06/q2,3.70/q3}{
  \node[qbar, minimum width=0.98cm, anchor=west] (\n) at (\Aa+1.50,\y) {};
  \fill[compileamber] (\Aa+1.50,\y-0.10) rectangle (\Aa+1.54,\y+0.10);
  \draw[compileamber, line width=0.5pt] (\Aa+1.62,\y+0.035) -- (\Aa+2.38,\y+0.035);
  \draw[compileamber, line width=0.5pt] (\Aa+1.62,\y-0.04) -- (\Aa+2.14,\y-0.04);
  \draw[archgray, line width=0.4pt, densely dashed] (\Aa+1.24,\y) -- (\Aa+1.50,\y);
}
\draw[archgray, line width=0.4pt, densely dashed] (\Aa+0.98,4.06) -- (\Aa+1.24,4.06) (\Aa+1.24,4.42) -- (\Aa+1.24,3.70);
\draw[discblue, line width=0.6pt] (\Aa+1.70,3.58) -- (\Aa+1.70,3.42) -- (\Aa+2.30,3.42) -- (\Aa+2.30,3.58);
\draw[discblue, line width=0.6pt] (\Aa+2.00,3.42) -- (\Aa+2.00,3.18);
\node[hex=discblue, minimum size=0.36cm, font=\fontsize{6}{6}\selectfont\bfseries, text=white] (tc) at (\Aa+2.00,2.98) {?};
\node[lg, anchor=east] at (\Aa+1.76,2.98) {target conclusion};
\node[lb, anchor=west] at (\Aa+0.14,2.52) {ten routes start from};
\foreach \y/\i/\t in {2.28/{\faFile[regular]}/sources, 2.06/{\faIcon[regular]{list-alt}}/published analyses, 1.84/{\faSnowflake[regular]}/earlier verified tasks, 1.62/{\faEyeSlash[regular]}/model failures}{
  \node[font=\fontsize{4.6}{4.6}\selectfont, text=discblue, anchor=center] at (\Aa+0.30,\y) {\i};
  \node[lg, anchor=west] at (\Aa+0.50,\y) {\t};
}
\node[card, draw=discblue, minimum width=1.46cm, minimum height=0.30cm] (nov) at (\Aa+0.87,1.26) {};
\node[lbs] at (\Aa+0.87,1.26) {novelty check};
\node[lbs, anchor=west, align=left, text=inkblack!62] at (\Aa+1.70,1.26) {duplicates\\exit};
\node[card, draw=discblue, minimum width=1.46cm, minimum height=0.30cm] (dif) at (\Aa+0.87,0.78) {};
\node[lbs] at (\Aa+0.87,0.78) {difficulty screen};
\node[lbs, anchor=west, align=left, text=inkblack!62] at (\Aa+1.70,0.78) {easy ones\\exit};
\draw[flow] (nov.south) -- (dif.north);

\foreach \x in {0.24,1.27,2.30}{
  \node[qbar, minimum width=0.92cm, anchor=west] at (\Ba+\x,4.50) {};
  \fill[compileamber] (\Ba+\x,4.40) rectangle (\Ba+\x+0.04,4.60);
  \draw[compileamber, line width=0.5pt] (\Ba+\x+0.12,4.535) -- (\Ba+\x+0.80,4.535);
  \draw[compileamber, line width=0.5pt] (\Ba+\x+0.12,4.46) -- (\Ba+\x+0.58,4.46);
}
\node[bot] (w1) at (\Ba+0.70,3.94) {};
\pic[inkblack] at (\Ba+0.70,3.94) {boticon};
\node[bot] (w2) at (\Ba+2.76,3.94) {};
\pic[inkblack] at (\Ba+2.76,3.94) {boticon};
\draw[archgray, line width=0.5pt] (\Ba+0.70,4.40) -- (\Ba+0.70,4.30) -- (\Ba+2.76,4.30) -- (\Ba+2.76,4.40) (\Ba+1.73,4.40) -- (\Ba+1.73,4.30);
\draw[flow] (\Ba+0.70,4.30) -- (w1.north);
\draw[flow] (\Ba+2.76,4.30) -- (w2.north);
\node[circle, fill=white, minimum size=0.23cm] at (\Ba+3.08,4.16) {\includegraphics[height=5.2pt]{figures/icons/calculator-gpt.pdf}};
\foreach \dx/\g in {-0.54/g1,-0.27/g2,0/g3,0.27/g4}{
  \node[atom] at (\Ba+0.78+\dx,3.27) {}; \pic at (\Ba+0.78+\dx,3.27) {\g};
  \node[atom] at (\Ba+2.64+\dx,3.27) {}; \pic at (\Ba+2.64+\dx,3.27) {\g};
}
\node[hex=buildgreen] at (\Ba+1.34,3.27) {};
\node[hex=buildgreen] at (\Ba+3.20,3.27) {};
\draw[flow] (w1.south) -- (\Ba+0.70,3.42);
\draw[flow] (w2.south) -- (\Ba+2.76,3.42);
\node[lg, anchor=north] at (\Ba+0.88,3.11) {reference analysis};
\node[lb, anchor=north, text=feedred, font=\figfont\fontsize{5.5}{6.1}\selectfont\bfseries] at (\Ba+2.70,3.11) {independent\\reproduction};
\draw[rline] (\Ba+1.73,4.20) -- (\Ba+1.73,3.80);
\draw[rline] (\Ba+1.73,3.44) -- (\Ba+1.73,2.62);
\node[circle, draw=feedred, line width=0.5pt, fill=white, minimum size=0.33cm] at (\Ba+1.73,3.62) {\includegraphics[height=5.5pt]{figures/icons/eye-off-red.pdf}};
\draw[rounded corners=1.2pt, line width=0.5pt, draw=archgray!70, fill=white] (\Ba+0.26,2.30) rectangle (\Ba+3.26,1.06);
\draw[flow] (\Ba+0.90,2.88) |- (\Ba+1.30,2.50) -- (\Ba+1.30,2.32);
\draw[flow] (\Ba+2.70,2.66) |- (\Ba+2.22,2.50) -- (\Ba+2.22,2.32);
\node[lg, anchor=north] at (\Ba+1.76,2.22) {compare atom by atom};
\foreach \x/\g/\m/\c in {0.64/g1/okmark/buildgreen, 1.38/g2/okmark/buildgreen, 2.12/g3/okmark/buildgreen, 2.86/g4/nomark/feedred}{
  \node[atom] at (\Ba+\x-0.13,1.80) {}; \pic at (\Ba+\x-0.13,1.80) {\g};
  \node[atom] at (\Ba+\x+0.13,1.80) {}; \pic at (\Ba+\x+0.13,1.80) {\g};
  \pic[\c] at (\Ba+\x,1.53) {\m};
}
\node[lb, text=buildgreen, anchor=north] at (\Ba+1.38,1.38) {equivalent};
\node[lb, text=feedred, anchor=north] at (\Ba+2.86,1.38) {differs};
\draw[rflow] (\Ba+2.86,1.06) -- (\Ba+2.86,0.97) -- (\Ba+0.06,0.97) -- (\Ba+0.06,3.94) -- (w1.west);
\node[red, anchor=north] at (\Ba+1.73,0.91) {task defect: repair\\reproducer error: reproduce again};

\node[circle, fill=compileamber, minimum size=0.25cm, font=\figfont\fontsize{5.5}{5.5}\selectfont\bfseries, text=white] at (\Ca+0.32,4.57) {1};
\node[hex=buildgreen] (ans) at (\Ca+0.68,4.57) {};
\draw[flow] (\Ca+0.93,4.57) -- (\Ca+1.38,4.57);
\node[circle, fill=compileamber, minimum size=0.25cm, font=\figfont\fontsize{5.5}{5.5}\selectfont\bfseries, text=white] at (\Ca+1.60,4.57) {2};
\node[card, draw=archgray, minimum size=0.29cm, font=\figfont\fontsize{6.5}{6.5}\selectfont\bfseries, text=inkblack] at (\Ca+1.98,4.57) {?};
\node[lg, anchor=north] at (\Ca+0.68,4.37) {conclusion};
\node[lg, anchor=north] at (\Ca+1.98,4.37) {question};
\node[lb, anchor=south west] at (\Ca+0.20,3.92) {question};
\draw[rounded corners=1.2pt, line width=0.7pt, draw=inkblack, fill=white] (\Ca+0.20,3.88) rectangle (\Ca+1.56,3.32);
\node[font=\figfont\fontsize{8}{8}\selectfont\bfseries, text=inkblack] at (\Ca+0.44,3.60) {?};
\draw[archgray!60, line width=0.5pt] (\Ca+0.70,3.71) -- (\Ca+1.44,3.71);
\draw[archgray!60, line width=0.5pt] (\Ca+0.70,3.58) -- (\Ca+1.24,3.58);
\draw[archgray!60, line width=0.5pt] (\Ca+0.70,3.45) -- (\Ca+1.34,3.45);
\fill[compileamber!35] (\Ca+1.62,3.86) -- (\Ca+2.10,3.60) -- (\Ca+1.62,3.34) -- cycle;
\node[bot] (vr) at (\Ca+2.40,3.72) {};
\pic[inkblack] at (\Ca+2.40,3.72) {boticon};
\node at (\Ca+2.40,3.35) {\includegraphics[height=5.5pt]{figures/icons/eye-ink.pdf}};
\node[lb, anchor=north] (vrl) at (\Ca+2.40,3.24) {validity\\review};
\node[lg, anchor=north west] at (\Ca+0.20,3.25) {hidden reference};
\draw[rounded corners=1.2pt, line width=0.5pt, draw=archgray, fill=archgray!10] (\Ca+0.20,3.01) rectangle (\Ca+1.56,2.61);
\pic[archgray!150] at (\Ca+0.88,2.82) {lockicon};
\node at (\Ca+1.84,2.81) {\includegraphics[height=5.5pt]{figures/icons/eye-off-red.pdf}};
\node[lg, anchor=north west] at (\Ca+0.20,2.54) {scoring contract};
\draw[rounded corners=1.2pt, line width=0.5pt, draw=archgray, fill=white] (\Ca+0.20,2.30) rectangle (\Ca+1.56,1.70);
\foreach \y in {2.15,2.00,1.85}{
  \node[draw=archgray, rounded corners=0.4pt, line width=0.4pt, minimum size=0.11cm] at (\Ca+0.36,\y) {};
  \draw[archgray!60, line width=0.5pt] (\Ca+0.54,\y) -- (\Ca+1.44,\y);
}
\draw[rflow] (\Ca+2.40,2.76) -- (\Ca+2.40,2.14) -- (\Ca+1.60,2.14);
\node[red, anchor=north] at (\Ca+2.20,2.04) {defect: repair,\\refreeze,\\review again};
\draw[compileamber, line width=0.8pt, line cap=round] (\Ca+0.10,3.88) -- (\Ca+0.10,1.46);
\node[rectangle, rounded corners=3pt, fill=compileamber, minimum height=0.24cm, inner xsep=2pt, anchor=west,
      font=\figfont\fontsize{5.5}{5.5}\selectfont\bfseries, text=white] at (\Ca+0.06,1.40) {{\fontsize{4.2}{4.2}\selectfont\faSnowflake}\,frozen};
\node[bot, minimum size=0.44cm] (cr) at (\Ca+0.38,0.86) {};
\pic[inkblack, scale=0.85] at (\Ca+0.38,0.86) {boticon};
\node[circle, fill=white, minimum size=0.19cm] at (\Ca+0.58,0.68) {\includegraphics[height=4.6pt]{figures/icons/eye-ink.pdf}};
\node[lb, anchor=west] at (\Ca+0.76,0.97) {consistency review};
\node[lgs, anchor=west] at (\Ca+0.76,0.75) {sees the full task};

\node[bot] (sv) at (\Da+0.42,4.34) {};
\pic[inkblack] at (\Da+0.42,4.34) {boticon};
\node[circle, fill=white, minimum size=0.25cm] at (\Da+0.70,4.56) {};
\pic[evalpurple, scale=1.15] at (\Da+0.70,4.56) {searchicon};
\draw[archgray, line width=0.4pt, fill=white] (\Da+1.26,3.90) -- (\Da+1.26,4.74) -- (\Da+1.88,4.74) -- (\Da+2.00,4.62) -- (\Da+2.00,3.90) -- cycle;
\foreach \y in {4.56,4.42,4.28,4.14,4.02}{\draw[archgray!60, line width=0.5pt] (\Da+1.36,\y) -- (\Da+1.90,\y);}
\draw[flow] (sv.east) -- (\Da+1.24,4.34);
\node[lg, anchor=north] at (\Da+1.63,3.82) {answer};
\draw[archgray, line width=0.4pt, densely dashed] (sv.south) -- (\Da+0.42,3.86);
\draw[archgray, line width=0.4pt, fill=white] (\Da+0.24,3.36) -- (\Da+0.24,3.84) -- (\Da+0.52,3.84) -- (\Da+0.60,3.76) -- (\Da+0.60,3.36) -- cycle;
\draw[archgray!60, line width=0.5pt] (\Da+0.31,3.64) -- (\Da+0.50,3.64);
\draw[archgray!60, line width=0.5pt] (\Da+0.31,3.52) -- (\Da+0.44,3.52);
\node[lg, anchor=north] at (\Da+0.42,3.28) {sources};
\draw[flow] (\Da+1.63,3.56) -- (\Da+1.63,2.94);
\draw[rounded corners=1.2pt, line width=0.5pt, draw=archgray, fill=white] (\Da+1.15,2.88) rectangle (\Da+2.11,1.90);
\fill[archgray!18] (\Da+1.16,2.78) rectangle (\Da+2.10,2.87);
\foreach \y/\m/\c in {2.60/oksmall/buildgreen, 2.42/oksmall/buildgreen, 2.24/oksmall/buildgreen, 2.06/nosmall/feedred}{
  \node[draw=archgray, rounded corners=0.4pt, line width=0.4pt, minimum size=0.145cm] at (\Da+1.31,\y) {};
  \pic[\c] at (\Da+1.31,\y) {\m};
  \draw[archgray!60, line width=0.5pt] (\Da+1.49,\y) -- (\Da+1.99,\y);
}
\node[lg, anchor=north] at (\Da+1.63,1.84) {scoring\\contract};
\pic[inkblack] at (\Da+0.52,2.80) {gavelicon};
\node[bot, minimum size=0.36cm] at (\Da+0.32,2.40) {};
\pic[inkblack, scale=0.72] at (\Da+0.32,2.40) {boticon};
\node[bot, minimum size=0.36cm] at (\Da+0.70,2.40) {};
\pic[inkblack, scale=0.72] at (\Da+0.70,2.40) {boticon};
\node[lgs, anchor=north] at (\Da+0.52,2.16) {two judges,\\a third one\\adjudicates};
\node[rectangle, rounded corners=3pt, draw=feedred, line width=0.5pt, fill=feedred!10, minimum height=0.25cm, inner xsep=3pt,
      font=\figfont\fontsize{5.5}{5.5}\selectfont\bfseries, text=feedred] (any) at (\Da+1.145,1.10) {any failed item fails};

\draw[archgray, line width=0.5pt, densely dashed, -{Stealth[length=1.3mm,width=1.1mm]}]
  (\Da+1.145,0.97) -- (\Da+1.145,0.20) -- (0.72,0.20) -- (0.72,0.50);
\node[lg, fill=white, inner xsep=3pt, inner ysep=1pt] at (6.98,0.20) {failure signals, judge disagreements and specification gaps seed the next discovery round};
\path[use as bounding box] (0,0.02) rectangle (13.97,5.70);
\end{tikzpicture}
\caption{The \pipename{} engine. Each panel hands one artifact to the next: a candidate found in professional sources, a reference analysis verified by independent reproduction, and the frozen task, whose question is written only after the reference analysis is verified and is all the solver receives. A defect returns to the stage that caused it (dashed red).}
\label{fig:pipeline}
\end{figure}

\pipename{} turns sources used in professional practice into tasks whose reference analyses are verified before any question is written.
Its design follows the asymmetry of \S\ref{sec:intro}: an analysis is
hard to solve because its evidence must first be found and interpreted,
but much easier to verify once that evidence is fixed.
The engine therefore builds a reference analysis from evidence in the
sources, verifies it by independent reproduction from the same
evidence, and only then writes the question.

Each delivered task has three parts (Figure~\ref{fig:glance}).
The question states one analytical objective, an information cutoff,
and the allowed tools; it is all the solver sees.
The reference analysis is the complete answer to the question, built
from the sources and hidden from the solver.
It reaches its conclusion through \emph{answer atoms}: the values and categories that a correct answer must contain.
Each atom records its tolerance, the evidence it rests on, and the derivation that produces it.
The scoring contract turns the reference analysis into scored items:
every answer atom, and the fatal conditions that fail an answer
outright.
It also lists the accepted alternatives and maps every requirement of
the question to the atoms that answer it.
\begin{figure}[t]
\centering
\fbox{\parbox{0.96\linewidth}{\scriptsize
\textbf{Question.} SouthState's consolidated allowance for credit losses
(ACL) rose 33\% from December~2024 to June~2025. Is this representative
of higher portfolio loss content, or non-diagnostic once gross-loan
growth and loan-population mix are controlled for? Reconcile the
acquired and non-acquired loan populations at both dates before
judging.\\[2pt]
\textbf{Answer atoms} (8, all required), e.g.\
total gross loans \$33{,}902{,}927K $\to$ \$47{,}267{,}247K (+39.42\%);
ACL coverage 1.3724\% $\to$ 1.3139\%;
\texttt{representativeness\_classification}: \emph{non-diagnostic}.\\[2pt]
\textbf{Fatal conditions} (7), e.g.\
\texttt{incorrect\_judgment}: calls the increase representative of higher loss content;
\texttt{unsupported\_inference}: treats arithmetic growth contributions as acquisition causality.\\[2pt]
\textbf{Accepted alternatives}, e.g.\ ``purchased credit deteriorated'' for PCD; amounts in millions, not thousands.
}}
\caption{A task at a glance: the SouthState allowance task, with part
of its scoring contract. The task's reference analysis and scoring
contract are summarized in Appendix~\ref{case:southstate}.}
\label{fig:glance}
\end{figure}

Sections~\ref{sec:discovery}--\ref{sec:evaluation} follow the four
panels of Figure~\ref{fig:pipeline}.
The complete procedure is described step by step in
Appendix~\ref{app:method}, with the full flowchart
(Figure~\ref{fig:pipeline_full}), what each role is allowed to see
(Figure~\ref{fig:visibility}), and a worked example
(Appendix~\ref{app:ps}).
An orchestrator agent coordinates the flow: it dispatches each step, assesses the result, and decides whether a candidate is repaired or
archived, but writes nothing that enters a task.
Each step is carried out by a separate worker agent that receives only
the inputs the step needs; the worker that reproduces a reference
analysis never sees its conclusion, and a reviewer judges each question first from the solver's view alone.
No worker output is accepted until code checks that every quotation
appears verbatim in the sources, that every arithmetic step recomputes
to its stated result, and that every cross-reference resolves.
A defect found at any step returns the candidate to the earliest step
it affects; a candidate that fails a step outright, or cannot be
repaired within its budget, is archived with its reason
(Figure~\ref{fig:funnel}).

\subsection{Discovery}
\label{sec:discovery}
\label{sec:difficulty}
Sources are registered in the form in which they are obtained: web
pages, structured filing data, PDF documents, tables and workbooks,
audio and video, and images.
Each is rendered into text that can be quoted verbatim, and the rendering is frozen together with the original file.
A candidate is not yet a question but an analysis the sources can
support: a central object, a target conclusion, and the relationships
in the sources that must hold for the conclusion to follow (for the
task of Figure~\ref{fig:glance}: the allowance for credit losses rose
33\% while loans grew 39\%).
Candidates enter through ten discovery routes that start from primary
sources, published analyses, earlier verified tasks, or known model
failures (Appendix~\ref{app:routes}).
A candidate that passes a novelty check, which removes duplicates, goes to a worker that ties each relationship to an evidence fragment (a verbatim quotation with its location) and writes the smallest set of answer atoms needed to evaluate the candidate.
A candidate whose sources do not support a unique answer exits here.

The sources support far more relationships than are worth turning into
tasks, so a difficulty screen runs before any verification and without solving the candidate.
From a view of the candidate with every answer value hidden, a
worker without tools scores the shortest path a professional would
take to the outputs, and a second worker ranks the batch
(Appendix~\ref{app:method}).
The top candidates proceed and the bottom ones are archived.
The middle ones undergo difficulty evolution: a source-supported
relationship that changes the conclusion is added and the result
re-enters.

\subsection{Reference Analysis and Independent Reproduction}
\label{sec:construction}
A reference analysis counts only once a second worker has reproduced
it independently and every atom agrees.
A worker first reopens the sources and treats nothing from discovery as
established.
For each evidence fragment, it checks that the quotation supports the
stated relationship and that its entity, period, unit, basis, and population match those of the analysis, and it searches the sources for omissions that would
change the answer.
It then writes out every computation from the evidence fragments to the
conclusion and produces the reference analysis, including the
conditions under which its conclusion holds (in Figure~\ref{fig:glance}, the atoms include the coverage ratio at each
date).

The reference analysis is then verified by independent reproduction.
A second worker, the reproducer, receives only the analytical
objective, the evidence fragments, and a description of each
required output that does not reveal its value, and must derive every
output from the evidence alone.
Because the evidence is already fixed, the reproducer works on the easier, verification side of the asymmetry.
A third worker compares the two analyses atom by atom.
Agreement is not proof, so this worker also reopens the evidence and
checks the premises the two analyses may share.
Every difference that could change an atom is traced to its cause: a defect in
the task returns the candidate to the earliest affected step for
repair, while an error by the reproducer leads to a fresh reproduction.
Only a candidate whose two analyses are equivalent continues.
The verified analysis is then registered as an instance of a
problem family, defined by its analytical objective, the roles its
evidence plays, and the shape of its relationships and outputs; later candidates may extend a family only after it passes a separate review on multiple instances
(Appendix~\ref{app:method}).

\subsection{Question and Scoring Contract}
\label{sec:compilation}
The question is written last.
Starting from the verified reference analysis, a worker writes the question: one high-level analytical objective, with the basis, period, and information cutoff that make its answer unique; the cutoff is
set from the dates on which the sources became public.
The scoring contract is written with the question, and code checks
that every requirement of the question maps to atoms and every atom to
a requirement.
All three parts are then frozen together as one version; any later change creates a new version
and keeps the old one.

The frozen task then enters the quality gate.
In the validity review, a reviewer sees only what the solver will see
and, with access to public sources, judges whether the question is
professionally meaningful, whether its answer is unique, and whether
it can be answered with the allowed tools from information available
before the cutoff.
Then, in the consistency review, a reviewer receives the full task and
the source files and checks that the
question, the reference analysis, and the scoring contract agree with
one another and that reasonable alternatives are accepted.
When a review fails, an adjudicator confirms each finding against the
task; only confirmed defects are repaired, and the repaired task is
refrozen and reviewed again.
A final pass may polish only the question's wording; code then confirms that every step has passed before delivery.

\subsection{Solving and Judging}
\label{sec:evaluation}
The solver receives the question and the tools of a solving environment
modeled on an expert-annotated benchmark in the domain, and must
locate and read the sources itself.
Two judges rule on every item of the scoring contract, a third settles
their disagreements, and each domain decides how the item rulings combine into a verdict on the whole
answer (\S\ref{sec:setup}).

\section{Production at Scale}
\label{sec:research-set}

\subsection{Preparation}
\label{sec:sources}

The sources of the finance run are the public materials an analyst
works from: periodic and current filings of listed companies and their
exhibits, earnings releases and call transcripts, investor materials,
regulators' orders and letters, and market and macroeconomic data.
The sources of the legal run are the public materials a lawyer works
from: codes, regulations, and the Federal Register; court dockets and
opinions; regulators' enforcement documents and guidance; and transaction
agreements and securities litigation filings.
We collected \nfiles{} finance files across \nindustries{} GICS industry groups and \nlawfiles{} legal files across \nlawareas{} practice areas that we defined; Appendix~\ref{app:source-files} gives their breakdown by channel
and the sources of the energy and medicine runs.

The orchestrator and the workers at every stage ran mainly on GPT-5.6
Sol; steps that require several agents also drew on other model
families, including DeepSeek V4 Flash, Grok~4.6, and Claude Opus~5.

\subsection{Production Run}
\label{sec:results-prod}
\label{sec:funnel}

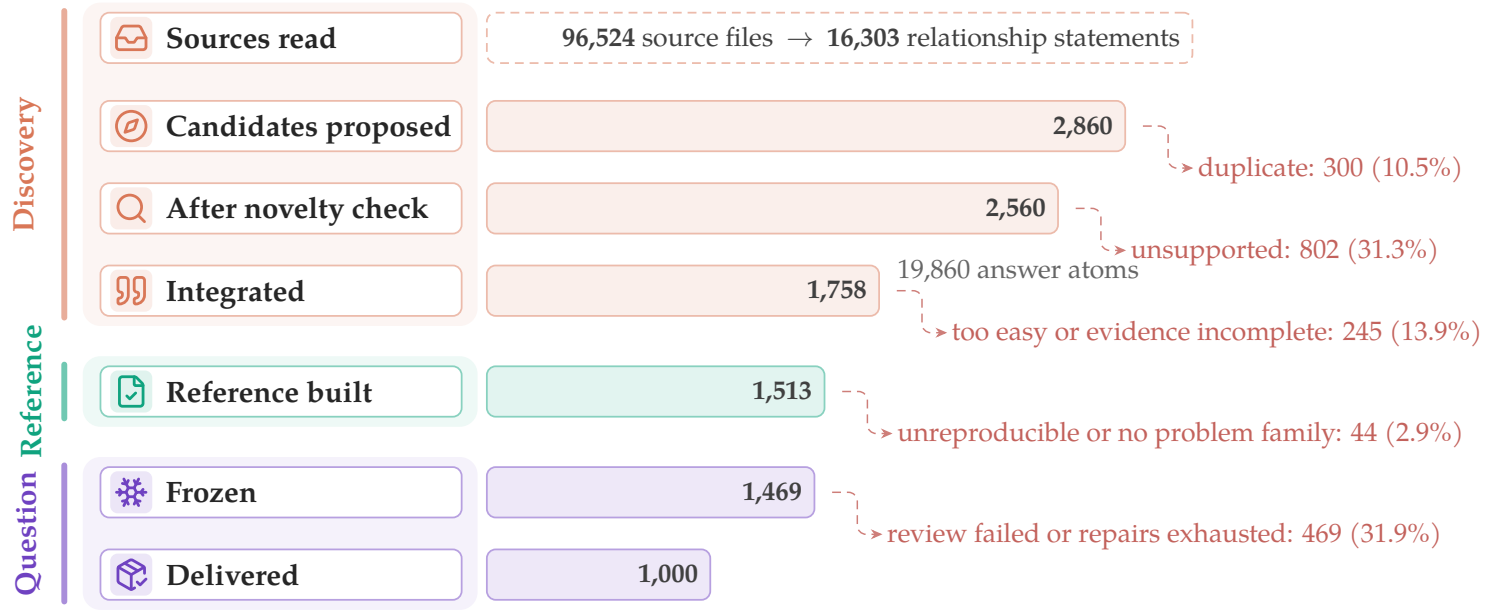
\begin{figure}[t]
\centering
\providecolor{brandclaude}{HTML}{D97757}
\providecolor{brandgpt}{HTML}{10A37F}
\providecolor{brandqwen}{HTML}{6F42C1}
\providecolor{feedred}{HTML}{B9463F}
\resizebox{0.70\textwidth}{!}{%
\begin{tikzpicture}[
  font=\figfont,
  st/.style={rectangle, rounded corners=3pt, draw=#1!50, line width=0.6pt, fill=white, anchor=west, minimum width=4.3cm, minimum height=0.6cm, inner sep=0pt},
  bar/.style={rectangle, rounded corners=3pt, draw=#1!50, line width=0.6pt, fill=#1!12, anchor=west, minimum height=0.6cm, inner sep=0pt},
  badge/.style={rectangle, rounded corners=2.5pt, draw=none, fill=#1!13, minimum size=0.5cm, inner sep=0pt, anchor=west},
  band/.style={rectangle, rounded corners=6pt, draw=none, fill=#1!7},
  stage/.style={anchor=base west, inner sep=0pt, font=\figfont\normalsize\bfseries, text=black!85},
  count/.style={anchor=base east, inner sep=0pt, font=\figfont\footnotesize\bfseries, text=black!75},
  side/.style={anchor=south west, inner sep=0pt, font=\figfont\footnotesize, text=black!60},
  exit/.style={anchor=base west, inner sep=0pt, font=\figfont\footnotesize, text=feedred!85},
  exitarr/.style={-{Stealth[length=1.4mm,width=1.2mm]}, line width=0.5pt, feedred!70, densely dashed, rounded corners=3pt},
  lanelbl/.style={font=\figfont\normalsize\bfseries, rotate=90, anchor=south, text=#1},
]
\def\xb{4.6}\def\xR{4.5}
\def\yZ{0.2}
\def\yA{-0.85}\def\yB{-1.83}\def\yC{-2.81}\def\yD{-4.01}\def\yE{-5.21}\def\yF{-6.19}
\newcommand{\ic}[1]{\vadeicon{#1}{12pt}}
\newcommand{\frow}[7]{%
  \node[st=#2] (s#1) at (0,#7) {};
  \node[badge=#2] at ($(s#1.west)+(0.12,0)$) {\ic{#3}};
  \node[stage] at ($(s#1.west)+(0.78,-0.12)$) {#4};
  \node[bar=#2, minimum width=#5cm] (r#1) at (\xb,#7) {};
  \node[count] at ($(r#1.east)+(-0.15,-0.09)$) {#6};}
\newcommand{\fexit}[2]{%
  \draw[exitarr] ($(r#1.east)+(0.2,0)$) -- ++(0.3,0) |- ($(r#1.east)+(0.8,-0.5)$);
  \node[exit] at ($(r#1.east)+(0.85,-0.59)$) {#2};}
 
\fill[band=brandclaude] (-0.2,\yZ+0.42) rectangle (\xR,\yC-0.42);
\fill[band=brandgpt]    (-0.2,\yD+0.42) rectangle (\xR,\yD-0.42);
\fill[band=brandqwen]   (-0.2,\yE+0.42) rectangle (\xR,\yF-0.42);
\draw[brandclaude!60, line width=2pt, line cap=round] (-0.42,\yZ+0.32) -- (-0.42,\yC-0.32);
\draw[brandgpt!60,    line width=2pt, line cap=round] (-0.42,\yD+0.32) -- (-0.42,\yD-0.32);
\draw[brandqwen!60,   line width=2pt, line cap=round] (-0.42,\yE+0.32) -- (-0.42,\yF-0.32);
\node[lanelbl=brandclaude] at (-0.57,{(\yZ+\yC)/2}) {Discovery};
\node[lanelbl=brandgpt]    at (-0.57,\yD) {Reference};
\node[lanelbl=brandqwen]   at (-0.57,{(\yE+\yF)/2}) {Question};
 
\node[st=brandclaude] (s0) at (0,\yZ) {};
\node[badge=brandclaude] at ($(s0.west)+(0.12,0)$) {\ic{inbox-claude}};
\node[stage] at ($(s0.west)+(0.78,-0.12)$) {Sources read};
\node[bar=brandclaude, minimum width=8.4cm, densely dashed, fill=white] (r0) at (\xb,\yZ) {};
\node[anchor=base east, inner sep=0pt, font=\figfont\footnotesize, text=black!75] at ($(r0.east)+(-0.15,-0.1)$)
  {\textbf{96{,}524} source files\ \ $\rightarrow$\ \ \textbf{16{,}303} relationship statements};
 
\frow{1}{brandclaude}{compass-claude}{Candidates proposed}{7.60}{2{,}860}{\yA}
\frow{2}{brandclaude}{search-claude}{After novelty check}{6.80}{2{,}560}{\yB}
\frow{3}{brandclaude}{quote-claude}{Integrated}{4.67}{1{,}758}{\yC}
\frow{4}{brandgpt}{file-check-gpt}{Reference built}{4.02}{1{,}513}{\yD}
\frow{5}{brandqwen}{snowflake-qwen}{Frozen}{3.90}{1{,}469}{\yE}
\frow{6}{brandqwen}{package-check-qwen}{Delivered}{2.66}{1{,}000}{\yF}
\node[side] at ($(r3.east)+(0.2,0.1)$) {19{,}860 answer atoms};
 
\fexit{1}{duplicate: 300 (10.5\%)}
\fexit{2}{unsupported: 802 (31.3\%)}
\fexit{3}{too easy or evidence incomplete: 245 (13.9\%)}
\fexit{4}{unreproducible or no problem family: 44 (2.9\%)}
\fexit{5}{review failed or repairs exhausted: 469 (31.9\%)}
\end{tikzpicture}}
\caption{Production funnel of the finance run: the candidates remaining after each stage, with the archive reasons between stages and their share of the candidates entering the stage. The first row, the input, is not drawn to scale.}
\label{fig:funnel}
\end{figure}

Figure~\ref{fig:funnel} shows the funnel of the finance run, from
\nfiles{} source files to the \ntasks{} delivered tasks.
On average, one task was delivered for every 2.9 candidates proposed.
Attrition is concentrated before the reference analysis is built and at
review.
The legal run proposed 871 candidates and delivered
\nlawtasks{} tasks.

At least 76.7\% of the delivered tasks needed one or more repairs, and
the repaired tasks went through a median of two repair rounds.
The most common cause of repair was a mismatch between the question and
the scoring contract: a required item the question did not ask for, a
precision stricter than the sources support, or a reasonable equivalent
answer the contract did not accept.
In independent reproduction, the comparison exposed a defect on the
task side, which was then repaired, in 5.9\% of the finance candidates
and 16.2\% of the law candidates.
At first review, the validity review returned 22\% of the finance tasks
and 7\% of the law tasks, and the consistency review 47\% and 38\%.
Among the delivered tasks, the validity review returned 21.2\% of the
finance tasks and 5.6\% of the law tasks at least once, and the
consistency review 44.5\% and 24.0\%
(Appendix~\ref{app:step-contribution}).

Measured in tokens of the constructing model, the median task costs
24.4M tokens (interquartile range 17.7--34.1M).
To keep the tasks verifiable, the engine spends more on verifying a
task than on finding it: evidence and reference construction,
independent reproduction and comparison, and review and repair together account for 34\% of the total, compared
with 28\% for discovery,
integration, and difficulty screening.
By agent rather than by stage, the orchestrator's own judgment
and checking account for 88\% of the total.

\subsection{The Task Sets}
\label{sec:task-composition}

In all, \pipename{} delivered \ntasks{} tasks in finance and
\nlawtasks{} tasks in law, together with preview sets of
\nenergytasks{} energy and \nmedtasks{} medicine tasks from smaller runs.
The finance set covers all \nindustries{} GICS industry groups and uses the nine task categories of Finance Agent
Benchmark~v2~\citep{fabv2_2026}; the legal set covers \nlawareas{}
practice areas and uses the four task types of LAB~\citep{harvey_lab2026}.
On average, a task draws on 1.6 sources and 13.1 evidence fragments,
and its scoring contract has 11.9 answer atoms and 8.0 fatal conditions.

Larger runs of the same engine produced extended sets: 642 law tasks
in the same \nlawareas{} practice areas and four task types,
\energyfull{}, with 211 tasks in ten task categories, and \medfull{},
with 205 tasks in six (Appendix~\ref{app:judging}).

\subsection{Human Review}
\label{sec:human-review}

Two reviewers with backgrounds in finance independently reviewed
\nreviewed{} tasks.
For each task they checked four things: whether the task is realistic and close to the work finance
professionals actually do, whether every
cited value can be found at its stated location in the sources,
whether the core conclusion is correct, and whether a reasonable
alternative exists that the scoring contract does not accept.
At least one reviewer flagged 15 tasks, and one of the reviewers
then re-examined these against the sources.
Of the \nreviewed{} tasks, 93.5\% needed no substantive revision, and
the re-examination confirmed substantive problems in the remaining
12, of which 9 concern scoring fairness.

One expert with a legal background reviewed 50 law tasks under the
same criteria and found a substantive problem in 1.
Two models also reviewed 50 tasks in each of the four domains under the
same criteria: on the finance tasks they found every problem the
experts found, and in energy and medicine they flagged 2 to 4 of the 50
tasks, of which a re-review by the authors confirmed 2 in energy and
none in medicine.
Appendix~\ref{app:human-review} gives the agreement between the
reviewers and the full results.
We will release 100 tasks sampled from the delivered tasks, with their
sources, reference analyses, and review notes, on the project website
(\url{https://tiger-ai-lab.github.io/VADE/}) as a quality showcase.

\section{The Tasks as a Benchmark}
\label{sec:setup}
\label{sec:results-tasks}

\subsection{Setup}

We turn the delivered tasks into a benchmark and evaluate models in
each domain's solving environment.
The finance environment follows the tool interface of the Finance Agent
Benchmark: the solver finds its sources using web search, EDGAR
search, and a document reader, and submits its answer as text.
The legal environment follows the sandbox of LAB: the solver works on
local materials with file tools and delivers files or text.
Both environments are detailed in Appendix~\ref{app:harness}.

Following the implementations of the Finance Agent Benchmark and LAB,
we grade with language-model
judges~\citep{zheng2023judging,judgebench2024} under a fixed procedure.
Every answer is graded by two judges, GPT-5.6 Sol and Claude Opus~5,
item by item under the scoring contract.
In the finance environment, the judge first reads the whole answer and
checks the solver's retrieval record without seeing the scoring
contract, then rules on every item and fatal condition; in the legal environment, it reads the delivered files and rules on every item.
In both, an answer passes only if every item passes.
When the two judges disagree, their itemized findings are passed to a
third judge, whose ruling is final.
The full judging rules are in Appendix~\ref{app:judging}.
On 60 answers that human reviewers graded independently, the judge
and the reviewers agree on 96.7\% of the item rulings
(Appendix~\ref{app:judging}, Table~\ref{tab:judge-human}).

\subsection{Model Performance}
\label{sec:model-perf}

\begin{table}[t]
\caption{Pass rates on \dsname{}, \lawname{}, \energyname{}, and
\medname{} (top to bottom) by model and task category. A task passes
when every item of its scoring contract passes; the highest rate in each
column of a block is in bold. \energyname{} and \medname{} are preview
sets from smaller runs. Category abbreviations are expanded in Appendix~\ref{app:judging}.}
\label{tab:performance}
\centering
\scriptsize
\setlength{\tabcolsep}{3.5pt}
\ifdefined\hcw\else\newlength{\hcw}\fi
\setlength{\hcw}{0.95cm}
\newcommand{\hc}[1]{\makebox[\hcw][r]{#1}}
\newcommand{\law}[4]{\multicolumn{9}{l}{\makebox[\dimexpr 9\hcw+16\tabcolsep\relax]{\hc{#1}\hfill\hc{#2}\hfill\hc{#3}\hfill\hc{#4}}}}
\newcommand{\enrg}[7]{\multicolumn{9}{l}{\makebox[\dimexpr 9\hcw+16\tabcolsep\relax]{\hc{#1}\hfill\hc{#2}\hfill\hc{#3}\hfill\hc{#4}\hfill\hc{#5}\hfill\hc{#6}\hfill\hc{#7}}}}
\newcommand{\med}[5]{\multicolumn{9}{l}{\makebox[\dimexpr 9\hcw+16\tabcolsep\relax]{\hc{#1}\hfill\hc{#2}\hfill\hc{#3}\hfill\hc{#4}\hfill\hc{#5}}}}
\begin{tabular}{@{}lrrrrrrrrrr@{}}
\toprule
\multicolumn{11}{@{}l}{\textbf{\dsname{}}} \\
Model & \hc{Qual.} & \hc{Quant.} & \hc{Mkt.} & \hc{Comp.} & \hc{Prec.} & \hc{Adj.} & \hc{Earn.} & \hc{Discl.} & \hc{Model.} & Overall \\
\midrule
Grok~4.6          & \textbf{54.1} & \textbf{76.2} & \textbf{69.4} & \textbf{65.5} & 55.0 & \textbf{68.1} & \textbf{65.3} & \textbf{65.7} & \textbf{63.0} & \textbf{66.4} \\
DeepSeek V4 Flash & 24.3 & 46.5 & 44.4 & 51.7 & \textbf{60.0} & 48.9 & 40.6 & 40.1 & 42.5 & 42.9 \\
Claude Opus~5     & 18.9 & 49.5 & 30.6 & 31.0 & 35.0 & 40.4 & 39.6 & 40.1 & 30.1 & 37.9 \\
Qwen3.8-27B       & 27.0 & 36.6 & 25.0 & 34.5 & 30.0 & 34.0 & 37.6 & 32.8 & 37.0 & 34.1 \\
GLM-5.2           & 21.6 & 33.7 & 36.1 & 31.0 & 20.0 & 23.4 & 25.7 & 28.5 & 23.3 & 27.7 \\
MiniMax-M3        & 5.4 & 28.7 & 16.7 & 17.2 & 20.0 & 21.3 & 22.8 & 16.1 & 16.4 & 19.4 \\
\midrule[\heavyrulewidth]
\multicolumn{11}{@{}l}{\textbf{\lawname{}}} \\
Model & \law{Analyze}{Draft}{Review}{Research} & Overall \\
\midrule
Claude Opus~5     & \law{\textbf{73.6}}{41.7}{64.3}{\textbf{76.0}} & \textbf{66.0} \\
Grok~4.6          & \law{70.6}{37.5}{\textbf{68.3}}{72.0} & 64.5 \\
DeepSeek V4 Flash & \law{52.9}{31.8}{47.5}{56.0} & 48.6 \\
Qwen3.8-27B       & \law{41.5}{\textbf{45.8}}{39.0}{41.7} & 41.6 \\
GLM-5.2           & \law{37.7}{20.8}{52.4}{40.0} & 39.6 \\
MiniMax-M3        & \law{20.8}{25.0}{22.0}{25.0} & 22.5 \\
\midrule[\heavyrulewidth]
\multicolumn{11}{@{}l}{\textbf{\energyname{}}} \\
Model & \enrg{Reg.}{Mkt.}{Gen.}{Rel.}{Scen.}{Stor.}{Dem.} & Overall \\
\midrule
Grok~4.6          & \enrg{46.9}{\textbf{100.0}}{\textbf{83.3}}{0.0}{71.4}{28.6}{\textbf{71.4}} & \textbf{59.8} \\
DeepSeek V4 Flash & \enrg{46.9}{16.7}{58.3}{\textbf{60.0}}{\textbf{100.0}}{0.0}{28.6} & 45.0 \\
Claude Opus~5     & \enrg{\textbf{50.0}}{58.3}{41.7}{0.0}{71.4}{\textbf{40.0}}{0.0} & 43.6 \\
GLM-5.2           & \enrg{37.5}{16.7}{41.7}{0.0}{71.4}{28.6}{28.6} & 34.1 \\
Qwen3.8-27B       & \enrg{31.3}{16.7}{\textbf{83.3}}{0.0}{28.6}{0.0}{40.0} & 32.1 \\
MiniMax-M3        & \enrg{6.3}{0.0}{16.7}{0.0}{28.6}{0.0}{40.0} & 10.3 \\
\midrule[\heavyrulewidth]
\multicolumn{11}{@{}l}{\textbf{\medname{}}} \\
Model & \med{Reg.}{Trial}{Discl.}{Safety}{Label} & Overall \\
\midrule
Grok~4.6          & \med{\textbf{75.0}}{\textbf{76.2}}{\textbf{85.7}}{\textbf{57.1}}{\textbf{55.6}} & \textbf{73.4} \\
Claude Opus~5     & \med{40.0}{33.3}{35.7}{28.6}{\textbf{55.6}} & 38.3 \\
DeepSeek V4 Flash & \med{32.1}{47.4}{35.7}{0.0}{\textbf{55.6}} & 36.4 \\
GLM-5.2           & \med{7.1}{23.8}{64.3}{28.6}{0.0} & 22.8 \\
Qwen3.8-27B       & \med{23.3}{9.5}{50.0}{0.0}{22.2} & 22.2 \\
MiniMax-M3        & \med{0.0}{9.5}{0.0}{0.0}{0.0} & 2.5 \\
\bottomrule
\end{tabular}
\end{table}

On the finance tasks, overall pass rates range from 19.4\% to 66.4\%.
Grok~4.6 leads by more than twenty points; DeepSeek
V4 Flash, Claude Opus~5, and Qwen3.8-27B cluster between 34\% and
43\%; GLM-5.2 and MiniMax-M3 are lower.
Qualitative analysis is the weakest category for four of the six
models: Grok reaches only 54.1\% on it, Opus 18.9\%, and
MiniMax 5.4\%.
Precedents and comparables break the overall order: DeepSeek passes 60.0\% of precedent tasks, more than Grok.

On the legal tasks, Claude Opus~5 and Grok~4.6 are close, at 66.0\%
and 64.5\%, and the other four models keep the order they have in
finance.
Opus leads on the analysis and research tasks and Grok on review.
Drafting is the weakest task type for four of the six models, yet Qwen3.8-27B has the highest drafting score of any model.
On the two preview sets, Grok~4.6 again leads, at 59.8\% in energy and
73.4\% in medicine, and MiniMax-M3 is again last.

Results of six more recent models on finance tasks and on the extended
law, energy, and medicine sets are in Appendix~\ref{app:recent-models}, and turn, time, and length statistics in Appendix~\ref{app:harness}.

\subsection{Failure Analysis}
\label{sec:failure}

For every failed answer, the judge records which of six whole-answer checks it fails; one answer can fail several.
Unsupported facts and inferences are by far the most common, present in 85.9\% of failed answers.
The SouthState task of Figure~\ref{fig:glance} illustrates the pattern:
the allowance for credit losses rose 33\% in nominal terms, and the
sources determine that this rise is non-diagnostic, because gross loans
grew 39\% over the same period and acquisitions changed the loan mix,
so coverage actually fell.
All six models reach this conclusion, but each then adds a claim: three
attribute the change in the allowance to the acquisitions, an inference
the sources do not support and one of the task's fatal conditions
(Appendix~\ref{case:southstate}).
Internal inconsistencies appear in 64.8\% of failed answers and unmet
requirements in 57.2\%, usually alongside an unsupported inference;
errors of scope or period (37.6\%), computation or units (32.6\%), and
professional concept (29.6\%) are less frequent but rarely occur alone.
Appendix~\ref{app:failures} gives the distribution by model and task
type, the classification procedure, and case studies for each kind.

\section{The Tasks as Supervision}
\label{sec:results-learn}
\label{sec:train-protocol}
\label{sec:train-results}

\subsection{Setup}

We split the finance tasks into \ntraintasks{} training and
\ntesttasks{} test tasks, keeping tasks that share a source, a family,
or near-duplicate content on the same side and matching the
industry and category mix across sides.
On the training tasks we fine-tune Qwen3.5-9B and Qwen3.5-27B on verified solver trajectories (Appendix~\ref{app:training}), and
evaluate on the held-out test tasks, \dstest{}, with the solving
environment and judges of \S\ref{sec:setup}, reporting the pass rate
and the item pass rate, the share of an answer's scored items that pass.
We also evaluate on the public questions of four external finance
benchmarks, FAB~v1~\citep{fabv1_2025}, FAB~v2~\citep{fabv2_2026},
BigFinanceBench~\citep{bigfinancebench2026}, and
FrontierFinance~\citep{frontierfinance2026}, with a leakage filter
(Appendix~\ref{app:training}).
In law, energy, and medicine, we fine-tune Qwen3.5-9B for four epochs
on 182, 94, and 68 training tasks and evaluate on 63, 25, and 23
held-out tasks; in each domain, the training and test tasks share no
task and no source.
We also evaluate on LAB in law and on the public tasks of LifeSci
Diamond~\citep{lifescidiamond2026} in medicine
(Appendix~\ref{app:training}).

\subsection{Results}

\begin{table}[t]
\caption{Pass rate (P) and item pass rate (R) of Qwen3.5-9B and
Qwen3.5-27B before fine-tuning and after 4 and 10 epochs, on \dstest{}
and on the public questions of four external finance benchmarks. The best value of each model in each column is in
bold.}
\label{tab:qwen}
\centering
\scriptsize
\setlength{\tabcolsep}{3pt}
\begin{tabular}{@{}ll*{10}{w{r}{0.7cm}}@{}}
\toprule
 & & \multicolumn{2}{c}{\shortstack{VADE-Finance-\\Test}}
 & \multicolumn{2}{c}{FAB v1} & \multicolumn{2}{c}{FAB v2}
 & \multicolumn{2}{c}{\shortstack{BigFinance-\\Bench}}
 & \multicolumn{2}{c}{\shortstack{Frontier-\\Finance}} \\
\cmidrule(lr){3-4}\cmidrule(lr){5-6}\cmidrule(lr){7-8}\cmidrule(lr){9-10}\cmidrule(l){11-12}
Model & Stage & P & R & P & R & P & R & P & R & P & R \\
\midrule
Qwen3.5-9B  & base     & 0.0 & 40.8 & 30.0 & 46.5 & 11.1 & 35.4 & 6.0 & 30.0 & 1.0 & 17.1 \\
            & epoch 4  & 0.0 & 54.8 & 42.0 & 71.5 & \textbf{25.9} & \textbf{63.6} & 10.0 & 56.1 & 1.0 & 38.6 \\
            & epoch 10 & 0.0 & \textbf{73.1} & \textbf{54.0} & \textbf{78.5} & \textbf{25.9} & \textbf{63.6} & \textbf{18.0} & \textbf{58.7} & \textbf{3.0} & \textbf{40.8} \\
\midrule
Qwen3.5-27B & base     & 6.7 & 70.1 & 58.0 & 79.6 & 29.6 & 61.0 & 14.0 & 41.8 & 0.0 & 24.1 \\
            & epoch 4  & 17.3 & \textbf{85.3} & 58.0 & 79.6 & 40.7 & 75.7 & \textbf{22.0} & \textbf{64.3} & 0.0 & 41.2 \\
            & epoch 10 & \textbf{21.3} & \textbf{85.3} & \textbf{62.0} & \textbf{82.6} & \textbf{51.9} & \textbf{81.2} & \textbf{22.0} & 59.2 & \textbf{4.0} & \textbf{43.7} \\
\bottomrule
\end{tabular}
\end{table}

On \dstest{}, after ten epochs Qwen3.5-27B raises its pass rate from
6.7\% to 21.3\% and its item pass rate from 70.1\% to 85.3\%.
Qwen3.5-9B passes no whole task, but its item pass rate rises from
40.8\% to 73.1\%.

Both models also improve on all four external benchmarks.
The largest gains in pass rate are on FAB~v2 for Qwen3.5-27B, from
29.6\% to 51.9\%, and on FAB~v1 for Qwen3.5-9B, from 30.0\% to 54.0\%.
The item pass rate mostly rises more than the pass rate, and since training
used only \pipename{} tasks, the external gains suggest the tasks teach a
transferable procedure rather than answers.

\begin{table}[t]
\caption{Pass rate (P) and item pass rate (R) of Qwen3.5-9B before
fine-tuning and after 4 epochs in law, energy, and medicine, on the
held-out \pipename{} tasks of each domain and on two external
benchmarks. The best value in each column is in bold.}
\label{tab:qwen-domains}
\centering
\scriptsize
\setlength{\tabcolsep}{3pt}
\begin{tabular}{@{}ll*{10}{w{r}{0.7cm}}@{}}
\toprule
 & & \multicolumn{4}{c}{Law} & \multicolumn{2}{c}{Energy} & \multicolumn{4}{c}{Medicine} \\
\cmidrule(lr){3-6}\cmidrule(lr){7-8}\cmidrule(l){9-12}
 & & \multicolumn{2}{c}{\shortstack{VADE-Law-\\Test}}
 & \multicolumn{2}{c}{LAB}
 & \multicolumn{2}{c}{\shortstack{VADE-Energy-\\Test}}
 & \multicolumn{2}{c}{\shortstack{VADE-Medicine-\\Test}}
 & \multicolumn{2}{c}{\shortstack{LifeSci\\Diamond}} \\
\cmidrule(lr){3-4}\cmidrule(lr){5-6}\cmidrule(lr){7-8}\cmidrule(lr){9-10}\cmidrule(l){11-12}
Model & Stage & P & R & P & R & P & R & P & R & P & R \\
\midrule
Qwen3.5-9B  & base     & 1.6 & 27.6 & 0.8 & 47.5 & 0.0 & 43.6 & 0.0 & 46.9 & 0.0 & 26.4 \\
            & epoch 4  & \textbf{12.7} & \textbf{44.7} & \textbf{4.1} & \textbf{63.3} & \textbf{8.0} & \textbf{67.9} & \textbf{13.0} & \textbf{53.8} & 0.0 & \textbf{42.8} \\
\bottomrule
\end{tabular}
\end{table}

Table~\ref{tab:qwen-domains} gives the results in the other three
domains.
On the held-out \pipename{} tasks, the pass rate rises from 1.6\% to
12.7\% in law, from 0.0\% to 8.0\% in energy, and from 0.0\% to 13.0\%
in medicine, and the item pass rate rises in all three.
On LAB, the pass rate rises from 0.8\% to 4.1\% and the item pass rate
from 47.5\% to 63.3\%.
On LifeSci Diamond, no task passes in full before or after fine-tuning,
and the item pass rate rises from 26.4\% to 42.8\%.

\section{Related Work}
\label{sec:related}

\begin{table}[t]
\caption{Automated task construction methods compared.
``Indep.'': a second agent reproduces the reference from the evidence
without seeing it; rejection sampling and solver panels re-solve the
task but do not rebuild the reference.
\checkmark: yes; \textasciitilde: partial; --: no.}
\label{tab:related}
\centering\scriptsize
\setlength{\tabcolsep}{3pt}
\renewcommand{\arraystretch}{1.08}
\newcolumntype{L}[1]{>{\raggedright\arraybackslash}p{#1}}
\begin{tabular}{@{}L{3.35cm}L{2.05cm}L{2.55cm}L{2.95cm}c@{}}
\toprule
 & Deliverable & Reference from & Verification & Indep. \\
\midrule
\multicolumn{5}{@{}l}{\emph{Execution oracle available}} \\
SWE-Gym, SWE-smith, R2E-Gym
                                     & code patch        & repository tests        & execution                  & -- \\
CLI-Universe & command sequence  & technical materials     & fail-to-pass tests         & -- \\
TaskCraft      & short answer      & tool execution          & rejection sampling         & \textasciitilde \\
\midrule
\multicolumn{5}{@{}l}{\emph{No execution oracle}} \\
FinForge       & multiple choice   & LM synthesis, curated corpus & LM validation, expert spot-check & -- \\
AGORA              & single value      & workplace documents     & solver filtering           & \textasciitilde \\
SPARK              & short answer      & knowledge-graph path    & graph facts                & -- \\
WebSailor-V2   & short answer      & entity in web-expanded graph & LM judge              & -- \\
QUEST              & long-form report  & synthesized rubric tree & rubric reward              & -- \\
\midrule
\pipename{} (ours)                   & complete analysis, item contract & professional sources & independent reproduction, cross-family review & \checkmark \\
\bottomrule
\end{tabular}
\end{table}

\paragraph{Automated task construction.}
Methods differ in where the reference answer comes from and what
checks it (Table~\ref{tab:related}).
With an execution oracle, a task is verified by running it: software
tasks against repository
tests~\citep{swegym2024,swesmith2025,r2egym2025,swenext2026}, terminal
tasks against fail-to-pass tests~\citep{cliuniverse2026}, and tool
tasks by re-executing the tool~\citep{taskcraft2026,apigenmt2025}.
Without one, existing methods narrow the task until it can be checked,
to multiple choice~\citep{finforge2026}, a single value over workplace
documents~\citep{agora2026}, or an answer fixed by a knowledge
graph~\citep{spark2026,websailor2026}, or keep a long-form deliverable
but trust a reference from the same process that wrote the
task~\citep{quest2026,sweqapro2026}.
\pipename{} keeps the deliverable a complete analysis and replaces the
oracle with a reference reproduced independently from the sources.

\paragraph{Professional benchmarks and agentic training.}
Expert-authored benchmarks grade professional deliverables against
expert judgment or rubrics, across occupations~\citep{gdpval2025,apexagents2026,apex2026},
in finance~\citep{fabv1_2025,fabv2_2026,bigfinancebench2026,frontierfinance2026,hedgebench2026},
and in law~\citep{harvey_lab2026}; they define the task format and
grading that \pipename{} reproduces without expert authoring, and we
adopt their solving environments (\S\ref{sec:setup}) and public
questions as transfer targets (\S\ref{sec:results-learn}).
Our training is the rejection-sampling recipe of STaR~\citep{star2022}
as applied to agents~\citep{agenttuning2023,swegym2024}, with the
scoring contract as the filter; neighboring methods go on to
reinforcement learning on their own
tasks~\citep{taskcraft2026,quest2026,websailor2026,spark2026}, for
which the item-level contract is a reward of the same kind
(Appendix~\ref{app:related}).

\section{Discussion}
\label{sec:discussion}

\subsection{Which Steps Need a Frontier Model}
\label{sec:ablation}
Appendix~\ref{app:ablation-findings} reports the ablations: direct
generation without the engine (\ref{app:direct-generation}), a
simple inspector in place of independent reproduction
(\ref{app:direct-review}), and what each step contributes
(\ref{app:step-contribution}).
Here we ask how capable a model each step needs.
We replayed the complete production trajectory of one finance task
with the open-weight Qwen3.8-27B, a capable solver
(Table~\ref{tab:performance}): at each of the 81 model steps, it redid
the step from a snapshot of the environment, and we rated both its
output and the original on whether each could be used downstream as
is (Table~\ref{tab:replacement}).
Qwen's output was usable as is at 27 steps and unusable at 21, compared
with 57 and 11 for the original run.
It matches the original on well-bounded steps, such as the novelty
check and numerical reconstruction, but falls short on steps that call
for judgment or self-checking: it passes a task after explaining away
a conflict it has noticed, and it reports checks it has not completed.
These steps require more than the ability to solve the tasks
(Appendix~\ref{app:replacement}).

\begin{table}[t]
\caption{Outputs of the original run and of Qwen3.8-27B at the 81 model
steps of one finance task: usable as is, needs fixes (the main work can
be kept), or unusable (a key decision is wrong or no complete output is
produced). A step counts under a kind of error if it made the error or,
when its job was to check, let it pass.}
\label{tab:replacement}
\centering
\scriptsize
\setlength{\tabcolsep}{3.5pt}
\begin{tabular}{@{}lcccc@{}}
\toprule
 & \multicolumn{2}{c}{Original run} & \multicolumn{2}{c}{Qwen3.8-27B} \\
\cmidrule(lr){2-3}\cmidrule(l){4-5}
Usable as is & \multicolumn{2}{c}{57} & \multicolumn{2}{c}{27} \\
\midrule
Kind of error & Needs fixes & Unusable & Needs fixes & Unusable \\
\midrule
Unstated requirement & 5 & 9 & 6 & 8 \\
Wrong decision & 2 & 1 & 3 & 8 \\
Unverified claim & 1 & 1 & 7 & 1 \\
Incomplete handoff & 4 & 0 & 7 & 0 \\
Wrong format & 1 & 0 & 6 & 0 \\
Factual slip & 0 & 0 & 4 & 1 \\
No output & 0 & 0 & 0 & 3 \\
\midrule
Total & 13 & 11 & 33 & 21 \\
\bottomrule
\end{tabular}
\end{table}

\subsection{Findings from Production}
\label{sec:findings}
\paragraph{The answer comes before the question.}
\label{sec:request-last}
If the question is written first and an agent then searches the sources
for its answer, the question tends to fix requirements that the evidence
may not support: of 20 such attempts, 15 reached an answer that
appeared to pass verification, but five of them first needed details
corrected, against one of the same 20 when the question is written last.
A task built this way can also be unreliable even when
its answer is reproduced independently: in one case, the question was written from
the selected evidence alone, while the sources supported a better
answer.
This is why \pipename{} writes the question last
(Appendix~\ref{app:findings}).

\paragraph{Length is not difficulty.}
\label{sec:hardness}
More sub-questions do not make a task harder: across tasks with 5 to more than 20 contract items, the six models'
pooled pass rate stays between 33\% and 40\%.
Instead, models fail at judgments that change the conclusion, such as
which scope, version, or definition of the data applies; difficulty
evolution therefore adds such a relationship rather than more sub-questions
(\S\ref{sec:discovery}, Appendix~\ref{app:findings}).

\subsection{Limitations and Future Work}
\label{sec:limitations}
\paragraph{Model and judge bias.}
An automated engine carries the biases of its models into the tasks,
and model judges carry their own; sources from professional practice,
workers from several model families, and judging procedures taken from
expert-annotated benchmarks reduce these biases but cannot remove them
(Appendix~\ref{app:judging}).
\paragraph{Dependence on the sources.}
The engine can only pose questions whose answers the collected sources
determine: every atom must rest on evidence that can be quoted and
recomputed, so analyses that need unpublished or unquotable data cannot become tasks, and
domains with a richer public record yield more and harder tasks.
\paragraph{Cost.}
A task costs a median of 24.4M constructing-model tokens
(\S\ref{sec:results-prod}).
Reusing verified problem families and moving substitutable steps to cheaper models would lower it.
\paragraph{Reinforcement learning.}
We fine-tuned only on verified solver trajectories; the answer atoms and scoring contracts can also serve as verifiable rewards for reinforcement learning, which we leave to future work.

\section{Conclusion}
\label{sec:conclusion}
We introduced \pipename{}, a data engine built on the asymmetry between
solving and verifying a professional analysis: each reference analysis
is verified by independent reproduction before the question is written,
so tasks stay open-ended without expert authoring.
The resulting tasks pass expert review at a high rate, expose where
current models fail, and improve a smaller model when used for training.
These results show the potential of scaling verifiable agent data for
professional work beyond what experts can write.

\bibliography{bib/references}
\bibliographystyle{vade_preprint}

\clearpage
\appendix
\phantomsection
\label{list:list_of_appendix}
\setcounter{tocdepth}{2}
\DoToC
\clearpage

\section{Task Construction}
\label{app:construction}
\subsection{Construction Procedure}
\label{app:method}

\begin{figure}[t]
\centering
\providecolor{brandclaude}{HTML}{D97757}
\providecolor{brandgpt}{HTML}{10A37F}
\providecolor{brandqwen}{HTML}{6F42C1}
\providecolor{branddeepseek}{HTML}{4D6BFE}
\providecolor{feedred}{HTML}{B9463F}
\resizebox{\textwidth}{!}{%
\begin{tikzpicture}[
  font=\figfont,
  st/.style={rectangle, rounded corners=3pt, draw=#1!50, line width=0.6pt, fill=white, anchor=north west,
             minimum width=\W cm, minimum height=\H cm, inner sep=0pt},
  badge/.style={rectangle, rounded corners=2.5pt, draw=none, fill=#1!13, minimum size=0.56cm, inner sep=0pt, anchor=north west},
  ttl/.style={anchor=base west, inner sep=0pt, font=\figfont\normalsize\bfseries},
  txt/.style={anchor=north west, inner sep=0pt, text width=\TW cm, align=flush left, font=\figfont\footnotesize, text=black!60},
  hid/.style={txt, text width=\TWH cm, text=feedred!85},
  lanebg/.style={rectangle, rounded corners=6pt, draw=none, fill=#1!7},
  lane/.style={font=\figfont\normalsize\bfseries, text=#1, rotate=90, anchor=south},
  flow/.style={-{Stealth[length=2mm,width=1.6mm]}, line width=0.7pt, black!45},
  turn/.style={flow, rounded corners=6pt},
  rep/.style={-{Stealth[length=1.8mm,width=1.4mm]}, line width=0.7pt, feedred!85, densely dashed, rounded corners=6pt},
  hand/.style={font=\figfont\footnotesize\bfseries, text=black!60, fill=white, inner sep=2pt},
  lab/.style={font=\figfont\scriptsize, text=feedred!85, anchor=north west, inner sep=1.5pt},
  ex/.style={font=\figfont\scriptsize, text=feedred!85, anchor=north west, inner sep=1pt},
]
\def\W{3.9}\def\H{2.25}\def\TW{3.6}\def\TWH{3.2}\def\dx{4.4}
\def\ya{0}\def\yb{-4.1}\def\yc{-8.2}\def\yd{-12.3}
\def\laneL{-0.3}\def\laneR{3*\dx+\W+0.75}
\newcommand{\step}[7]{%
  \node[st=#1] (#2) at (#3,#4) {};
  \node[badge=#1] at ($(#2.north west)+(0.15,-0.15)$) {\vadeicon{#5}{13pt}};
  \node[ttl] at ($(#2.north west)+(0.85,-0.55)$) {#6};
  \node[txt] (#2txt) at ($(#2.north west)+(0.15,-0.82)$) {#7};}
\newcommand{\hidden}[2]{%
  \node[anchor=north west, inner sep=0pt] at ($(#1txt.south west)+(0,-0.11)$) {\includegraphics[height=8pt]{figures/icons/eye-off-red.pdf}};
  \node[hid] at ($(#1txt.south west)+(0.4,-0.10)$) {#2};}
 
\fill[lanebg=brandclaude]   (\laneL,\ya+0.3) rectangle (\laneR,\ya-3.55);
\fill[lanebg=brandgpt]      (\laneL,\yb+0.3) rectangle (\laneR,\yb-3.55);
\fill[lanebg=brandqwen]     (\laneL,\yc+0.3) rectangle (\laneR,\yc-3.55);
\fill[lanebg=branddeepseek] (\laneL,\yd+0.3) rectangle (2*\dx+\W+0.3,\yd-2.6);
\fill[lanebg=brandclaude]   (3*\dx-0.3,\yd+0.3) rectangle (\laneR,\yd-2.6);
 
\draw[brandclaude!60,   line width=2pt, line cap=round] (\laneL-0.2,\ya+0.15) -- (\laneL-0.2,\ya-3.4);
\draw[brandgpt!60,      line width=2pt, line cap=round] (\laneL-0.2,\yb+0.15) -- (\laneL-0.2,\yb-3.4);
\draw[brandqwen!60,     line width=2pt, line cap=round] (\laneL-0.2,\yc+0.15) -- (\laneL-0.2,\yc-3.4);
\draw[branddeepseek!60, line width=2pt, line cap=round] (\laneL-0.2,\yd+0.15) -- (\laneL-0.2,\yd-2.45);
\node[lane=brandclaude]   at (\laneL-0.35,\ya-1.625) {Discovery};
\node[lane=brandgpt]      at (\laneL-0.35,\yb-1.625) {Reference};
\node[lane=brandqwen]     at (\laneL-0.35,\yc-1.625) {Question};
\node[lane=branddeepseek] at (\laneL-0.35,\yd-1.15) {Evaluation};
 
\step{brandclaude}{disc}{0}{\ya}{compass-claude}{Discovery}{Candidate via one of ten routes}
\step{brandclaude}{nov}{\dx}{\ya}{search-claude}{Novelty check}{Against delivered and in-progress tasks}
\step{brandclaude}{int}{2*\dx}{\ya}{quote-claude}{Integration}{Exact-quote bindings; answer atoms}
\step{brandclaude}{dif}{3*\dx}{\ya}{gauge-claude}{Difficulty screen}{Structural card; closed-batch ranking}
\hidden{dif}{Answer values}
 
\step{brandgpt}{egt}{0}{\yb}{file-check-gpt}{Evidence \& reference}{Fragments, derivation, atoms, conditions}
\step{brandgpt}{rpr}{\dx}{\yb}{calculator-gpt}{Reproduction}{Atoms rebuilt from evidence alone}
\hidden{rpr}{Reference and derivation}
\step{brandgpt}{cmp}{2*\dx}{\yb}{scale-gpt}{Comparison}{Atom by atom; conflicts attributed}
\step{brandgpt}{fam}{3*\dx}{\yb}{layers-gpt}{Problem family}{Contract and admission}
 
\step{brandqwen}{cmpl}{0}{\yc}{snowflake-qwen}{Compile \& freeze}{Question; requirements $\leftrightarrow$ atoms; fatal conditions}
\step{brandqwen}{gate}{\dx}{\yc}{shield-check-qwen}{Quality gate}{Validity, then consistency}
\hidden{gate}{Reference, contract, sources}
\step{brandqwen}{pol}{2*\dx}{\yc}{pen-line-qwen}{Polish}{Solver-facing wording; pass, edit or return}
\step{brandqwen}{acc}{3*\dx}{\yc}{package-check-qwen}{Acceptance}{Chain verified; package delivered}
 
\step{branddeepseek}{sol}{0}{\yd}{bot-deepseek}{Solver}{Question, tools, cutoff; finds sources itself}
\hidden{sol}{Reference and contract}
\step{branddeepseek}{jud}{\dx}{\yd}{gavel-deepseek}{Judge}{Audit, then per-item pass/fail}
\step{branddeepseek}{sig}{2*\dx}{\yd}{activity-deepseek}{Signals}{Failures, judge disagreement, specification issues}
\node[st=brandclaude] (entry) at (3*\dx,\yd) {};
\node[badge=brandclaude] at ($(entry.north west)+(0.15,-0.15)$) {\includegraphics[height=13pt]{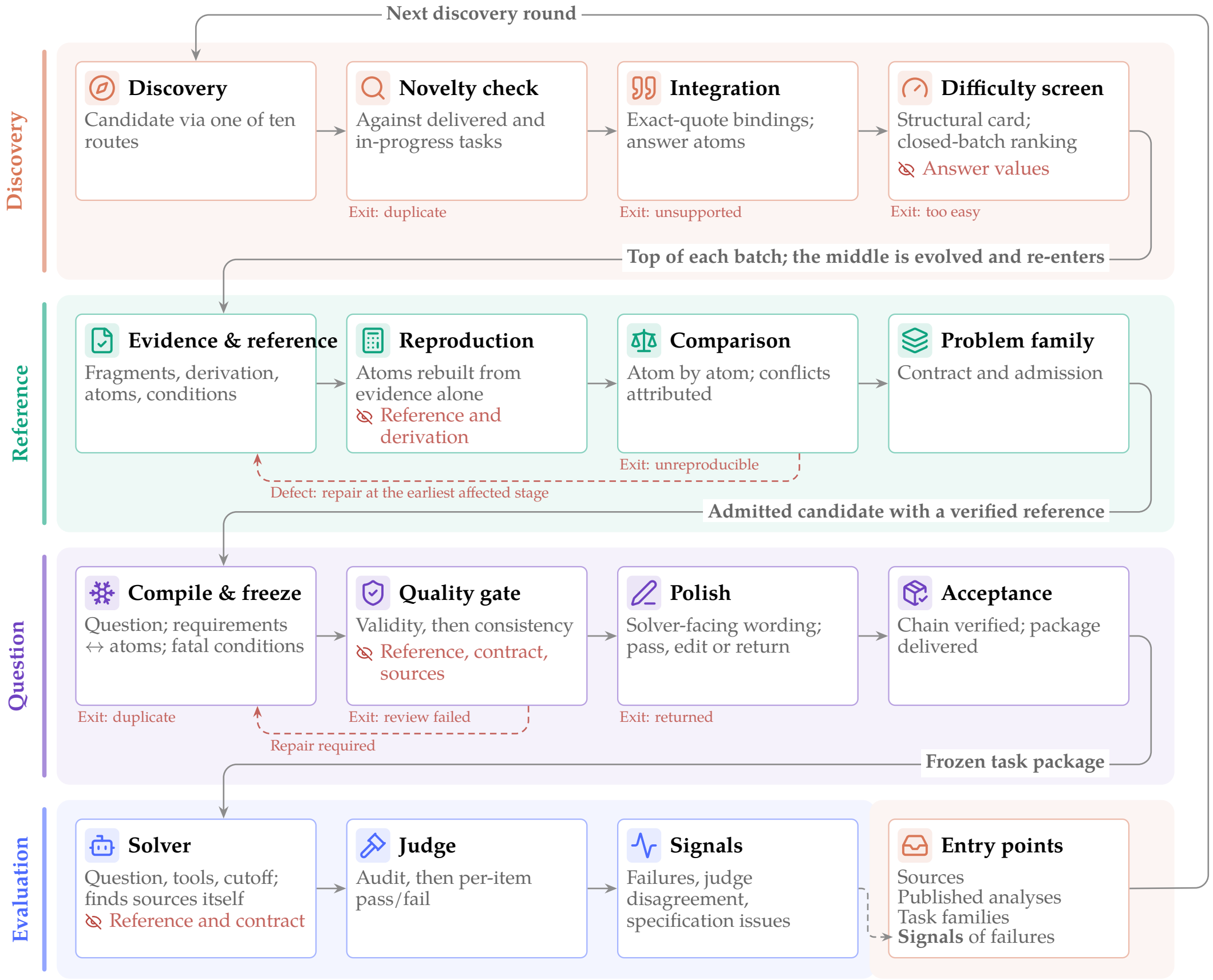}};
\node[ttl] at ($(entry.north west)+(0.85,-0.55)$) {Entry points};
\node[txt] (e1) at ($(entry.north west)+(0.15,-0.82)$) {Sources};
\node[txt] (e2) at ($(entry.north west)+(0.15,-1.14)$) {Published analyses};
\node[txt] (e3) at ($(entry.north west)+(0.15,-1.46)$) {Task families};
\node[txt] (e4) at ($(entry.north west)+(0.15,-1.78)$) {\textbf{Signals} of failures};
 
\foreach \a/\b in {disc/nov, nov/int, int/dif, egt/rpr, rpr/cmp, cmp/fam, cmpl/gate, gate/pol, pol/acc, sol/jud, jud/sig}
  \draw[flow] (\a) -- (\b);
\draw[turn] (dif.east) -- ++(0.35,0) |- ($(dif.south)+(0,-0.95)$) -| ($(egt.north)+(0.45,0)$);
\draw[turn] (fam.east) -- ++(0.35,0) |- ($(fam.south)+(0,-0.95)$) -| ($(cmpl.north)+(0.45,0)$);
\draw[turn] (acc.east) -- ++(0.35,0) |- ($(acc.south)+(0,-0.95)$) -| ($(sol.north)+(0.45,0)$);
\node[hand, anchor=east] at (3*\dx+\W-0.3, \yb+0.9) {Top of each batch; the middle is evolved and re-enters};
\node[hand, anchor=east] at (3*\dx+\W-0.3, \yc+0.9) {Admitted candidate with a verified reference};
\node[hand, anchor=east] at (3*\dx+\W-0.3, \yd+0.9) {Frozen task package};
 
\draw[rep] ($(cmp.south)+(1.0,0)$) -- ++(0,-0.45) -| ($(egt.south)+(1.0,0)$);
\node[lab] at ($(egt.south)+(1.15,-0.48)$) {Defect: repair at the earliest affected stage};
\draw[rep] ($(gate.south)+(1.0,0)$) -- ++(0,-0.45) -| ($(cmpl.south)+(1.0,0)$);
\node[lab] at ($(cmpl.south)+(1.15,-0.48)$) {Repair required};
 
\node[ex] at ($(nov.south west)+(0,-0.04)$) {Exit: duplicate};
\node[ex] at ($(int.south west)+(0,-0.04)$) {Exit: unsupported};
\node[ex] at ($(dif.south west)+(0,-0.04)$) {Exit: too easy};
\node[ex] at ($(cmp.south west)+(0,-0.04)$) {Exit: unreproducible};
\node[ex] at ($(cmpl.south west)+(0,-0.04)$) {Exit: duplicate};
\node[ex] at ($(gate.south west)+(0,-0.04)$) {Exit: review failed};
\node[ex] at ($(pol.south west)+(0,-0.04)$) {Exit: returned};
 
\draw[rep, black!45, rounded corners=3pt] (sig.east) -- ++(0.15,0) |- ($(entry.west |- e4.west)+(0.08,0.02)$);
\draw[turn] (entry.east) -- (\laneR+0.55, \yd-1.125) -- (\laneR+0.55, \ya+0.75) -| (disc.north);
\node[hand] at (1.5*\dx, \ya+0.75) {Next discovery round};
\end{tikzpicture}}
\caption{The complete construction flow of \pipename{}, one box per step, with what the step produces under its title; a crossed eye marks what the step does not see (at the quality gate, the validity review). \textbf{Discovery:} a candidate enters through one of ten routes, is checked for novelty, is bound to verbatim quotations with answer atoms, and is screened for difficulty from a blind view without answer values; the top of each batch proceeds, the middle is evolved once and re-enters as a new candidate, and the bottom is kept as reserve. \textbf{Reference:} evidence fragments and a reference analysis are built, the reproducer rebuilds every answer atom from the evidence alone, a third worker compares the two atom by atom and attributes every difference, and the verified analysis is admitted to a problem family. \textbf{Question:} the question and its scoring contract are compiled from the verified analysis and frozen, pass the validity and consistency reviews and a final wording pass, and are accepted as a task package. \textbf{Evaluation:} the solver receives only the question, the allowed tools, and the cutoff; the judge rules item by item; failures, judge disagreements, and specification issues are recorded as signals for later discovery. Every worker result is checked by code before the next step uses it; a defect returns to the earliest affected step (dashed red), and red labels mark where candidates exit.}
\label{fig:pipeline_full}
\end{figure}

Figure~\ref{fig:pipeline_full} shows the complete construction flow; the
paragraphs below follow it step by step.
Before the next step uses a worker's result, code checks it against the
files the worker was given: every quotation must appear verbatim in its
source, every arithmetic expression must recompute to its stated result,
and every reference must resolve; a result that fails is redone.
Whether a result is good enough to hand on is decided by the
orchestrator.

\paragraph{Candidate discovery.}
The orchestrator sets the scope for a set of sources: which sources to
use, their boundaries, the parts to exclude, and the excerpts it has
already checked.
A worker then proposes one to three candidates along an assigned route.
Each candidate states the research question a professional would
actually ask, the decision it informs, the one judgment its answer turns
on, and every relationship needed to reach the conclusion.
Among the directions rejected at this step are those whose difficulty
comes only from stacking unrelated sub-questions, those that depend on
non-public data, and those that can end only in an unsupported causal
or predictive claim.

\paragraph{Novelty check.}
Code lists the closest delivered tasks and in-progress candidates by text
similarity; the orchestrator compares each of them with the candidate on
the central analysis object, the conclusion target, the necessary
relationships, and the deliverable, and decides whether the candidate is
a duplicate.
Changing only the company or the period while the analysis stays the
same does not make a candidate new.
Duplicates exit.

\paragraph{Integration.}
The orchestrator first confirms that the conclusion target has one
determinate reading: one measure, one period, one population, one unit.
A worker reopens the sources, binds every relationship to a verbatim
quotation and its location, and writes the smallest set of answer atoms
needed to evaluate the candidate.
Each atom has an expected value or category, keeps reported and
computed values apart, and names the quotations it rests on; its label
says what is measured, never what was found.
A candidate whose sources do not support a unique answer exits; one that
lacks a load-bearing public source waits until the source is supplied.

\paragraph{Difficulty screen.}
Code projects the candidate onto a blind view: a neutral objective, the
bound excerpts, and the form of each output (label, type, and unit),
with every answer value removed, and checks that no answer value appears
outside the excerpts.
From this view alone, a worker without tools estimates the shortest
professional path to the outputs and scores it from 0 to 4 on eight
structural dimensions: relation depth, cross-source reconciliation,
entity or population identity, temporal or version reconciliation,
conditional branching, professional judgment, state transitions, and
post-action verification.
It also raises two flags: a material ambiguity returns the candidate to
integration to fix the scope, and a task that can be bypassed as a
whole, with its conclusion stated outright in an excerpt or forced by
units or sign, is set aside as basic reserve, a pool of easier
candidates that are not developed further here.
The remaining candidates wait until a batch is full.
A second model then ranks the batch by comparing the candidates with one
another and gives a reason for each adjacent pair; the first and last
candidates of the previous batch serve as anchors that keep the batches
on one scale.
The top of the batch proceeds first to reference construction, and the bottom is kept as basic reserve.
The middle undergoes one round of difficulty evolution: a relationship
from the authorized sources that changes the conclusion is added, for
example a temporal or version reconciliation, a population
reconciliation, a hierarchy of source authority, or a conditional
branch, and the result passes again through the novelty check,
integration, and a later batch as a new candidate.
Both the scoring and the ranking criteria were refined using solver
results on constructed tasks.

\paragraph{Evidence and reference analysis.}
A worker reopens the sources and treats nothing from the candidate stage
as established.
It checks, binding by binding, that the quotation supports the stated
relationship and that entity, period, unit, basis, and population agree,
and it searches the sources for omissions that would change the answer,
a ranking, a sign, an attribution, or a classification.
That a quotation exists and that it supports the claim are separate
decisions.
The worker then extracts the evidence fragments, completes the domain's
semantic review (in finance, the reporting entity and the accounting
basis, among others), writes out every derivation step from the
fragments to the conclusion, and produces the reference analysis: a core
conclusion, answer atoms with tolerances, accepted alternatives, and the
conditions under which the conclusion holds.
Every atom rests on at least one evidence fragment and one derivation
step.
When it finds a defect, the worker names the earliest step that must
change: integration if a binding or the relationship itself is wrong,
or this step if only its own work must be redone.

\paragraph{Independent reproduction.}
The reproducer receives only the analytical objective, the evidence
fragments, and a neutral description of each required output; it does
not see the reference analysis, the derivation, the scoring contract, or
the question.
It derives every output from the evidence alone and records the
calculation or classification and the fragments it used.
An output that the evidence does not determine uniquely is reported as
a gap rather than answered.

\paragraph{Comparison.}
A third worker compares the reference analysis with the reproduction
atom by atom and records each atom as an exact match, a match within
tolerance, a difference, or not reproduced; tolerances are never
widened.
Every difference keeps both sides' content, the dimensions it touches
(entity, period, population, basis, unit, calculation, or conclusion),
and its effect (none, load-bearing, or unresolved).
Agreement is not proof: the worker reopens the evidence and checks the
premises the two results may share, namely the entity and period, the
source and its version, the load-bearing calculation or classification,
and the boundary of the inference.
Every load-bearing difference is attributed.
A defect in the task returns the candidate to the earliest affected
step, and only this attribution counts as a repair; an error of the
reproducer leads to a fresh reproduction; an execution failure reruns
only what failed; and a difference that cannot be attributed goes to
independent adjudication.
The two analyses are equivalent only when every atom is reproduced, no
load-bearing or unresolved difference remains, and the shared premises
hold.
The same conflict may be repaired at most twice; a candidate that still
does not converge exits.

\paragraph{Problem family.}
A worker selects a problem family that is already reusable, or defines a
narrow boundary for this analysis: its objective, positive boundary,
explicit exclusions, evidence roles, relationship shape, and output
contract.
It then checks the analysis against each of these six parts separately
and admits it only if all six pass.
A family defined from one instance cannot claim to be reusable or be
used to extend to other instances; reusability requires a separate
review with two distinct positive examples, one near-negative case, a
boundary draft, and an independent verification.
A candidate whose boundary cannot be settled exits.

\paragraph{Compilation and freezing.}
From the verified analysis, a worker writes the question the solver will
see: one high-level analytical objective, with the basis, period, and
information cutoff that make its answer unique.
The question does not list the hidden calculation steps, the order of
the sources, or intermediate conclusions, and it does not state hidden
answers.
The scoring contract is written with it.
Every requirement of the question maps to answer atoms; every atom is
either scored or explicitly unscored, and no load-bearing atom may go
unscored; every load-bearing atom has a stated fatal condition; every
accepted alternative of the reference analysis is marked as accepted or
as not applicable to this question; and helpful but unrequired
derivations are recorded as unscored.
Code checks that these links hold in both directions and that each
requirement is anchored in the exact wording of the question; once they
do, the question, the reference analysis, and the scoring contract are
frozen together as one version.
A second novelty check then compares the frozen question with the
delivered tasks, and a duplicate exits.

\paragraph{Quality gate.}
Two reviews run in sequence: the validity review, then the consistency
review.
The validity review sees only what the solver will see (the question,
the cutoff, and the allowed tools) and uses public sources to judge
whether the entity, period, unit, version, and cutoff are unambiguous;
whether the requested measures are compatible in definition, unit, and
basis; whether the question separates association, official findings,
and causal claims; whether every load-bearing input can be obtained with
the allowed tools before the cutoff; whether another supported
definition or version would give a materially different answer; and
whether the question asks for one high-level result.
The consistency review receives the frozen task and the source files and
independently re-derives every load-bearing number and conclusion.
It checks that the question and the reference analysis refer to the same
object, that every scored atom is supported by evidence, that reasonable
alternatives are accepted, that no item requires something the question
does not ask for, that the tolerances are reasonable, and that the
conclusion does not go beyond the evidence.
If either review fails, an adjudicator reads every review and confirms
each finding against the task; only confirmed defects are repaired.
A repair returns to compilation or to the evidence step, after which the
task is refrozen and reviewed again; a task may be repaired at most five
times before it exits.

\paragraph{Polish.}
A worker sees the question as the solver sees it, the full scoring
contract, and the allowed tools, but not the reference analysis, the
sources, or earlier reviews.
It checks that the question states the cutoff, what identifies each
source, the entity and scope, the deliverable, and the answer form and
boundary conditions that the scoring requires, and that it neither gives
away an answer nor recites the items.
It passes the question; edits its wording only, which leaves the scoring
contract and the reference analysis unchanged so that the gate's result
stands; or returns it for repair, for example when the problem lies in
the scoring contract.

\paragraph{Acceptance.}
Code confirms that the chain is complete: the reference analysis is
built, the comparison is equivalent, the problem family is admitted, and
the quality gate and the polish have passed, all for the current
version.
The task package is then delivered; solving and judging come after
delivery.

\paragraph{Solving, judging, and signals.}
The solver receives only the question, the allowed tools, and the
cutoff, and finds and reads the sources itself; the judge rules item by
item under the scoring contract (\S\ref{sec:setup},
Appendix~\ref{app:judging}).
Each evaluation records three kinds of signal: the items on which
solvers fail, the items on which judges disagree, and defects
that a judge finds in the scoring contract or the reference analysis.
The first two are kept as candidate directions for later discovery and
are verified against evidence before use; in the runs reported here,
they were not fed back automatically.
A defect that a judge finds returns the task to the earliest affected
step.

\paragraph{Returns and archiving.}
A defect returns the task to the earliest step it affects, and the steps
after it are redone on a new version; reviews, solver runs, and
judgments of an earlier version are not carried over.
An execution failure, such as a provider error, a timeout, or output
that does not match the required form, reruns only that step and does
not count as a repair.
Candidates exit for one of the following reasons, and every exit is kept
on record: duplicate, novelty undecided, unsupported by the sources,
basic reserve, not reproducible, family undecided, review failed, repair
limit reached, or cost above four times the batch median.

\begin{figure}[t]
\centering
\providecolor{brandclaude}{HTML}{D97757}
\providecolor{brandgpt}{HTML}{10A37F}
\providecolor{brandqwen}{HTML}{6F42C1}
\providecolor{branddeepseek}{HTML}{4D6BFE}
\resizebox{\textwidth}{!}{%
\begin{tikzpicture}[
  font=\sffamily,
  role/.style={rectangle, rounded corners=3pt, draw=#1!50, line width=0.6pt, fill=white, anchor=south,
               minimum width=1.95cm, minimum height=1.55cm, inner sep=0pt},
  badge/.style={rectangle, rounded corners=2.5pt, draw=none, fill=#1!13, minimum size=0.56cm, inner sep=0pt, anchor=north},
  rname/.style={anchor=north, inner sep=0pt, font=\sffamily\scriptsize\bfseries, text=black!80, align=center, text width=1.8cm},
  band/.style={rectangle, rounded corners=6pt, draw=none, fill=#1!7},
  lanelbl/.style={font=\sffamily\small\bfseries, text=#1, anchor=base},
  rowlbl/.style={anchor=west, inner sep=0pt, font=\sffamily\footnotesize, text=black!75},
  note/.style={anchor=north, inner sep=0pt, font=\sffamily\tiny, text=black!55},
  cell/.style={inner sep=0pt},
  legend/.style={anchor=north west, inner sep=0pt, font=\sffamily\scriptsize, text=black!60},
]
\def\cw{2.15}   %
\def\rh{0.78}   %
\def\xL{-4.6}   %
\def\yhdr{0.05} %
\newcommand{\ic}[2]{\vadeicon{#1}{#2}}
\newcommand{\sees}{\ic{eye-ink}{9.5pt}}
\newcommand{\hid}{\ic{eye-off-red}{9.5pt}}
\newcommand{\web}{\ic{globe-ink}{9.5pt}}
\newcommand{\writes}{\ic{pencil-ink}{9.5pt}}
\newcommand{\rolecard}[5]{%
  \node[role=#1] (#2) at (#3,\yhdr) {};
  \node[badge=#1] at ($(#2.north)+(0,-0.14)$) {\ic{#4}{13pt}};
  \node[rname] at ($(#2.north)+(0,-0.8)$) {#5};}
\newcommand{\rowl}[3]{%
  \node[anchor=west, inner sep=0pt] at (\xL, -#1*\rh) {\ic{#2-gray}{9.5pt}};
  \node[rowlbl] at (\xL+0.42, -#1*\rh) {#3};}
\newcommand{\cellat}[3]{\node[cell] at (#1*\cw, -#2*\rh) {#3};}
\newcommand{\cellnote}[4]{\node[cell] at (#1*\cw, -#2*\rh+0.08) {#3};\node[note] at (#1*\cw, -#2*\rh-0.1) {#4};}

\fill[band=brandclaude]   (0*\cw-1.02, \yhdr+1.85) rectangle (0*\cw+1.02, -7*\rh-0.4);
\fill[band=brandgpt]      (1*\cw-1.02, \yhdr+1.85) rectangle (2*\cw+1.02, -7*\rh-0.4);
\fill[band=brandqwen]     (3*\cw-1.02, \yhdr+1.85) rectangle (4*\cw+1.02, -7*\rh-0.4);
\fill[band=branddeepseek] (5*\cw-1.02, \yhdr+1.85) rectangle (6*\cw+1.02, -7*\rh-0.4);
\draw[brandclaude!60,   line width=2pt, line cap=round] (0*\cw-0.85, \yhdr+1.95) -- (0*\cw+0.85, \yhdr+1.95);
\draw[brandgpt!60,      line width=2pt, line cap=round] (1*\cw-0.85, \yhdr+1.95) -- (2*\cw+0.85, \yhdr+1.95);
\draw[brandqwen!60,     line width=2pt, line cap=round] (3*\cw-0.85, \yhdr+1.95) -- (4*\cw+0.85, \yhdr+1.95);
\draw[branddeepseek!60, line width=2pt, line cap=round] (5*\cw-0.85, \yhdr+1.95) -- (6*\cw+0.85, \yhdr+1.95);
\node[lanelbl=brandclaude]   at (0*\cw,   \yhdr+2.15) {Discovery};
\node[lanelbl=brandgpt]      at (1.5*\cw, \yhdr+2.15) {Reference};
\node[lanelbl=brandqwen]     at (3.5*\cw, \yhdr+2.15) {Question};
\node[lanelbl=branddeepseek] at (5.5*\cw, \yhdr+2.15) {Evaluation};

\rolecard{brandclaude}{card}{0*\cw}{gauge-claude}{Difficulty card}
\rolecard{brandgpt}{bld}{1*\cw}{file-check-gpt}{Evidence \& reference builder}
\rolecard{brandgpt}{rpr}{2*\cw}{calculator-gpt}{Reproducer}
\rolecard{brandqwen}{val}{3*\cw}{shield-check-qwen}{Validity reviewer}
\rolecard{brandqwen}{con}{4*\cw}{clipboard-check-qwen}{Consistency reviewer}
\rolecard{branddeepseek}{sol}{5*\cw}{bot-deepseek}{Solver}
\rolecard{branddeepseek}{jud}{6*\cw}{gavel-deepseek}{Judge}

\foreach \r in {1,...,6} \draw[black!10, line width=0.4pt] (\xL, -\r*\rh-\rh/2) -- (6*\cw+1.02, -\r*\rh-\rh/2);

\rowl{1}{file-text}{Source documents}
\cellnote{0}{1}{\sees}{Bound excerpts} \cellat{1}{1}{\sees} \cellat{2}{1}{\hid} \cellnote{3}{1}{\web}{Public web} \cellnote{4}{1}{\sees}{Files and web} \cellnote{5}{1}{\web}{Must locate} \cellnote{6}{1}{\sees}{Solver's record}
\rowl{2}{quote}{Evidence fragments}
\cellat{0}{2}{\hid} \cellat{1}{2}{\writes} \cellat{2}{2}{\sees} \cellat{3}{2}{\hid} \cellat{4}{2}{\sees} \cellat{5}{2}{\hid} \cellat{6}{2}{\sees}
\rowl{3}{target}{Reference atoms}
\cellat{0}{3}{\hid} \cellat{1}{3}{\writes} \cellat{2}{3}{\hid} \cellat{3}{3}{\hid} \cellat{4}{3}{\sees} \cellat{5}{3}{\hid} \cellnote{6}{3}{\sees}{After the audit}
\rowl{4}{workflow}{Derivation}
\cellat{0}{4}{\hid} \cellat{1}{4}{\writes} \cellat{2}{4}{\hid} \cellat{3}{4}{\hid} \cellat{4}{4}{\sees} \cellat{5}{4}{\hid} \cellnote{6}{4}{\sees}{After the audit}
\rowl{5}{clipboard-list}{Question}
\cellnote{0}{5}{\sees}{Objective only} \cellnote{1}{5}{\sees}{Objective only} \cellnote{2}{5}{\sees}{Objective only} \cellat{3}{5}{\sees} \cellat{4}{5}{\sees} \cellat{5}{5}{\sees} \cellat{6}{5}{\sees}
\rowl{6}{list-checks}{Scoring contract}
\cellat{0}{6}{\hid} \cellat{1}{6}{\hid} \cellat{2}{6}{\hid} \cellat{3}{6}{\hid} \cellat{4}{6}{\sees} \cellat{5}{6}{\hid} \cellnote{6}{6}{\sees}{After the audit}
\rowl{7}{route}{Solver's trajectory}
\cellat{0}{7}{\hid} \cellat{1}{7}{\hid} \cellat{2}{7}{\hid} \cellat{3}{7}{\hid} \cellat{4}{7}{\hid} \cellnote{5}{7}{\sees}{Its own} \cellat{6}{7}{\sees}

\node[legend] at (\xL, -7*\rh-0.7) {%
  \raisebox{-2pt}{\sees}\ Sees\qquad \raisebox{-2pt}{\writes}\ Writes\qquad \raisebox{-2pt}{\web}\ Reaches it only through public search and retrieval tools\qquad \raisebox{-2pt}{\hid}\ Never sees};
\end{tikzpicture}}
\caption{What each isolated role of Figure~\ref{fig:pipeline_full} is allowed to see. The difficulty card scores the relationship from the bound excerpts, without answer values and before any evidence fragment is built; the reproducer rebuilds the atoms from the evidence fragments alone; the validity reviewer and the solver work from the request and the public web; the consistency reviewer sees the whole package and the bound source files; the judge sees the solver's record and evidence first, and the reference and scoring contract only after it has filed a preliminary audit. The question exists only after the reference analysis is verified, so the earlier roles see just the analytical objective.}
\label{fig:visibility}
\end{figure}
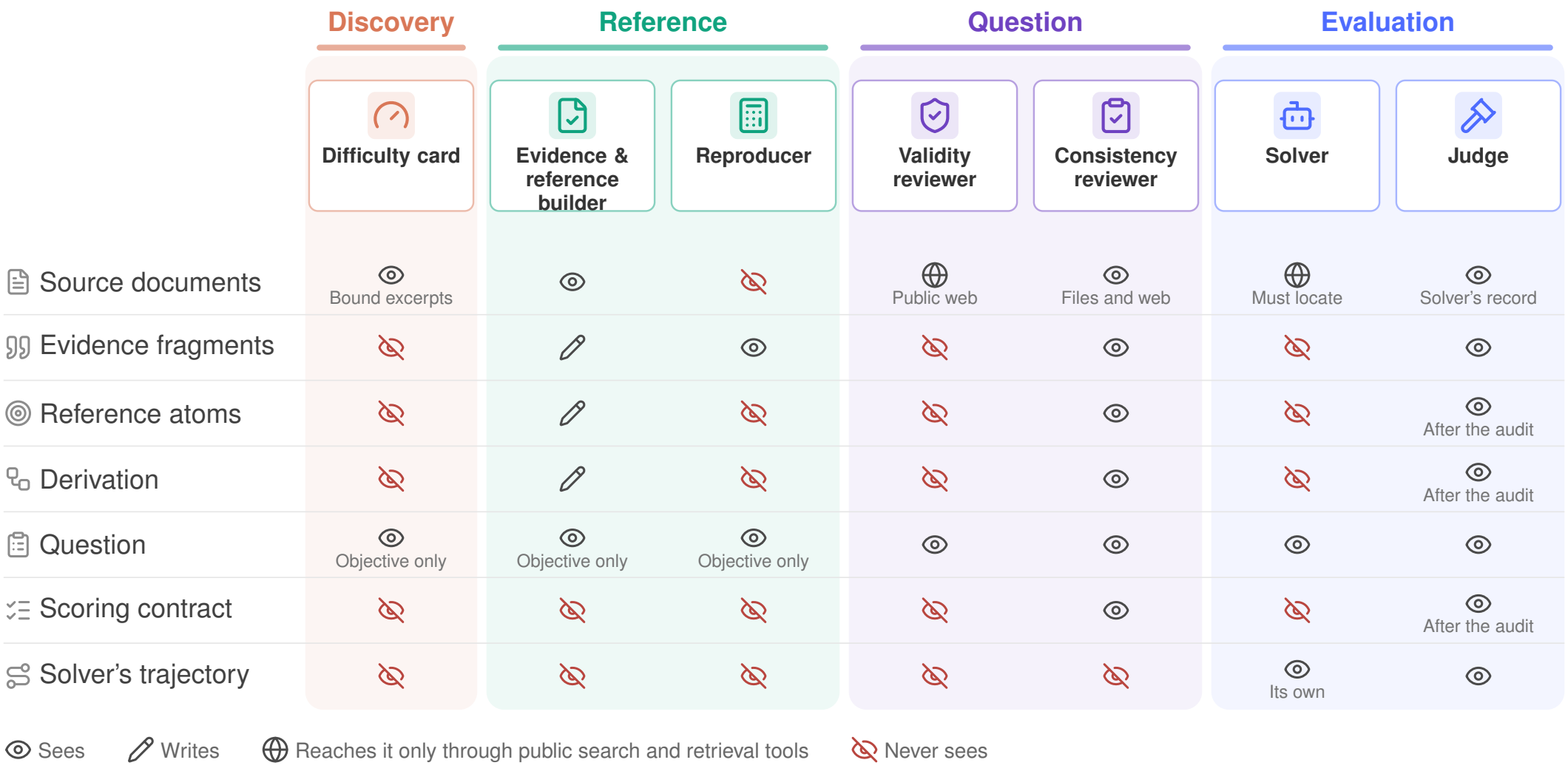

\paragraph{What each role sees.}
Figure~\ref{fig:visibility} shows what each isolated role is allowed to
see.
These are the roles whose separation the construction relies on: the
difficulty worker judges a candidate without its answer values, the
reproducer rebuilds the answer without the reference analysis, the
validity review judges the question from the solver's position, and the
solver never sees the reference analysis or the scoring contract.
The other roles are not separated in this way.
The orchestrator reads every result but writes nothing into the task
package; the comparison worker sees both analyses and the evidence; and
the polish worker sees the question, the scoring contract, and the
allowed tools.

\subsection{Discovery Routes}
\label{app:routes}

A route tells the discovery worker where to start looking; it is a
starting preference and does not guarantee a candidate.
The worker tries the assigned route first and may switch to another
route before a candidate forms only when the sources require it; once a
candidate forms, the route it actually came from is recorded with it.
The two failure-first routes use what is known about model failures to
choose where to look, but every relationship they propose must be closed
from the sources like that of any other candidate.
Table~\ref{tab:routes} lists the ten routes.

\begin{table}[h]
\caption{The ten discovery routes, grouped by where they start.}
\label{tab:routes}
\centering\footnotesize
\begin{tabular}{@{}lp{0.62\textwidth}@{}}
\toprule
Route & Starting point and rule \\
\midrule
\multicolumn{2}{@{}l}{\emph{Source-first}} \\
Raw source relationships & Extract analytical relationships directly from primary sources. \\
Source-supported extension & Use a verified simple relationship as a search scope; keep only external extensions that add a load-bearing relation. \\
\multicolumn{2}{@{}l}{\emph{Analysis-first}} \\
Complete-analysis reverse authoring & Start from a public analysis with conclusion, method, facts, and sources; verify against primary sources. \\
Verified-relationship reverse authoring & Build a higher-level task on a verified data or evidence relationship. \\
Real problem, then answer & Start from a real professional objective; compile no question until evidence and reference analysis close. \\
\multicolumn{2}{@{}l}{\emph{Family-first}} \\
Mature-family expansion & Add a new entity, period, event, or state to a validated narrow family. \\
Necessary-relation accumulation & Create a child task only when every inherited relation stays necessary and new evidence adds a conclusion-sensitive dependency. \\
Taxonomy-gap discovery & Prioritize underrepresented categories without relabeling ordinary tasks to fill quotas. \\
\multicolumn{2}{@{}l}{\emph{Failure-first}} \\
Historical error-structure transfer & Use known model failure structures to find new source-grounded instances, without copying old questions. \\
Model-boundary hypothesis & A frontier model names relations missing from its own prospective solution path; every hypothesis must be closed from sources. \\
\bottomrule
\end{tabular}
\end{table}

\subsection{Running Example: Public Storage Margin Analysis}
\label{app:ps}

\begin{figure}[h]
\centering
\providecolor{brandclaude}{HTML}{D97757}
\providecolor{brandgpt}{HTML}{10A37F}
\providecolor{brandqwen}{HTML}{6F42C1}
\providecolor{branddeepseek}{HTML}{4D6BFE}
\providecolor{feedred}{HTML}{B9463F}
\resizebox{\textwidth}{!}{%
\begin{tikzpicture}[
  font=\sffamily,
  card/.style={rectangle, rounded corners=3pt, draw=#1!50, line width=0.6pt, fill=white, anchor=north west,
               minimum width=\W cm, minimum height=\H cm, inner sep=0pt},
  badge/.style={rectangle, rounded corners=2.5pt, draw=none, fill=#1!13, minimum size=0.56cm, inner sep=0pt, anchor=north west},
  title/.style={anchor=north west, inner sep=0pt, font=\sffamily\small\bfseries, text=black!85},
  body/.style={anchor=north west, inner sep=0pt, text width=\TW cm, align=flush left, font=\sffamily\scriptsize, text=black!55},
  chip/.style={rectangle, rounded corners=3pt, inner sep=2pt, font=\sffamily\tiny, fill=brandgpt!12, text=brandgpt!80!black},
  band/.style={rectangle, rounded corners=6pt, draw=none, fill=#1!7},
  lanelbl/.style={font=\sffamily\small\bfseries, text=#1, anchor=base},
  flow/.style={-{Stealth[length=2mm,width=1.6mm]}, line width=0.7pt, black!45},
  flowlbl/.style={font=\sffamily\tiny\itshape, text=black!55, fill=white, inner sep=1.5pt},
]
\def\W{3.9}\def\H{4.55}\def\TW{3.4}\def\dx{5.1}
\newcommand{\ic}[2]{\raisebox{-2pt}{\vadeicon{#1}{#2}}}
\newcommand{\cardat}[5]{%
  \node[card=#1] (#2) at (#3,0) {};
  \node[badge=#1] at ($(#2.north west)+(0.15,-0.15)$) {\vadeicon{#4}{13pt}};
  \node[title] at ($(#2.north west)+(0.85,-0.2)$) {#5};}

\fill[band=brandclaude] (-0.3, 0.35) rectangle (\W+0.3, -\H-0.3);
\fill[band=brandgpt]    (\dx-0.3, 0.35) rectangle (2*\dx+\W+0.3, -\H-0.3);
\fill[band=brandqwen]   (3*\dx-0.3, 0.35) rectangle (3*\dx+\W+0.3, -\H-0.3);
\draw[brandclaude!60, line width=2pt, line cap=round] (-0.1, 0.55) -- (\W+0.1, 0.55);
\draw[brandgpt!60,    line width=2pt, line cap=round] (\dx-0.1, 0.55) -- (2*\dx+\W+0.1, 0.55);
\draw[brandqwen!60,   line width=2pt, line cap=round] (3*\dx-0.1, 0.55) -- (3*\dx+\W+0.1, 0.55);
\node[lanelbl=brandclaude] at (\W/2, 0.75) {Discovery};
\node[lanelbl=brandgpt]    at (1.5*\dx+\W/2, 0.75) {Reference};
\node[lanelbl=brandqwen]   at (3*\dx+\W/2, 0.75) {Question};

\cardat{brandclaude}{c1}{0}{table-claude}{Source}
\node[body] at ($(c1.north west)+(0.25,-0.85)$) {%
  \textbf{Public Storage}, Q2 2026 earnings supplement, same-store operating table\\[5pt]
  {\setlength{\tabcolsep}{3pt}\renewcommand{\arraystretch}{1.1}%
  \begin{tabular}{@{}lrr@{}}
    \$M & Q2 2026 & Q2 2025\\
    Revenue & 1{,}006.5 & 1{,}012.4\\
    Costs & 260.2 & 249.1\\
    Margin & 74.2\% & 75.4\%\\
  \end{tabular}}\\[6pt]
  \ic{calendar-gray}{8pt}\ Cutoff: 29 July 2026};

\cardat{brandgpt}{c2}{\dx}{file-check-gpt}{Evidence \& reference}
\node[body] at ($(c2.north west)+(0.25,-0.85)$) {%
  Four evidence columns, each cell bound to the source\\[6pt]
  \textbf{Nine answer atoms}, among them (pp)\\[4pt]
  \tikz[baseline=-0.5ex]\node[chip]{Q2 $-$1.2};\;
  \tikz[baseline=-0.5ex]\node[chip]{H1 $-$0.5};\;
  \tikz[baseline=-0.5ex]\node[chip]{Gap $-$0.7};\\[4pt]
  \tikz[baseline=-0.5ex]\node[chip]{Gap from amounts $-$0.76};\\[4pt]
  \tikz[baseline=-0.5ex]\node[chip]{Direct $-$0.70};\;
  \tikz[baseline=-0.5ex]\node[chip]{Indirect $-$0.06};\\[6pt]
  Conditions: same-store population; disclosure rounding};

\cardat{brandgpt}{c3}{2*\dx}{calculator-gpt}{Reproduction}
\node[body] at ($(c3.north west)+(0.25,-0.85)$) {%
  Sees the objective and the evidence fragments\\[3pt]
  {\color{feedred!85}\ic{eye-off-red}{8pt}\ The reference analysis}\\[7pt]
  \textbf{Round 1}\\
  Attribution atom missing\\
  {\color{feedred!85}\ic{rotate-ccw-red}{8pt}\ Load-bearing difference}\\[5pt]
  \textbf{Round 2}\\
  {\color{brandgpt!80!black}\ic{circle-check-gpt}{8pt}\ All nine atoms equivalent}};

\cardat{brandqwen}{c4}{3*\dx}{snowflake-qwen}{Frozen package}
\node[body] at ($(c4.north west)+(0.25,-0.85)$) {%
  \textbf{Question}\\
  Q2 versus H1 margin movement, with a bounded cost-to-revenue attribution\\[6pt]
  \textbf{Scoring contract}\\
  Numeric items with tolerances\\
  Qualifiers: population, rounding\\
  {\color{feedred!85}Fatal: causal claims}\\[6pt]
  \ic{lock-gray}{8pt}\ Frozen, version-locked};

\draw[flow] (c1.east) -- node[flowlbl,above]{Build} (c2.west);
\draw[flow] (c2.east) -- node[flowlbl,above]{Evidence only} (c3.west);
\draw[flow] (c3.east) -- node[flowlbl,above]{Compile} (c4.west);
\end{tikzpicture}}
\caption{The PS-MARGIN running example through the first three rows of Figure~\ref{fig:pipeline_full}. One earnings table becomes four bound evidence columns and a reference analysis of nine answer atoms, five of which are shown; the one-decimal gap of $-$0.7 points and the gap of $-$0.76 points recomputed from the thousand-dollar amounts are both part of it. The reproducer, which never sees the reference analysis, missed the attribution atom in its first reproduction and reproduced all nine atoms in the next; the frozen package exposes only the question and the tools to the solver.}
\label{fig:example}
\end{figure}
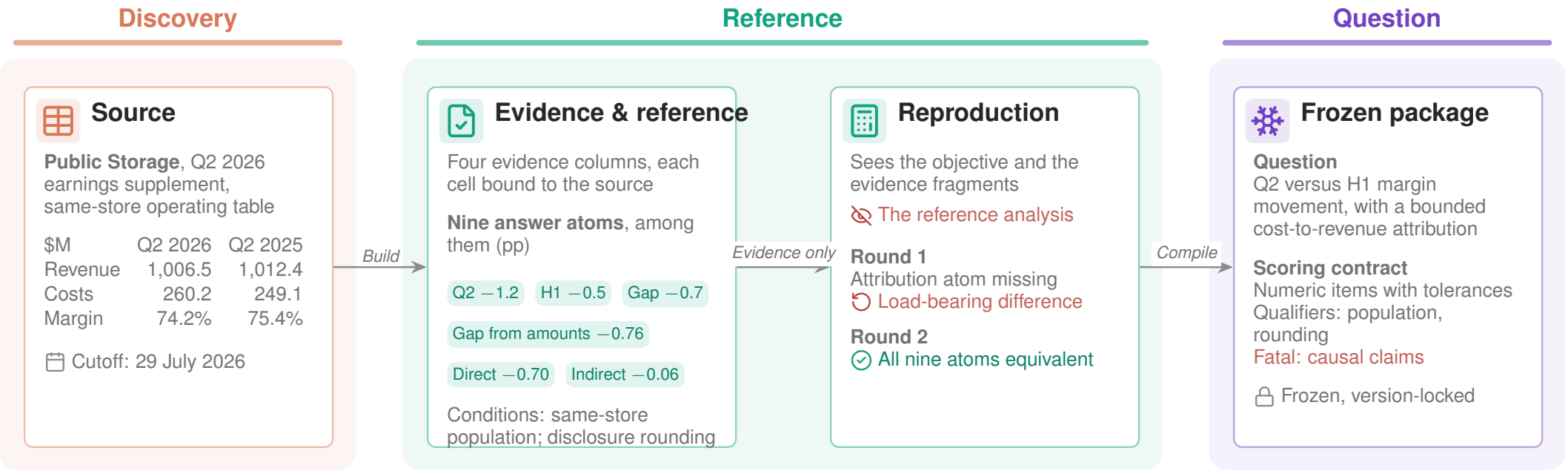

Figure~\ref{fig:example} follows one task, PS-MARGIN, through the first
three rows of Figure~\ref{fig:pipeline_full}.
The task is not part of the benchmark described in
\S\ref{sec:task-composition}; we use it only to illustrate the method.

\paragraph{Question.}
``Using Public Storage's public disclosures available through
July~29, 2026, assess whether its year-over-year Same Store Facilities
gross margin after indirect costs movement in Q2~2026 differed from
the movement for the six months ended June~30, 2026, and quantify the
difference.
Confine the analysis to the disclosed Same Store Facilities population
and exclude separately reported non-same-store facility populations.
Support the judgment by reconciling the disclosed revenue and direct
and indirect operating costs to reported same-store NOI and by
comparing the revenue and cost relationships for both periods.
Explain the differing margin movements with a bounded cost-to-revenue
arithmetic attribution, preserving the disclosure's rounding limits
and avoiding causal claims.''

\paragraph{Source data.}
The evidence comes from the Same Store Operating Performance table of
Public Storage's Q2~2026 earnings supplement, which covers 2{,}755
facilities and 192.1M net rentable square feet, about 83\% of the U.S.\
consolidated portfolio (Table~\ref{tab:ps-data}).
All four columns refer to the same facility population.

\begin{table}[h]
\caption{PS-MARGIN source data (thousands of USD).
Indirect costs shown as positive amounts; the source presents them as
deductions from Direct NOI.}
\label{tab:ps-data}
\centering
\footnotesize
\begin{tabular}{@{}lrrrr@{}}
\toprule
\textbf{Item} & \textbf{Q2 2026} & \textbf{Q2 2025} & \textbf{H1 2026} & \textbf{H1 2025} \\
\midrule
Total revenue        & 1{,}006{,}549 & 1{,}012{,}439 & 2{,}007{,}382 & 2{,}013{,}460 \\
Property taxes       &   111{,}362 &   105{,}186 &   217{,}485 &   212{,}411 \\
On-site managers     &    33{,}713 &    34{,}325 &    68{,}075 &    68{,}907 \\
Repairs              &    20{,}824 &    20{,}172 &    43{,}431 &    44{,}182 \\
Utilities            &    11{,}888 &    11{,}506 &    27{,}344 &    27{,}469 \\
Marketing            &    22{,}029 &    20{,}732 &    44{,}625 &    44{,}382 \\
Other direct         &    27{,}855 &    26{,}437 &    55{,}999 &    53{,}946 \\
\cmidrule(lr){2-5}
Direct costs         &   227{,}671 &   218{,}358 &   456{,}959 &   451{,}297 \\
Indirect costs       &    32{,}505 &    30{,}748 &    64{,}650 &    62{,}133 \\
NOI                  &   746{,}373 &   763{,}333 & 1{,}485{,}773 & 1{,}500{,}030 \\
\midrule
\textbf{Disclosed margin}     &    \textbf{74.2\%} &    \textbf{75.4\%} &    \textbf{74.0\%} &    \textbf{74.5\%} \\
\bottomrule
\end{tabular}
\end{table}

\paragraph{Reference analysis.}
The reference analysis has nine answer atoms: the same-store scope; the
reported changes in revenue and in direct and indirect costs for the
quarter and for the six months; the reconciliation of the six direct-cost
components, and of revenue less costs to NOI, in all four columns; the
quarterly and the six-month margin movements; their difference; and a
bounded arithmetic attribution of that difference.
On the issuer's one-decimal margins, the margin fell 1.2 points in Q2
and 0.5 points over the six months, so the Q2 movement was 0.7 points
worse.
Recomputed from the thousand-dollar amounts, the additional Q2 erosion
is about 0.76 points, of which about 0.70 comes from the direct-cost
burden and about 0.06 from the indirect-cost burden.
The scoring contract accepts the amount-based figure when it is labeled
as a cross-check, and either rounding of the two burdens as long as they
add up to the amount-based gap.
The six-month period includes Q2, so the two periods are not
independent, and the attribution is arithmetic: an answer that states
the decline in percent rather than percentage points, or that explains
it by operational causes, fails.

\paragraph{Independent reproduction.}
The reproducer received only the analytical objective, the evidence
fragments, and a neutral description of each output.
Its first reproduction recovered the margins and their difference but
omitted the bounded attribution, which the comparison recorded as a
load-bearing difference.
In the next reproduction all nine atoms were reproduced, and the
comparison, after reopening the evidence to check the premises the two
analyses share, found them equivalent.

\subsection{Scoring Contract Example: Travelers Underwriting Decomposition}
\label{app:scoring}

This section reproduces the complete scoring contract of the Travelers
case study (\S\ref{case:travelers}), an Adjustments task.
The question reads:
``Using only public information available through July 17, 2026, assess
how much of The Travelers Companies, Inc.'s consolidated underwriting
improvement for the three months ended June 30, 2026 versus June 30,
2025 reflects improved current-accident-year performance rather than
changes in catastrophe losses and favorable prior-year reserve
development.
Support your judgment by presenting the signed reported-to-underlying
combined-ratio bridge for both quarters, decomposing the year-over-year
reported improvement and quantifying the share retained in the
underlying movement.
Use the consolidated three-month pre-tax underwriting-income
reconciliation as a non-interchangeable cross-check, likewise showing
each quarter's signed bridge, the year-over-year decomposition, and the
retained share.
Conclude with a bounded historical classification of the underlying
direction, retained share, and dominant driver; do not infer
persistence, causality, reserve adequacy, solvency, or future
performance.''

Table~\ref{tab:scoring-map} maps each of the question's four
requirements to the answer atoms it makes scorable and lists the five
accepted alternatives; Table~\ref{tab:scoring-atoms} gives the nine
atoms, and Table~\ref{tab:scoring-fatal} the eight fatal conditions.
All atoms are fatal, so omitting or materially misstating any one fails
the answer, and no atom is left unscored.
In the tables, PYD stands for favorable prior-year reserve development,
and condition IDs are quoted from the contract, where the prefix
\texttt{fec} marks a fatal error condition.

\begin{table}[h]
\caption{Travelers question requirements and the answer atoms each makes scorable (top), and the
accepted alternatives with the atoms they apply to (bottom). Each requirement is anchored in the
question's own wording; all five alternatives are accepted for scoring.}
\label{tab:scoring-map}
\centering
\footnotesize
\setlength{\tabcolsep}{3.5pt}
\begin{tabular}{@{}>{\raggedright\arraybackslash}p{5.6cm}>{\raggedright\arraybackslash}p{7.0cm}@{}}
\toprule
Requirement (anchor in the question) & Answer atoms \\
\midrule
Assessment: ``assess how much of \ldots\ consolidated underwriting improvement \ldots\ reflects improved current-accident-year performance''
  & earnings\_quality\_judgment \\
Ratio support: ``presenting the signed reported-to-underlying combined-ratio bridge for both quarters \ldots''
  & ratio\_bridge\_2026, ratio\_bridge\_2025, ratio\_change\_decomp., ratio\_retained\_share \\
Pre-tax cross-check: ``use the consolidated three-month pre-tax underwriting-income reconciliation as a non-interchangeable cross-check \ldots''
  & pretax\_bridge\_2026, pretax\_bridge\_2025, pretax\_change\_decomp., pretax\_retained\_share \\
Conclusion boundary: ``conclude with a bounded historical classification \ldots; do not infer persistence, causality \ldots''
  & earnings\_quality\_judgment \\
\midrule
Accepted alternative & Applies to \\
\midrule
Equivalent signed forms of the bridges, e.g.\ $83.6 + 5.4 - 4.9 = 84.1$
  & ratio\_bridge\_2026, ratio\_bridge\_2025 \\
Retained ratio share as about 9\%, or 8.96\% when marked as calculation precision
  & ratio\_retained\_share \\
Pre-tax retained share as about 6.1\% or approximately 6\%
  & pretax\_retained\_share \\
``Limited'', ``minority'', or ``modest'' in place of ``small'', with the 9.0\% and 6.1\% calculations
  & earnings\_quality\_judgment \\
Catastrophe and favorable prior-year development supplying about 91\% of the ratio improvement and 94\% of the pre-tax improvement, with bases kept apart
  & both decompositions, both retained shares, earnings\_quality\_judgment \\
\bottomrule
\end{tabular}
\end{table}

\begin{table}[h]
\caption{Travelers answer atoms; each is fatal if missing or materially incorrect.}
\label{tab:scoring-atoms}
\centering
\footnotesize
\setlength{\tabcolsep}{3.5pt}
\begin{tabular}{@{}p{3.4cm}p{7.2cm}cc@{}}
\toprule
Atom ID & Expected display & Tol.\ & Unit \\
\midrule
ratio\_bridge\_2026
  & 83.6\% $-$ ($-$5.4\,pp) $-$ 4.9\,pp $=$ 84.1\%
  & $\pm$0.1 & \% \\
ratio\_bridge\_2025
  & 90.3\% $-$ ($-$2.9\,pp) $-$ 8.5\,pp $=$ 84.7\%
  & $\pm$0.1 & \% \\
ratio\_change\_decomp.
  & 6.7\,pp improvement: 3.6\,pp catastrophe, 2.5\,pp PYD, 0.6\,pp underlying
  & --- & pp \\
ratio\_retained\_share
  & $\approx$9.0\% (0.6 of 6.7\,pp)
  & $\pm$0.2 & \% \\
pretax\_bridge\_2026
  & \$1{,}738M $-$ \$578M $+$ \$518M $=$ \$1{,}678M
  & $\pm$1 & \$M \\
pretax\_bridge\_2025
  & \$1{,}022M $-$ \$315M $+$ \$927M $=$ \$1{,}634M
  & $\pm$1 & \$M \\
pretax\_change\_decomp.
  & \$716M improvement: \$409M catastrophe, \$263M PYD, \$44M underlying
  & --- & \$M \\
pretax\_retained\_share
  & $\approx$6.1\% (\$44M of \$716M)
  & $\pm$0.1 & \% \\
earnings\_quality\_judgment
  & Underlying improved; retained share small; reported improvement predominantly catastrophe-and-PYD-driven.
  & --- & --- \\
\bottomrule
\end{tabular}
\end{table}

\begin{table}[h]
\caption{Fatal error conditions for the Travelers scoring contract.
Any triggered condition produces a \textbf{Fail} verdict regardless of
other atom results.}
\label{tab:scoring-fatal}
\centering
\footnotesize
\setlength{\tabcolsep}{3.5pt}
\begin{tabular}{@{}p{3.6cm}p{8.8cm}@{}}
\toprule
Condition ID & Description \\
\midrule
fec\_ratio\_bridge
  & Either quarterly combined-ratio bridge is omitted, uses an incorrect
    component or sign, or does not reproduce the underlying ratio. \\[3pt]
fec\_ratio\_decomposition
  & Reported combined-ratio change is not correctly decomposed into
    catastrophe, PYD, and underlying components, or the retained share
    is omitted or materially incorrect. \\[3pt]
fec\_pretax\_bridge
  & Either quarterly pre-tax underwriting-income bridge is omitted,
    uses an incorrect adjustment sign or amount, or fails to reproduce
    underlying pre-tax underwriting income. \\[3pt]
fec\_pretax\_decomposition
  & Reported pre-tax underwriting-income change is not correctly
    decomposed into catastrophe, PYD, and underlying components, or the
    retained share is omitted or materially incorrect. \\[3pt]
fec\_quality\_judgment
  & Answer omits the three-part quality classification or materially misclassifies the underlying direction, retained share, or dominant driver. \\[3pt]
fec\_scope\_population
  & Analysis uses a segment, year-to-date, audited-status assumption,
    later disclosure, or other population inconsistent with consolidated
    three-month Q2\,2026 vs.\ Q2\,2025 disclosures. \\[3pt]
fec\_measurement\_basis
  & The ratio and pre-tax-dollar retained shares are treated as
    interchangeable or combined without preserving their measurement bases. \\[3pt]
fec\_historical\_boundary
  & Conclusion asserts persistence, causality, reserve adequacy or
    redundancy, solvency, or future performance beyond the historical
    evidence. \\
\bottomrule
\end{tabular}
\end{table}

\subsection{Source Files}
\label{app:source-files}

Table~\ref{tab:source-channels} breaks down the source files of the
finance and legal runs (\S\ref{sec:sources}) by the channel from which
they were collected.

\begin{table}[h]
\caption{Source files of the finance and legal runs by the channel
from which they were collected.}
\label{tab:source-channels}
\centering\scriptsize
\setlength{\tabcolsep}{3pt}
\begin{tabular}{@{}p{3.6cm}rr@{\hspace{1.2em}}p{3.9cm}rr@{}}
\toprule
\multicolumn{3}{@{}l}{Finance (\nfiles{} files)} & \multicolumn{3}{l}{Law (\nlawfiles{} files)} \\
\cmidrule(r){1-3}\cmidrule(l){4-6}
Channel & Files & Share & Channel & Files & Share \\
\midrule
SEC & 75{,}892 & 78.6\% & SEC & 27{,}538 & 51.5\% \\
FDIC and its Office of Inspector General & 5{,}114 & 5.3\% & GovInfo, eCFR, and the U.S.\ Code & 9{,}249 & 17.3\% \\
Federal Reserve & 5{,}057 & 5.2\% & Federal Register & 4{,}022 & 7.5\% \\
Company websites and investor-relations channels & 996 & 1.0\% & CPSC and SaferProducts & 2{,}379 & 4.4\% \\
OCC & 773 & 0.8\% & CourtListener / RECAP & 1{,}644 & 3.1\% \\
CFPB & 392 & 0.4\% & Supreme Court and federal court websites & 1{,}561 & 2.9\% \\
FFIEC & 257 & 0.3\% & Regulations.gov & 1{,}400 & 2.6\% \\
CMS & 206 & 0.2\% & State environmental regulators & 1{,}106 & 2.1\% \\
FHFA & 140 & 0.1\% & California Attorney General and privacy agencies & 1{,}070 & 2.0\% \\
DOJ & 126 & 0.1\% & DOL and OSHA & 728 & 1.4\% \\
FinCEN & 99 & 0.1\% & NLRB & 567 & 1.1\% \\
Other & 7{,}472 & 7.7\% & Other & 2{,}234 & 4.2\% \\
\bottomrule
\end{tabular}
\end{table}

In energy, the sources are the primary records of the bodies that
regulate and operate the U.S. energy sector: event reports from the
nuclear regulator (NRC); orders and filings from the federal energy
regulator (FERC) and the state public utility commissions; statistics
from the Energy Information Administration (EIA); publications of the
grid reliability organization (NERC); and market data and reports from
the regional grid operators (PJM, SPP, CAISO, ERCOT, ISO-NE) and their
market monitors, together with federal regulations and the Federal
Register.
In medicine, the sources are FDA review documents, labels and their
revision histories, approval letters, and safety communications;
ClinicalTrials.gov records and their version histories; and PubMed.

\subsection{Prompts of Key Steps}
\label{app:prompts}

{  %
Each step of the engine has its own prompt; the question and the
reference analysis are each built across several steps.
We reproduce in full the prompts of two steps central to verification:
independent reproduction and the validity review.
The prompts of the other steps are several times longer; what each of
them requires is described step by step in
Appendix~\ref{app:method}, and the judging rules are in
Appendix~\ref{app:judging}.
Role and step names follow the terms of this paper; file and field
identifiers are quoted as they are.

\paragraph{Independent reproduction (full prompt).}
\begin{quote}\small
Reproduce the requested analysis from the supplied analytical objective
and verified evidence fragments. You must not seek or infer the hidden
reference analysis. The packet deliberately excludes the reference
analysis, its derivation, the scoring contract, the drafted question,
prior model attempts, and the difficulty result. You see only
\texttt{reproduction\_input.json}: the analytical objective, the
verified evidence fragments, and the required outputs under opaque
identifiers (\texttt{output-001}, \ldots) with a neutral description,
value type and unit each; and the domain profile under
\texttt{policy/} for the domain's units, sign and precision
conventions.

Attempt each required output and report it under its opaque identifier
as \texttt{atom\_id}. Show the calculation or classification used and
bind every answer to the evidence fragment IDs actually used. If the
evidence does not uniquely support an output, report the gap and use
\texttt{blocked} rather than inventing an answer. Preserve entity,
date, unit, population, scope, version, rounding, and disclosure
limitations.

SOURCES\_USED BARE IDENTIFIER RULE. The \texttt{sources\_used} array
must contain bare source identifiers only. NEGATIVE EXAMPLE:
``src-1 - archived annual report capture''. The required form is
``src-1'', with no dash, description, or annotation of any kind
appended.

NO UNSUPPORTED AFFIRMATIVE INFERENCE RULE. State only what the bound
evidence establishes. NEGATIVE EXAMPLE: writing ``the supplied context
identifies the bank as a Category III institution'' when the bound
fragment only lists the institutions that are NOT Category III and the
bank is absent from it; absence from a list of exceptions does not
identify a category. Converting a fragment's silence or a list's
absence into an affirmative identification is an unsupported inference
and fails the atom.
\end{quote}

\paragraph{Validity review (full prompt).}
\begin{quote}\small
You are an isolated, blind validity reviewer for one public-evidence
agent benchmark question. Another model built the question. You have
not been given its reference analysis, its derivation, the scoring
contract, the independent reproduction, prior reviews, solver
trajectories, or expected difficulty, and you must not try to locate
any of those. Your inputs are exactly what a solver would receive: the
question text, the information cutoff, visible definitions or
assumptions, and the solver's tools, plus public web research through
the tools you are given.

Public research: reopen official primary sources (regulators, issuers,
courts, registries, statutes, agency data portals). Respect the
information cutoff: treat later-dated material as unavailable evidence,
and say so if a load-bearing fact only exists after the cutoff.

\textbf{What to decide.} Determine whether the question, exactly as
written, is professionally meaningful, uniquely answerable,
evidence-closable with the allowed tools by the cutoff, and scorable.
Do not solve the question for its own sake; recover only the
load-bearing relationships needed to test validity. Do not reject a
valid high-level question because it omits intermediate steps, is
difficult, or is retrieval-intensive. Do not accept it merely because
the wording is technical or one number can be reproduced.

Audit at least these questions:
\begin{enumerate}[leftmargin=*,itemsep=1pt,topsep=2pt]
\item Are the entity, facility, unit, jurisdiction, event, period,
interval, time zone, source version, and information cutoff
unambiguous?
\item Are the requested measures compatible in definition, unit,
status, population, and aggregation basis? Is a stock silently treated
as a flow, gross as net, preliminary as final, reported as adjusted,
one label version as another?
\item Does the task distinguish observed association, official
findings, mechanism, and causal or counterfactual claims?
\item Can a competent solver retrieve every load-bearing input using
only the allowed tools and public sources available by the cutoff?
\item Would another primary-source-supported definition, version,
baseline, or procedural path produce a materially different answer
that the wording does not resolve?
\item Does the wording ask for a single high-level analytical result
rather than expose an ordered solution procedure or demand hidden
enumeration?
\item Does any undefined threshold, rounding policy, missing-data
treatment, or source-revision choice prevent a unique scoring target?
\item For the domain-specific dimensions listed in the packet
(\texttt{validity\_dimensions}), return exactly one
\texttt{dimension\_checks} entry per \texttt{dimension\_id}, in packet
order.
\end{enumerate}

Verdicts:
\begin{itemize}[leftmargin=*,itemsep=1pt,topsep=2pt]
\item \texttt{VALID}: professionally meaningful, uniquely answerable,
evidence-closable, and safe to give to the solver exactly as written.
\item \texttt{VALID\_WITH\_REFRAMING}: the underlying analytical object
is valid but the visible wording, cutoff, version, definition, or
conclusion strength must be repaired. Blocks delivery.
\item \texttt{MULTIPLE\_VALID\_INTERPRETATIONS}: materially different
answers remain defensible. Blocks delivery.
\item \texttt{INSUFFICIENT\_PUBLIC\_EVIDENCE}: the allowed public
environment cannot close the task by the cutoff. Blocks delivery.
\item \texttt{DOMAIN\_INVALID}: the requested relationship, formula,
attribution, legal status, or analytical object is invalid even if the
source facts are real. Blocks delivery.
\end{itemize}

\texttt{unique\_answer\_status}: \texttt{UNIQUE} (one correct answer,
which you confirmed), \texttt{UNIQUE\_INDETERMINATE} (the question
admits one correct answer but you could not confirm its value),
\texttt{CONDITIONAL} (unique only under an assumption the question does
not state), \texttt{NOT\_UNIQUE}. Set
\texttt{safe\_to\_enter\_solver=true} only for \texttt{VALID} with
\texttt{unique\_answer\_status} of \texttt{UNIQUE} or
\texttt{UNIQUE\_INDETERMINATE}. For every material objection, cite the
exact public source fact or counterexample and its URL when one exists.
Submit exactly one JSON object matching the output schema.
\end{quote}
}

\section{Evaluation Protocol}
\label{app:judging}

Every answer is graded independently by two judges, GPT-5.6 Sol and
Claude Opus~5, each following the procedure of its domain and ruling
item by item.
When their verdicts on an answer differ, the two sets of item rulings
are passed to a third judge, whose ruling is final.
The item pass rate is computed from the final item rulings.

\paragraph{Task categories.}
The category abbreviations of Tables~\ref{tab:performance} and~\ref{tab:performance-new-full} are as follows.
Finance: Qual./Quant.\ general
qualitative/quantitative analysis, Mkt.\ market analysis, Comp.\
comparables, Prec.\ precedents, Adj.\ adjustments, Earn.\ earnings
analysis, Discl.\ disclosure analysis, Model.\ financial modeling.
Energy: Reg.\ regulation and infrastructure, Mkt.\ markets and
interchange, Gen.\ generation and capacity, Rel.\ reliability events,
Scen.\ scenarios and mitigation, Stor.\ storage, Dem.\ demand and
forecasting. Medicine: Reg.\ regulatory decisions, Trial trial
registration, Discl.\ disclosure and versions, Safety safety and
benefit-risk, Label product labeling.
Law uses the four task types of LAB: analysis, drafting, review, and research.
The extended sets add three energy categories, Discl.\ disclosure,
Grid grid operations, and Emis.\ emissions and efficiency, and one
medicine category, Evid.\ biomedical evidence synthesis.

\paragraph{More recent models.}
\label{app:recent-models}
We also evaluated six more recent models on finance tasks and on
the extended law, energy, and medicine sets
(Table~\ref{tab:performance-new}).
In finance, these models were evaluated on a subset of the tasks.
Grok~4.7 and Claude Opus~5.5 lead in finance, and GPT-6.1 Sol leads in
the other three domains, where the six models keep the same order.
Table~\ref{tab:performance-new-full} gives their results by category.

\begin{table}[!ht]
\caption{Overall pass rates of six more recent models on finance tasks and on the extended law, energy, and medicine sets; Table~\ref{tab:performance-new-full} gives the rates by task category.}
\label{tab:performance-new}
\centering\scriptsize
\setlength{\tabcolsep}{5pt}
\begin{tabular}{@{}lrrrr@{}}
\toprule
Model & Finance & Law & Energy & Medicine \\
\midrule
GPT-6.1 Sol & 54.5 & \textbf{61.7} & \textbf{71.1} & \textbf{78.0} \\
Grok~4.7 & \textbf{66.2} & 46.3 & 57.3 & 57.1 \\
Claude Opus~5.5 & 64.9 & 39.4 & 46.0 & 34.1 \\
DeepSeek V4.1 Flash & 27.3 & 11.8 & 29.9 & 25.9 \\
GLM-5.3 & 28.6 & 10.6 & 29.4 & 17.1 \\
MiMo V2.6 Flash & 28.6 & 9.3 & 23.2 & 14.6 \\
\bottomrule
\end{tabular}
\end{table}

\begin{table}[h]
\caption{Pass rates of six more recent models by task category, on finance tasks and on the extended law, energy, and medicine sets. The highest rate in each column of a block is in bold.}
\label{tab:performance-new-full}
\centering
\scriptsize
\setlength{\tabcolsep}{3.5pt}
\ifdefined\nhcw\else\newlength{\nhcw}\newlength{\nbw}\fi
\setlength{\nhcw}{0.86cm}
\setlength{\nbw}{9.6cm}
\newcommand{\nhc}[1]{\makebox[\nhcw][r]{#1}}
\begin{tabular}{@{}llr@{}}
\toprule
\multicolumn{3}{@{}l}{\textbf{Finance}} \\
Model & \makebox[\nbw]{\nhc{Qual.}\hfill\nhc{Quant.}\hfill\nhc{Mkt.}\hfill\nhc{Comp.}\hfill\nhc{Prec.}\hfill\nhc{Adj.}\hfill\nhc{Earn.}\hfill\nhc{Discl.}\hfill\nhc{Model.}} & Overall \\
\midrule
Grok~4.7 & \makebox[\nbw]{\nhc{\textbf{100.0}}\hfill\nhc{\textbf{100.0}}\hfill\nhc{60.0}\hfill\nhc{28.6}\hfill\nhc{\textbf{77.8}}\hfill\nhc{\textbf{87.5}}\hfill\nhc{\textbf{66.7}}\hfill\nhc{60.0}\hfill\nhc{41.7}} & \textbf{66.2} \\
Claude Opus~5.5 & \makebox[\nbw]{\nhc{\textbf{100.0}}\hfill\nhc{75.0}\hfill\nhc{\textbf{80.0}}\hfill\nhc{28.6}\hfill\nhc{66.7}\hfill\nhc{81.2}\hfill\nhc{58.3}\hfill\nhc{\textbf{70.0}}\hfill\nhc{\textbf{50.0}}} & 64.9 \\
GPT-6.1 Sol & \makebox[\nbw]{\nhc{\textbf{100.0}}\hfill\nhc{\textbf{100.0}}\hfill\nhc{60.0}\hfill\nhc{\textbf{42.9}}\hfill\nhc{66.7}\hfill\nhc{56.2}\hfill\nhc{50.0}\hfill\nhc{50.0}\hfill\nhc{33.3}} & 54.5 \\
GLM-5.3 & \makebox[\nbw]{\nhc{50.0}\hfill\nhc{75.0}\hfill\nhc{20.0}\hfill\nhc{14.3}\hfill\nhc{11.1}\hfill\nhc{25.0}\hfill\nhc{33.3}\hfill\nhc{50.0}\hfill\nhc{16.7}} & 28.6 \\
MiMo V2.6 Flash & \makebox[\nbw]{\nhc{50.0}\hfill\nhc{75.0}\hfill\nhc{20.0}\hfill\nhc{0.0}\hfill\nhc{22.2}\hfill\nhc{50.0}\hfill\nhc{8.3}\hfill\nhc{30.0}\hfill\nhc{25.0}} & 28.6 \\
DeepSeek V4.1 Flash & \makebox[\nbw]{\nhc{50.0}\hfill\nhc{0.0}\hfill\nhc{0.0}\hfill\nhc{14.3}\hfill\nhc{55.6}\hfill\nhc{37.5}\hfill\nhc{25.0}\hfill\nhc{30.0}\hfill\nhc{16.7}} & 27.3 \\
\midrule[\heavyrulewidth]
\multicolumn{3}{@{}l}{\textbf{\lawname{}}} \\
Model & \makebox[\nbw]{\nhc{Analyze}\hfill\nhc{Draft}\hfill\nhc{Review}\hfill\nhc{Research}} & Overall \\
\midrule
GPT-6.1 Sol & \makebox[\nbw]{\nhc{\textbf{55.2}}\hfill\nhc{\textbf{68.8}}\hfill\nhc{\textbf{70.8}}\hfill\nhc{\textbf{56.2}}} & \textbf{61.7} \\
Grok~4.7 & \makebox[\nbw]{\nhc{49.6}\hfill\nhc{35.4}\hfill\nhc{47.0}\hfill\nhc{41.0}} & 46.3 \\
Claude Opus~5.5 & \makebox[\nbw]{\nhc{44.4}\hfill\nhc{41.7}\hfill\nhc{32.4}\hfill\nhc{40.0}} & 39.4 \\
DeepSeek V4.1 Flash & \makebox[\nbw]{\nhc{10.0}\hfill\nhc{29.2}\hfill\nhc{13.7}\hfill\nhc{4.8}} & 11.8 \\
GLM-5.3 & \makebox[\nbw]{\nhc{9.6}\hfill\nhc{2.1}\hfill\nhc{11.9}\hfill\nhc{14.3}} & 10.6 \\
MiMo V2.6 Flash & \makebox[\nbw]{\nhc{10.0}\hfill\nhc{8.3}\hfill\nhc{8.2}\hfill\nhc{10.5}} & 9.3 \\
\midrule[\heavyrulewidth]
\multicolumn{3}{@{}l}{\textbf{\energyfull{}}} \\
Model & \makebox[\nbw]{\nhc{Reg.}\hfill\nhc{Mkt.}\hfill\nhc{Gen.}\hfill\nhc{Rel.}\hfill\nhc{Scen.}\hfill\nhc{Stor.}\hfill\nhc{Dem.}\hfill\nhc{Discl.}\hfill\nhc{Grid}\hfill\nhc{Emis.}} & Overall \\
\midrule
GPT-6.1 Sol & \makebox[\nbw]{\nhc{\textbf{74.7}}\hfill\nhc{66.7}\hfill\nhc{\textbf{63.2}}\hfill\nhc{\textbf{77.8}}\hfill\nhc{70.0}\hfill\nhc{\textbf{70.0}}\hfill\nhc{\textbf{75.0}}\hfill\nhc{\textbf{66.7}}\hfill\nhc{\textbf{62.5}}\hfill\nhc{\textbf{50.0}}} & \textbf{71.1} \\
Grok~4.7 & \makebox[\nbw]{\nhc{66.7}\hfill\nhc{\textbf{86.7}}\hfill\nhc{31.6}\hfill\nhc{59.3}\hfill\nhc{\textbf{80.0}}\hfill\nhc{30.0}\hfill\nhc{37.5}\hfill\nhc{42.9}\hfill\nhc{37.5}\hfill\nhc{33.3}} & 57.3 \\
Claude Opus~5.5 & \makebox[\nbw]{\nhc{48.3}\hfill\nhc{53.3}\hfill\nhc{\textbf{63.2}}\hfill\nhc{55.6}\hfill\nhc{60.0}\hfill\nhc{10.0}\hfill\nhc{37.5}\hfill\nhc{28.6}\hfill\nhc{25.0}\hfill\nhc{33.3}} & 46.0 \\
DeepSeek V4.1 Flash & \makebox[\nbw]{\nhc{34.5}\hfill\nhc{40.0}\hfill\nhc{21.1}\hfill\nhc{22.2}\hfill\nhc{60.0}\hfill\nhc{10.0}\hfill\nhc{25.0}\hfill\nhc{28.6}\hfill\nhc{0.0}\hfill\nhc{33.3}} & 29.9 \\
GLM-5.3 & \makebox[\nbw]{\nhc{41.4}\hfill\nhc{33.3}\hfill\nhc{31.6}\hfill\nhc{29.6}\hfill\nhc{10.0}\hfill\nhc{10.0}\hfill\nhc{25.0}\hfill\nhc{14.3}\hfill\nhc{0.0}\hfill\nhc{0.0}} & 29.4 \\
MiMo V2.6 Flash & \makebox[\nbw]{\nhc{21.8}\hfill\nhc{33.3}\hfill\nhc{26.3}\hfill\nhc{37.0}\hfill\nhc{60.0}\hfill\nhc{10.0}\hfill\nhc{0.0}\hfill\nhc{4.8}\hfill\nhc{0.0}\hfill\nhc{33.3}} & 23.2 \\
\midrule[\heavyrulewidth]
\multicolumn{3}{@{}l}{\textbf{\medfull{}}} \\
Model & \makebox[\nbw]{\nhc{Reg.}\hfill\nhc{Trial}\hfill\nhc{Discl.}\hfill\nhc{Safety}\hfill\nhc{Label}\hfill\nhc{Evid.}} & Overall \\
\midrule
GPT-6.1 Sol & \makebox[\nbw]{\nhc{\textbf{72.9}}\hfill\nhc{\textbf{68.4}}\hfill\nhc{\textbf{87.0}}\hfill\nhc{\textbf{84.2}}\hfill\nhc{\textbf{92.3}}\hfill\nhc{\textbf{77.8}}} & \textbf{78.0} \\
Grok~4.7 & \makebox[\nbw]{\nhc{52.9}\hfill\nhc{63.2}\hfill\nhc{73.9}\hfill\nhc{57.9}\hfill\nhc{46.2}\hfill\nhc{55.6}} & 57.1 \\
Claude Opus~5.5 & \makebox[\nbw]{\nhc{38.8}\hfill\nhc{31.6}\hfill\nhc{52.2}\hfill\nhc{26.3}\hfill\nhc{23.1}\hfill\nhc{22.2}} & 34.1 \\
DeepSeek V4.1 Flash & \makebox[\nbw]{\nhc{28.2}\hfill\nhc{57.9}\hfill\nhc{47.8}\hfill\nhc{10.5}\hfill\nhc{7.7}\hfill\nhc{7.4}} & 25.9 \\
GLM-5.3 & \makebox[\nbw]{\nhc{27.1}\hfill\nhc{5.3}\hfill\nhc{17.4}\hfill\nhc{5.3}\hfill\nhc{23.1}\hfill\nhc{7.4}} & 17.1 \\
MiMo V2.6 Flash & \makebox[\nbw]{\nhc{20.0}\hfill\nhc{31.6}\hfill\nhc{4.3}\hfill\nhc{2.6}\hfill\nhc{7.7}\hfill\nhc{14.8}} & 14.6 \\
\bottomrule
\end{tabular}
\end{table}

\paragraph{Finance.}
The judge works in two steps.
First, before seeing the scoring contract, it reads the whole answer,
checks the solver's retrieval record, and notes preliminary findings.
Second, it rules on every item and every fatal condition of the scoring
contract, quoting the answer and giving the evidence and the reason for
each ruling.
It then runs six checks on the whole answer---professional concepts,
scope and period, facts and inference, internal consistency,
computation and units, and completion of the task requirements---and
resolves each of its preliminary findings.
To fail an item or a check, the judge must identify the statement in
the answer that fails it.
An answer passes only if every item passes, no fatal condition is
triggered, and all six checks pass; an item the judge cannot decide
does not count as passed.

\paragraph{Law.}
The judge has no tools.
It reads the files or text that the answer delivers and rules on the
items of the scoring contract one at a time, following the
per-criterion procedure of LAB.
An answer passes only if every item passes, and an answer that delivers
nothing fails.

\paragraph{Agreement with human grading.}
We compared the judge with human grading.
One reviewer with a finance background and one reviewer with a legal
background independently graded 60 answers, 30 in finance and 30 in
law, sampled from model answers to our finance and law tasks, half of
which the judge had passed and half failed.
The reviewers received exactly what the judge receives: the question,
the scoring contract, the reference analysis, the evidence, and the
answer.
They did not see any ruling of the judge.
Table~\ref{tab:judge-human} compares the judge's final rulings with the
reviewers'.
On scored items, the judge and the reviewers agree on 96.7\% of the
rulings.
Agreement on whole answers is lower because an answer passes only if
every item passes, so a single item on which the two differ changes the
verdict.
The disagreements go in both directions.

\begin{table}[h]
\caption{The judge's final rulings and the human reviewers' rulings on
60 answers. The third and fourth columns give, of the items the judge
passed, how many the reviewer also passed, and of the items the judge
failed, how many the reviewer also failed.}
\label{tab:judge-human}
\centering\footnotesize
\begin{tabular}{@{}lrcccc@{}}
\toprule
Level & Items & \shortstack{Judge: pass $\to$\\reviewer: pass} & \shortstack{Judge: fail $\to$\\reviewer: fail} & \shortstack{Overall\\agreement} & Cohen's $\kappa$ \\
\midrule
Scored items & 632 & 573 / 586 & 38 / 46 & 96.7\% & 0.77 \\
Whole answer & 60  & 24 / 30   & 25 / 30 & 81.7\% & 0.63 \\
\bottomrule
\end{tabular}
\end{table}

\subsection{Solving Environments}
\label{app:harness}

\paragraph{Finance.}
The environment follows the tool interface of the Finance Agent
Benchmark: web search, EDGAR search, a reader for web pages and
documents, a calculator, price history, and FFIEC bank call reports.
Every source tool returns only material published before the task's
information cutoff, and the solver delivers its answer through a
submission tool.
A task allows at most 120 tool calls and one hour.

\paragraph{Law.}
The environment follows the sandbox of LAB: the task's materials are
local, the solver works with seven file tools in a container without
network access, and it delivers files or text.
A task allows at most 200 turns and one hour, with at most ten minutes
per model request and two minutes per shell command.

Table~\ref{tab:runs} summarizes the finance runs of the six models of Table~\ref{tab:performance}.

\begin{table}[h]
\caption{Finance solving runs of the six models of Table~\ref{tab:performance}: turns per run (median
and interquartile range), wall-clock minutes per run (median), answer
length in characters (median), and mean turns of passed and failed
answers.}
\label{tab:runs}
\centering\footnotesize
\begin{tabular}{@{}lcrrc@{}}
\toprule
Model & Turns & Minutes & Answer length & Turns, pass / fail \\
\midrule
Grok~4.6          & 30 (20--43) & 4.7 & 8{,}752 & 30.4 / 39.5 \\
DeepSeek V4 Flash & 25 (17--32) & 5.1 & 7{,}614 & 24.2 / 28.1 \\
Claude Opus~5     & 20 (13--27) & 2.5 & 9{,}927 & 18.7 / 23.8 \\
Qwen3.8-27B       & 22 (14--31) & 9.2 & 7{,}807 & 21.0 / 25.4 \\
GLM-5.2           & 16 (11--24) & 2.9 & 7{,}674 & 17.0 / 20.6 \\
MiniMax-M3        & 22 (15--34) & 2.5 & 7{,}570 & 21.1 / 30.6 \\
\bottomrule
\end{tabular}
\end{table}

\section{Failure Analysis Details}
\label{app:failures}

\subsection{How Failures Are Classified}
\label{app:failures-method}

Besides checking each item of the scoring contract, the judge runs six
checks over the whole answer, and an answer fails if any check fails:
\begin{itemize}[leftmargin=*,itemsep=1pt,topsep=2pt]
\item \textbf{Fact and inference}: a statement contradicts the sources,
or a conclusion exceeds the evidence;
\item \textbf{Internal consistency}: numbers, directions, or statements
in the answer contradict one another;
\item \textbf{Task requirements}: an output the task asks for is missing
or incomplete;
\item \textbf{Scope or period}: the wrong population, period, statistical
scope, or basis of comparison is used;
\item \textbf{Computation or units}: an operation, unit, sign, or order
of magnitude is wrong;
\item \textbf{Professional concept}: an accounting, financial, or
regulatory concept is misused.
\end{itemize}
To fail a check, the judge must identify the statement in the answer,
the evidence or task requirement it conflicts with, and the error that
results.
An answer can trigger several checks, so the shares below sum to more
than 100\%.
All shares are of failed answers, taken over the failed answers of the
six models on the finance tasks of Table~\ref{tab:performance}.

\subsection{By Model}
\label{app:failures-models}

\begin{table}[h]
\caption{Share of each model's failed answers that trigger each whole-answer
check (\%). An answer can trigger several checks.}
\label{tab:fail-models}
\centering
\footnotesize
\setlength{\tabcolsep}{5pt}
\begin{tabular}{@{}lrrrrrr@{}}
\toprule
Model & \shortstack{Fact and\\inference} & \shortstack{Internal\\consistency} & \shortstack{Task\\requirements}
 & \shortstack{Scope or\\period} & \shortstack{Computation\\or units} & \shortstack{Professional\\concept} \\
\midrule
Grok~4.6          & 75.0 & 47.3 & 47.8 & 25.0 & 28.3 & 19.0 \\
DeepSeek V4 Flash & 85.0 & 60.7 & 54.5 & 34.6 & 28.0 & 26.2 \\
Claude Opus~5     & 83.8 & 60.4 & 47.6 & 34.5 & 29.3 & 21.1 \\
Qwen3.8-27B       & 88.3 & 67.6 & 55.3 & 36.2 & 33.8 & 32.7 \\
GLM-5.2           & 86.3 & 68.6 & 65.1 & 40.9 & 35.4 & 30.7 \\
MiniMax-M3        & 90.1 & 72.7 & 65.0 & 45.6 & 36.8 & 39.4 \\
\midrule
All               & 85.9 & 64.8 & 57.2 & 37.6 & 32.6 & 29.6 \\
\bottomrule
\end{tabular}
\end{table}

For every model, errors of fact and inference are the most common.
Weaker models tend to trigger more checks per failed answer: 2.4 on average for
Grok~4.6 and 3.5 for MiniMax-M3 (Table~\ref{tab:fail-models}).

\subsection{By Task Type}
\label{app:failures-types}

\begin{table}[h]
\caption{Share of the failed answers in each task type that trigger each
whole-answer check (\%), over the six models of Table~\ref{tab:performance}.}
\label{tab:fail-types}
\centering
\footnotesize
\setlength{\tabcolsep}{5pt}
\begin{tabular}{@{}lrrrrrr@{}}
\toprule
Task type & \shortstack{Fact and\\inference} & \shortstack{Internal\\consistency} & \shortstack{Task\\requirements}
 & \shortstack{Scope or\\period} & \shortstack{Computation\\or units} & \shortstack{Professional\\concept} \\
\midrule
Qualitative analysis  & 79.9 & 52.4 & 67.1 & 48.2 & 22.0 & 30.5 \\
Quantitative analysis & 83.0 & 69.4 & 55.2 & 31.9 & 33.8 & 33.4 \\
Market analysis       & 92.6 & 57.0 & 44.4 & 35.6 & 25.2 & 24.4 \\
Comparables           & 87.4 & 61.1 & 54.7 & 37.9 & 33.7 & 36.8 \\
Precedents            & 85.3 & 66.7 & 69.3 & 53.3 & 40.0 & 24.0 \\
Adjustments           & 89.1 & 73.1 & 48.7 & 27.6 & 36.5 & 24.4 \\
Earnings analysis     & 84.2 & 63.0 & 57.1 & 31.9 & 36.2 & 25.1 \\
Disclosure analysis   & 90.1 & 64.1 & 59.3 & 44.2 & 27.4 & 31.0 \\
Financial modeling    & 81.7 & 70.5 & 58.6 & 35.6 & 41.7 & 32.4 \\
\bottomrule
\end{tabular}
\end{table}

Errors of fact and inference lead in every task type
(Table~\ref{tab:fail-types}).
Scope or period errors peak in precedents (53.3\%) and
qualitative analysis (48.2\%), errors of computation or units in
financial modeling (41.7\%), and unmet task requirements in precedents
(69.3\%) and qualitative analysis (67.1\%).

\subsection{Cases}
\label{app:failures-cases}

Each of the following six cases corresponds to one check.
In each, most of the answer is right, and the error lies at a point that
changes the conclusion.

\paragraph{Fact and inference: a recorded adjustment called final (EOG
Resources, MiniMax-M3).}
The model correctly rebuilt the rollforward of asset retirement
obligations: a net increase of \$37 million in 2026 and \$24 million in
2025, a difference of \$13 million; it also correctly placed the \$15
million from Encino within the \$19 million of estimate revisions.
It then wrote that the adjustment ``was finalized as a
measurement-period adjustment.''
Note~12 of the filing still describes the purchase price allocation as
preliminary, and an adjustment that has been recorded is not thereby
final.

\paragraph{Internal consistency: a correct table with the opposite
explanation (Boeing, DeepSeek V4 Flash).}
The model's table gives GAAP operating earnings of \$604 million and
core operating earnings of \$294 million for the first half of 2026,
both correct.
Explaining the FAS/CAS adjustment, it wrote that ``excluding it raises
operating earnings'' and gave $604 + 310 = 294$; the issuer's
reconciliation is $604 - 310 = 294$.
The table and the explanation cannot both hold.

\paragraph{Task requirements: eligibility conditions without product
matching (CMS Part~D, GLM-5.2).}
The task asks when an immediate substitution qualifies.
The model introduced five conditions with ``only if all of the
following are satisfied'' but omitted the requirement that the
substitute match the original product in dosage form, route of
administration, and strength, which the CMS document states on page~47.
Without it, the set of acceptable substitutes changes.

\paragraph{Scope or period: a half year in place of the quarter the
task names (EOG Resources, Grok~4.6).}
The task asks to compare actual production in the second quarter of
2026 with guidance for the third quarter and the full year.
The model's data include the quarter, the half year, and the guidance,
but for the full-year comparison it used the half year.
For natural gas liquids, the quarter was 346.8 thousand barrels per
day, the half year 339.5, and the midpoint of full-year guidance 342:
the quarter lies 4.8 above the midpoint and the half year 2.5 below it,
so the change of basis reverses the direction.

\paragraph{Computation or units: basis points inflated a hundredfold
(Visa, GLM-5.2).}
The model correctly took net revenue of \$40,000 and \$35,926 million
for fiscal 2025 and 2024 and total payments volume of \$13,894 and
\$12,988 billion.
With units aligned, net revenue is 28.79 basis points of payments
volume, against 27.66 a year earlier, a change of $+1.13$ basis points.
The answer gives levels of about 2,878.94 and 2,766.09 basis points and
a change of $+1{,}128.49$: the levels are about 100 times too large and
the change about 1,000 times.

\paragraph{Professional concept: receivables called a contract asset
(Best Buy, Claude Opus~5).}
The model correctly summed four contract liabilities to \$1,333 million
and correctly noted that the \$287 million of restricted cash is not
broken down for the membership program.
It then called the \$528 million of receivables, net ``a contract
asset, not a liability.''
An unconditional right to payment is presented as a receivable, whereas
a contract asset is a right that still depends on something other than
the passage of time; the error changes the class of the asset.

\subsection{Borderline Verdicts}
\label{app:failures-borderline}

Under the strict standard, some failed answers contain errors of little
consequence.
For NVIDIA, $22{,}563 / 26{,}044 = 86.634\%$, which rounds to 86.63\%,
but Grok~4.6 wrote 86.64\% in its additional analysis.
For Curtiss-Wright, the task asks for one decimal place, and MiniMax-M3
wrote 108\% instead of 108.0\%.
For Boeing, DeepSeek V4 Flash stated that every page carries the
preliminary-data header, which the cover page does not.
We keep these verdicts under the strict standard.
To gauge how common such cases are, we screened for failed answers that
pass every item, fail exactly one whole-answer check, and give a
reason involving rounding, decimal places, pages, or formatting; they
make up 2.8\% of failed answers.

\section{Case Studies}
\label{app:maincases}
\label{app:cases}

Three cases show what the tasks demand (source data, scoring
contracts, and model outcomes below).
\emph{Travelers}: how much of a 6.7-point combined-ratio
improvement is underlying? Removing catastrophe losses and reserve
releases leaves 0.6 points, about 9\%, and the solver must build this
bridge in both ratio points and pre-tax dollars without treating the
two retained shares as interchangeable.
\emph{Northrop Grumman}: why did operating income fall
23\% on 5\% sales growth? Four segment changes, an intersegment
elimination, a pension adjustment, and an absent divestiture gain must
be layered without treating any one layer as exhaustive---and the
strongest model overall fails this task while a weaker one passes.
\emph{SouthState}: does a 33\% rise in the loan-loss
allowance mean worse loans? Loans grew 39\%, mostly through
acquisitions, and coverage fell; the correct judgment is that the
increase is non-diagnostic.

For each case, we describe the analyst's question, what the source
determines, how the scoring contract encodes it, and how models
performed.

\subsection{Travelers: separating underlying underwriting improvement}
\label{case:travelers}

The task asks how much of Travelers' consolidated underwriting
improvement in Q2~2026 versus Q2~2025 reflects improved
current-accident-year performance rather than lower catastrophe
losses and favorable prior-year reserve development (PYD).
The task tests whether the solver keeps two measures apart: it must
construct two independent bridges---one in combined-ratio points,
one in pre-tax dollars---and compute the retained share in each, without treating the two metrics as interchangeable or merging their bases.

The source determines that the reported combined ratio improved
6.7~percentage points, from 90.3\% to 83.6\%.
Removing catastrophe impact (8.5 to 4.9~pp) and favorable PYD
($-$2.9 to $-$5.4~pp) yields underlying combined ratios of 84.7\%
and 84.1\%, an improvement of only 0.6~pp---approximately 9\% of
the reported change.
A parallel pre-tax bridge shows reported underwriting income rising
\$716M, of which underlying income contributes only \$44M
(approximately 6.1\%).
Both bridges lead to the same qualitative conclusion: the underlying
direction improved, but the retained share was small; the reported
improvement was predominantly catastrophe-and-PYD-driven.

\begin{table}[h]
\caption{Travelers Q2 improvement decomposition.}
\label{tab:travelers}
\centering
\footnotesize
\begin{tabular}{@{}lccc|c@{}}
\toprule
\textbf{Metric} & \textbf{Catastrophe} & \textbf{PYD} & \textbf{Underlying} & \textbf{Reported} \\
\midrule
Combined ratio (pp) & 3.6 & 2.5 & 0.6 & 6.7 \\
Pre-tax (\$M)       & 409 & 263 &  44 & 716 \\
\midrule
Retained share      & --- & --- & $\sim$9\% / $\sim$6\% & 100\% \\
\bottomrule
\end{tabular}
\end{table}

The scoring contract has nine answer atoms, covering both quarterly ratio bridges, the ratio and pre-tax change
decompositions, both retained shares, and an earnings-quality
judgment.
Its eight fatal conditions guard against scope errors (using a
segment-level or year-to-date population), measure conflation
(treating the 9\% ratio share and 6.1\% dollar share as the same
number), and unsupported inference (asserting reserve adequacy or
future persistence from one quarter's result).
Five of the six models pass; GLM-5.2 fails.

\subsection{Northrop Grumman: reconciling segment drivers to consolidated income}
\label{case:northrop}

The task asks why Northrop Grumman's Q2~2026 consolidated operating
income declined 23\% (\$329M) despite 5\% sales growth.
The task tests whether the solver can reconcile layers of different
scope: it must correctly aggregate four segment changes, account for intersegment
eliminations, layer in FAS/CAS pension adjustments and an absent
prior-year divestiture gain, and reconcile the resulting
\$348M subtotal with the actual \$329M decline through a favorable
corporate residual---without treating any single layer as exhaustive.

\begin{table}[h]
\caption{Northrop Grumman Q2 operating-income bridge (\$M, YoY change).}
\label{tab:northrop}
\centering
\footnotesize
\begin{tabular}{@{}lr@{}}
\toprule
Component & $\Delta$ (\$M) \\
\midrule
Defense Systems    & $-$97 \\
Space Systems      & $-$44 \\
Mission Systems    & $+$60 \\
Aeronautics Systems & $+$41 \\
\cmidrule(lr){2-2}
Four-segment subtotal & $-$40 \\
Intersegment eliminations & $-$21 \\
\cmidrule(lr){2-2}
Total segment OI change & $-$61 \\
FAS/CAS adjustment      & $-$56 \\
Prior-year divestiture gain (absent) & $-$231 \\
\cmidrule(lr){2-2}
Named-driver subtotal & $-$348 \\
Other corporate (favorable) & $+$19 \\
\midrule
Consolidated OI change & $-$329 \\
\bottomrule
\end{tabular}
\end{table}

The scoring contract has 14 answer atoms, covering consolidated
operating income and margins in both quarters, the dollar and
percentage decline, aggregate segment operating income, the segment
bridge with the Defense and Space declines, the lower FAS/CAS
adjustment, the absent \$231M divestiture gain, and an explanation of
the decline despite sales growth.
Its seven fatal conditions include basing the conclusion on another
issuer or period; omitting or misstating the absent gain, the FAS/CAS
change, or the segment totals; and claiming that the named program EAC
amounts exhaust the segment changes.
Claude Opus~5, Qwen3.8-27B, and GLM-5.2 pass; Grok~4.6,
DeepSeek V4 Flash, and MiniMax-M3 fail.
Grok~4.6, the strongest model overall, fails where GLM-5.2 passes.

\subsection{SouthState: population normalization before judgment}
\label{case:southstate}

The task asks whether the nominal 33\% increase in SouthState's
consolidated allowance for credit losses (ACL) from December~2024 to
June~2025 is representative of higher portfolio loss content, or
whether it is non-diagnostic after controlling for gross-loan growth
and loan-population mix.
The task tests whether the solver draws only the conclusion the
evidence supports: it must build a quantitative bridge before judging, not treat
the nominal increase as self-evidently meaningful.

\begin{table}[h]
\caption{SouthState loan populations and ACL coverage.}
\label{tab:southstate}
\centering
\footnotesize
\begin{tabular}{@{}lrrr@{}}
\toprule
Population & Dec 2024 (\$K) & Jun 2025 (\$K) & Growth contrib. \\
\midrule
Acquired PCD       &    862{,}155 &  3{,}409{,}186 & 19.06\% \\
Acquired non-PCD   &  3{,}635{,}782 & 12{,}492{,}553 & 66.27\% \\
Non-acquired       & 29{,}404{,}990 & 31{,}365{,}508 & 14.67\% \\
\midrule
Total gross loans  & 33{,}902{,}927 & 47{,}267{,}247 & 100\% \\
\midrule
ACL coverage       & 1.3724\% & 1.3139\% & --- \\
\bottomrule
\end{tabular}
\end{table}

Total gross loans grew 39.42\%, driven almost entirely by acquired
portfolios (85.33\% of growth).
The ACL increased 33.48\% in nominal terms, but ACL-to-loan coverage
\emph{decreased} from 1.3724\% to 1.3139\% ($-$5.85~basis points).
The scoring contract classifies the correct answer as
``non-diagnostic'': the nominal ACL increase does not by itself
establish higher portfolio loss content because loans grew faster
than the allowance and the acquired-loan mix changed materially.

The contract has eight answer atoms, requiring the solver to
present population balances at both dates, reconcile them to total
gross loans, compute population shares and changes, build a
gross-loan growth bridge, and state the coverage movement before
rendering a judgment.
Its seven fatal conditions guard against omitting required quantitative
support, calling the increase ``representative,'' and drawing
unsupported inferences (treating arithmetic growth contributions as
acquisition causality or asserting reserve adequacy).
None of the six models passes. All reach the non-diagnostic
conclusion, but each adds an error: three attribute the change in the
allowance to the acquisitions beyond what the sources support, and the
others miscalculate the allowance needed to hold coverage flat,
misstate the allowance recorded for acquired loans, or miss the
required precision.

The five cases below give each objective, the trap that makes it hard, and how the six models fared.

\subsection{Marvell: foundry commitments without double counting
(Disclosure Analysis)}
A fabless chip company depends on foundry capacity, and an analyst
needs its future purchase obligations by year. The commitment table
totals \$2{,}185.5M, but the notes state that at least
\$482.5M of capacity-reservation purchases are already included
in it, while \$23.1M of fees and deposits and \$204.7M of
capital expenditure are disclosed separately without the same
mapping. The task is a set-membership problem: which amounts are
inside the table, which are outside, and which cannot be placed. None
of the six models passes it.

\subsection{CrowdStrike and Snowflake: can two retention rates be
ranked? (Comparables)}
Software analysts routinely place net retention rates side by side.
CrowdStrike measures retention on annualized contract value for a
fixed cohort of subscription customers; Snowflake measures it on
consumption-based product revenue over a two-year customer window
that includes reseller end-customers. The correct answer is that the
two figures cannot be ranked without a common basis---the task tests
judgment about whether a comparison is meaningful, not arithmetic. The
error is reading the metric's name rather than its definition. Only
DeepSeek V4 Flash passes it.

\subsection{EOG Resources: aligning actuals with guidance (Earnings
Analysis)}
An analyst updating a quarterly model must put actual production,
realized prices, unit costs, and capital spending on the same basis as
forward guidance. Production must be restricted to the geographies the
guidance covers; absolute realized prices cannot be subtracted from
differentials quoted relative to WTI; and some rows should be computed
while others must be left as not comparable. No model
passes it.

\subsection{Informatica: rebuilding a precedent-transaction set
(Precedents)}
An M\&A analyst reconstructs the disclosed precedent set behind a
fairness opinion: 28 dated transactions, with the summary rows at
the bottom of the table excluded, the minimum, maximum, and median
EV/NTM revenue multiples recomputed, and the offer premium tied to the
correct reference date. Precedent multiples and share-price premiums
are different reference points and must not be merged. Grok~4.6
and DeepSeek V4 Flash pass it.

\subsection{Advance Auto Parts: event window and comparator
contamination (Market Analysis)}
After an earnings release, the stock rises. The analyst must first
place the release relative to the trading session (the filing was
accepted before the open, so the prior close is the baseline), then
compare the move against peers, a sector index, and the market---while
noticing that one peer had its own event inside the comparison window.
The error is comparing against a contaminated reference set. Only
Qwen3.8-27B passes it.

\subsection{Discussion}
\label{app:case-discussion}

Performance on these cases does not follow the overall ranking of
Table~\ref{tab:performance}.
Grok~4.6, the strongest model overall, fails the Northrop Grumman task
that GLM-5.2 passes, and only Qwen3.8-27B passes the Advance Auto Parts task.
The failed answers are classified by check in
Appendix~\ref{app:failures}.

\section{Training Details}
\label{app:training}

\paragraph{Training data.}
Training data are selected from successful solving trajectories on the
training tasks.
A trajectory keeps the reasoning, the tool calls, and the tool results;
only the assistant's reasoning and actions are supervised.

\paragraph{Fine-tuning.}
Qwen3.5-9B~\citep{qwen35} is fine-tuned with all parameters of its
language model at a learning rate of $1\times10^{-5}$, on full
trajectories.
Qwen3.5-27B is fine-tuned with LoRA (rank 32, alpha 64) on all linear
layers of its language model at a learning rate of $5\times10^{-5}$;
trajectories longer than 48K tokens are truncated at a complete
assistant turn.
Both use a cosine schedule with a short warmup and BF16 precision, and
are trained for four epochs and then continued from the epoch-4 weights
to ten epochs.

\paragraph{Evaluation.}
Models before and after fine-tuning are evaluated in the same solving
environment and under the same budget (Appendix~\ref{app:harness}),
with thinking enabled, greedy decoding, at most 32{,}768 output tokens
per response, and a context of 262{,}144 tokens.

\paragraph{Leakage filter for external benchmarks.}
We do not have access to the full evaluation sets of the external
benchmarks, so we evaluate on the questions each has released
publicly.
Public questions and their answers can leak through web search, so
search results pass two filters during solving: a blacklist of domains,
and a language model that checks every search result and every opened
page and blocks copies of the questions, reference answers, and
rubrics.

\paragraph{Law, energy, and medicine.}
Qwen3.5-9B is fine-tuned with the configuration above for four epochs
in each domain.

\paragraph{LAB.}
The \lawname{} solving trajectories we train on answer in text, so we rewrote the output
instruction of each LAB task to ask for a text answer, and we count an
answer delivered either as text or as files.
After removing the tasks whose materials far exceed the context window
of Qwen3.5-9B, we sampled 243 tasks at random.
Answers are graded with the per-criterion procedure of LAB, and a task
passes only if every criterion passes.

\paragraph{LifeSci Diamond.}
We evaluate on the 10 public tasks, solved in our medicine
environment and scored with the benchmark's own scorer.

\section{Ablations and Production Findings}
\label{app:ablation-findings}
\subsection{Stage Replacement Study}
\label{app:replacement}

\paragraph{Setup.}
We took the complete production trajectory of one finance task in the
Utilities industry group: built on the resource adequacy
report of the Southwest Power Pool, it asks how the region's accredited
capacity mix changes from 2025 to 2030.
The trajectory has 104 steps, 82 carried out by models and 22 by code.
One model step was affected by a fault in the execution environment and
is left out; each of the other 81 is one replacement condition.
A condition starts from a snapshot of the environment just before the
step: the files and records the step could see and, for a step of the
orchestrator, its conversation so far.
Qwen3.8-27B redoes the step with the same instructions and tools, and
later steps keep the outputs of the original run, so the conditions do
not affect one another.
For four orchestrator steps whose conversation exceeded Qwen's context,
the history was written to files instead.
Each step allowed up to 512 model requests and 24 hours, with no limit
on tokens.

\paragraph{Review.}
Each of Qwen's outputs was checked against the step's instructions, its
sources, and what the next step needs.
The outputs of the original run had already been used by later steps
and were judged by whether a problem surfaced later in the run.
An output is usable as is, needs fixes (the main work can be kept), or
is unusable (a key decision is wrong or no complete output is
produced); a step whose job was to check counts an error it let pass as
its own.

\paragraph{Results by stage.}
Table~\ref{tab:replacement-groups} splits the results by stage.
The gap is widest in the stages of first version, difficulty, and review
and of repairs and re-review, where most steps decide whether to let a
task pass and to which step a defect returns; on problem family,
compilation, and novelty the two runs are on par.
Table~\ref{tab:replacement-full} describes what each kind of error
usually looks like.

\paragraph{Examples.}
\emph{Ranking requirement.}
The question asked for the changes by category to be ranked, while the
scoring contract accepted only a ranking by absolute change in MW;
ranking by percentage gives a different order.
In the original run, several steps from compilation to formal review
missed this, until, at the final step, the orchestrator adopted an
alternative reading raised by the validity review and repaired the
question.
At the same step, Qwen received the same counterexample and still judged
the question sound because the scoring rule could be satisfied.
\emph{Date conflict.}
The cover of the report is dated June~15, while the document's metadata
shows that it became public on June~16.
Qwen explained the conflict as an upload delay and let the task pass
without evidence for that explanation.
\emph{Unfinished check.}
Qwen read about 20{,}000 of the document's roughly 25{,}000 characters,
yet reported that the text matched the source exactly and, on that
basis, ruled out a revision after the cutoff.

\paragraph{Usage and execution.}
On the 67 steps for which token usage was logged in both runs, Qwen used
about 3.8 times the output tokens and 1.8 times the input tokens of the
original run.
In 76 of the 81 conditions, Qwen's run ended without error, which does
not mean that its output could be used.
Of the three conditions without a complete output, two were cut off by
the length limit of a single response and one failed its structured
submission.

\begin{table}[h]
\caption{Stage replacement results by stage: the number of steps whose
output was usable as is, needed fixes, or was unusable, for the original
run and for Qwen3.8-27B.}
\label{tab:replacement-groups}
\centering\footnotesize
\setlength{\tabcolsep}{4pt}
\begin{tabular}{@{}lr@{\hspace{8pt}}ccc@{\hspace{8pt}}ccc@{}}
\toprule
 & & \multicolumn{3}{c}{Original run} & \multicolumn{3}{c}{Qwen3.8-27B} \\
\cmidrule(lr){3-5}\cmidrule(l){6-8}
Stage & Steps & Usable & Fixes & Unusable & Usable & Fixes & Unusable \\
\midrule
Sources, candidates, early screening & 15 & 11 & 3 & 1 & 6 & 6 & 3 \\
Integration, evidence, rebuild inputs & 9 & 5 & 4 & 0 & 3 & 3 & 3 \\
Independent rebuild and comparison & 7 & 5 & 1 & 1 & 4 & 1 & 2 \\
Problem family, compilation, novelty & 7 & 5 & 0 & 2 & 5 & 1 & 1 \\
First version, difficulty, review & 13 & 9 & 2 & 2 & 3 & 6 & 4 \\
Repairs and re-review & 10 & 7 & 0 & 3 & 1 & 4 & 5 \\
Wording check and formal review & 8 & 3 & 3 & 2 & 1 & 4 & 3 \\
Ranking repair and final review & 12 & 12 & 0 & 0 & 4 & 8 & 0 \\
\midrule
All & 81 & 57 & 13 & 11 & 27 & 33 & 21 \\
\bottomrule
\end{tabular}
\end{table}

\begin{table}[h]
\caption{Outputs of the original run and of Qwen3.8-27B at the 81 model
steps of one finance task. The original run used GPT-5.6 Sol for most
steps and Claude Opus~5 for several reviews. The lower block counts the
steps with each kind of error.}
\label{tab:replacement-full}
\centering
\footnotesize
\setlength{\tabcolsep}{4pt}
\begin{tabular}{@{}>{\raggedright\arraybackslash}p{3.5cm}rr>{\raggedright\arraybackslash}p{6.0cm}@{}}
\toprule
 & Original run & Qwen3.8-27B & \\
\midrule
Usable as is         & \textbf{57} (70\%) & \textbf{27} (33\%) & \\
Needs fixes          & 13 (16\%) & 33 (41\%) & The main work can be kept \\
Unusable             & 11 (14\%) & 21 (26\%) & A key decision is wrong, or no complete output is produced \\
\midrule
Kind of error & & & What these errors usually look like \\
\midrule
Unstated requirement & 14 & 14 &
Checks that each scoring item has matching words in the question, not
whether a solver reading only the question would know to answer that
way; leaves a boundary or definition incomplete \\
\addlinespace
Wrong decision & 3 & 11 &
Notices a conflict or counterexample but explains it away with an
earlier conclusion, a program flag, or a scoring rule the solver cannot
see, and lets the task pass; or returns a defect to too early a step \\
\addlinespace
Unverified claim & 2 & 8 &
Reports an unfinished or unprovable check as confirmed: claims a full
match after reading part of a document, takes the date of an event for
the date of publication, or states a guessed cause as fact \\
\addlinespace
Incomplete handoff & 4 & 7 &
Passes on materials that are missing or misdescribed, or reports a step
as done when only part of it is \\
\addlinespace
Wrong format & 1 & 6 &
Breaks the agreed structure of the output, e.g., submits an object as a
string \\
\addlinespace
Factual slip & 0 & 5 &
Computes correctly but attaches a value to the wrong item, or
contradicts itself \\
\addlinespace
No output & 0 & 3 &
The output is cut off for length or the structured submission fails,
leaving no result \\
\bottomrule
\end{tabular}
\end{table}

\subsection{Findings from Production}
\label{app:findings}

\paragraph{The answer comes before the question.}
In 20 question-first attempts, a question was
written first, and an agent then searched the sources for its answer.
Fifteen reached an answer that appeared to pass verification, but five
of them first needed the comparison basis, a threshold, or the bounds of
the conclusion corrected, against one of the same 20 when the question
is written last.
In one banking task on a community program, the question passed
independent reproduction, yet its hidden scoring rejected an answer
based on the California amount, which a primary source that the
question allows explicitly supports: the reproducer had worked only within the
evidence selected for the question and could not see that the task admitted
broader evidence and another valid answer.

\paragraph{Length is not difficulty.}
Table~\ref{tab:items} groups the finance tasks by the number of items in
their scoring contracts.
Tasks with fewer than five items are easier; beyond that, more items
leave the pass rate between 33\% and 40\%.
Two rounds of experiments extended tasks that the constructing model had
already solved, adding reasoning layers that each consume the result of
the previous one.
In the first round, 16 of 29 designs passed a structural review and
three yielded complete new relationships, adding two, one, and one
layers, and a follow-up extension took one more task from four to six
layers.
In the second, larger round, 11 of 59 designs passed the structural
review, and two tasks passed answer verification, going from five to
seven and from five to eight layers.
Across 12 runs on these six extended tasks, the constructing model's
core conclusion was always correct; for one task, the solving time rose
from 188 to 297 seconds.
The difficulty screen had predicted success rates of 17\% to 78\% for
them.

\begin{table}[h]
\caption{Pass rate of the six models, pooled, by the number of items in
the scoring contract, with each group's share of the finance tasks.}
\label{tab:items}
\centering\footnotesize
\begin{tabular}{@{}lrr@{}}
\toprule
Contract items & Share of tasks & Pass rate \\
\midrule
1--4        & 8.6\%  & 48.8\% \\
5--6        & 22.2\% & 40.3\% \\
7--8        & 18.1\% & 40.2\% \\
9--11       & 20.0\% & 33.4\% \\
12--15      & 12.7\% & 39.5\% \\
16--20      & 10.5\% & 32.9\% \\
21 or more  & 7.9\%  & 33.0\% \\
\bottomrule
\end{tabular}
\end{table}

\paragraph{Where models do fail.}
From tasks the constructing model had answered wrongly, we derived five
kinds of judgment that change a conclusion: the scope of a risk stock
against a loss flow, a qualitative boundary the issuer leaves undefined,
double recognition across ordered accounting adjustments, the version
and revision scope of a disclosure, and exact thresholds applied to
rounded inputs.
New tasks built around these five kinds were the first on which the
constructing model erred in its core conclusion (four tasks, eight
runs).

\paragraph{Other observations.}
Adding material makes a task harder only when it changes a verifiable
relationship in the answer; more documents or background alone do not.
The difficulty screen orders candidates well but underestimates how
often models succeed.
A validity review confirms that a task stands, not that it is hard.

\subsection{Direct-Generation Baseline}
\label{app:direct-generation}

{  %
\paragraph{Setup.}
As a simple baseline, the same model writes a task directly from the
same sources; Autodata~\citep{autodata2026} uses prompt-based generation
of this kind as its control.
We took 50 delivered finance tasks, three or four from each of 13
industry groups, and gave the source files bound to each task,
unchanged, to the constructing model (GPT-5.6 Sol).
In a single session and with the prompt below, the model writes a
question, a reference answer with its answer atoms, and the scoring
points; there are no evidence fragments, no independent reproduction,
and no review.
\begin{quote}\small
You are a senior financial analyst asked to write ONE expert-level
benchmark task directly from the bound primary sources, in a single
pass. There is no pipeline: you alone write the question, the reference
answer, and the scoring points.

Read \texttt{packet.json} for the item id and the source inventory. Use
\texttt{read\_source} and \texttt{source\_search} on the bound sources
(neutral source IDs only). Then produce, via \texttt{submit\_report},
an \texttt{analysis} string that is a single JSON object with exactly
these fields:
\begin{itemize}[leftmargin=*,itemsep=1pt,topsep=2pt]
\item \texttt{question\_text}: the task a professional analyst would be
given (one paragraph; state the company/event, the analytical
objective, the information cutoff date, and the deliverables; no hints
about the solution path).
\item \texttt{information\_cutoff}: ISO date (YYYY-MM-DD) consistent
with the sources.
\item \texttt{reference\_answer}: a core conclusion and 4 to 12 atoms,
each with an id, a label (what is being determined), the expected
answer with value, unit, period and basis, and one sentence on how it
follows from the sources.
\item \texttt{scoring\_points}: list of strings, each a condition a
correct answer must satisfy (fatal ones prefixed ``FATAL:'').
\item \texttt{sources\_used}: list of neutral source IDs actually
relied on.
\end{itemize}
Requirements: the question must be answerable only from these sources;
the reference answer must be exactly supported by them (quote-level
precision on figures, periods, units, scope); do not invent facts.
\end{quote}

\paragraph{Checks.}
The 50 directly generated tasks and the 50 delivered tasks then went
through the same two checks: a review of the reference answer against
the sources, in the manner of the consistency review, and an
independent reproduction from the analytical objective, a description
of each required output that does not reveal its value, and the
sources, followed by a comparison atom by atom.

\paragraph{Results.}
The scored content of the reference answer is defective in 16 of the 50
directly generated tasks and in 4 of the 50 delivered tasks (Fisher's
exact test, two-sided, $p=0.005$; Table~\ref{tab:direct-generation}).
Of the 16, the review found 12, and 4 were found only by the
reproduction and comparison.
The types were assigned by three language-model annotators who worked
independently and did not know which group a task came from, by
majority vote (pairwise Cohen's $\kappa$ 0.90--0.95).

The defects take two forms.
In 10 directly generated tasks, the reference answer itself is wrong:
it omits a material item stated in the sources, uses the wrong basis,
period, or population, contains a numeric error, or misdescribes the
sources.
In 7, the question and the reference answer do not match: the question
requires an output that the reference answer does not give, or the
reference answer relies on documents published after the information
cutoff of the question.
One task shows both.
Among the delivered tasks, the two forms occur in 3 tasks and 1 task.
}

\begin{table}[h]
\caption{Tasks with a defect in the scored content of the reference
answer, by type of defect, out of 50 per group. One directly generated
task has two types.}
\label{tab:direct-generation}
\centering\footnotesize
\begin{tabular}{@{}lcc@{}}
\toprule
Type of defect & Direct generation & \pipename{} \\
\midrule
Relies on documents published after the information cutoff & 4 & 0 \\
An output required by the question is not given            & 3 & 1 \\
A material item stated in the sources is omitted           & 4 & 1 \\
Wrong basis, period, or population                         & 3 & 1 \\
Numeric error                                              & 2 & 1 \\
A statement misdescribes the sources                       & 1 & 0 \\
\midrule
Tasks with at least one defect                             & 16 & 4 \\
\bottomrule
\end{tabular}
\end{table}

\subsection{A Simple Inspector in Place of Independent Reproduction}
\label{app:direct-review}

{  %
\paragraph{Setup.}
To measure what independent reproduction contributes, we replaced it
with a simple inspector at the same position and asked whether the defects it
had found in production would still be found.
In independent reproduction, a second worker derives the answer from
the evidence without seeing the reference analysis; the inspector is
given the reference analysis together with the evidence and asked to
find what is wrong.
From the production records we took 133 defects in reference analyses
that the comparison with the independent reproduction had found and
that were then repaired, 56 in finance and 77 in law, and gave the
versions before repair to the inspector.
A first inspector receives the reference analysis and the evidence and
looks for defects; a second inspector sees the report of the first and
inspects again.
Both use the constructing model (GPT-5.6 Sol) and the budget of the
reproducer; the only difference is that there is no reproduction.
Language-model annotators then judged each case against the repair that
was made in production: found, if either inspector identified the
defect; partly found, if an inspector raised a concern about the right
item but the problem it described was not the actual defect; or not
found.

\paragraph{Results.}
The inspector found 57\% of the defects and missed 32\% entirely
(Table~\ref{tab:direct-review}); 29 of the 42 missed defects were
repaired by changing the core conclusion.
The misses are concentrated in one kind of defect: the reference
analysis is internally consistent, so nothing looks wrong when it is
read on its own, but the same evidence also supports another reasonable
answer that the scoring contract does not accept (16 of 28 such defects
were missed).
Conclusions that go beyond the evidence were also often missed (7 of
12), whereas missing evidence, numeric errors, and errors of scope were
mostly found.
An inspector reads the reference analysis and checks whether each step
holds, so an analysis that is internally consistent passes; the
reproducer never sees the reference analysis, arrives at the other
answer when the evidence supports one, and the comparison then shows
the difference.
The study covers only defects that independent reproduction found, and
the inspector stands at the position of the reproduction, not at the
later reviews.
}

\begin{table}[h]
\caption{Defects in reference analyses that independent
reproduction found in production, and how many of them an inspector at the
same position found.}
\label{tab:direct-review}
\centering\footnotesize
\begin{tabular}{@{}lrrrr@{}}
\toprule
 & Defects & Found & Partly found & Not found \\
\midrule
Finance & 56  & 35 & 3  & 18 \\
Law     & 77  & 41 & 12 & 24 \\
\midrule
Total   & 133 & 76 (57\%) & 15 (11\%) & 42 (32\%) \\
\bottomrule
\end{tabular}
\end{table}

\subsection{What Each Step Contributes}
\label{app:step-contribution}

{  %
Table~\ref{tab:step-contribution} goes through the steps of the
pipeline in the order in which they run (Appendix~\ref{app:method}), so
a problem found at one step had passed every step before it.
The ``Ablation'' column describes how removing or replacing a step
affects the final task quality, and the ``Result'' column gives the
numbers we measured, from the experiments of this appendix or from the
production records.

\paragraph{Code checks.}
Code matches every quotation in a worker's output against the sources.
About 6\% of the worker outputs in finance and 9\% in law contained a
quotation that is not in the sources verbatim; such an output is
rejected and the step is redone.

\paragraph{Validity review and consistency review.}
Table~\ref{tab:review-returns} covers only the delivered tasks and
gives the share that each review returned at least once and the average
number of returns per task.
A returned task goes back with the problems the review found, each
confirmed by the adjudicator; these are repaired and the task is
reviewed again, at most five times (Appendix~\ref{app:method}).
The consistency review returns the most because it is the first check
that sees both the question and the scoring contract, and a mismatch
between the two is the most common cause of repair
(\S\ref{sec:results-prod}).
After all of these steps, the expert review found a substantive problem
in 6.5\% of the finance tasks and in 1 of 50 law tasks
(\S\ref{sec:human-review}).

\paragraph{Wording polish.}
The polish makes sure the question tells the solver everything the
scoring expects and nothing more.
In finance, 32\% of the polish reviews edited the wording of the
question.
Table~\ref{tab:polish-edits} gives the share of these edits that made
each kind of change; one edit can make several.
}

\begin{table}[h]
\caption{The steps of the pipeline in the order in which they run:
how removing or replacing each step affects the final task quality,
and the numbers we measured.
Shares of candidates are from the finance run
(Figure~\ref{fig:funnel}).}
\label{tab:step-contribution}
\centering\scriptsize
\setlength{\tabcolsep}{3pt}
\renewcommand{\arraystretch}{1.1}
\newcolumntype{L}[1]{>{\raggedright\arraybackslash}p{#1}}
\begin{tabular}{@{}L{2.0cm}L{3.5cm}L{3.5cm}L{4.1cm}@{}}
\toprule
Step & What it does & Ablation & Result \\
\midrule
Discovery routes & Propose candidate analyses from the sources & Not a check; nothing to remove & The ten routes are listed in Table~\ref{tab:routes} \\
Novelty check & Removes candidates that duplicate an earlier one & Without it, the candidates it removes stay & 10.5\% of the candidates entering it \\
Integration & Ties every relationship of a candidate to a verbatim quotation; a candidate the sources do not support exits & Without it, the candidates it removes stay & 31.3\% of the candidates entering it \\
Difficulty screen & Ranks candidates without solving them; the easiest exit & Without it, the candidates it removes stay. It affects difficulty, not correctness & 13.9\% of the candidates entering it (including those whose reference analysis could not be completed) \\
Building the reference analysis & Derives every output from the quotations & Without it there is no answer. The last row replaces it with direct writing & See the last row \\
Code checks & Match every quotation in a worker's output against the sources and recompute every step & Without them, an output with a quotation that is not in the sources goes on to the next step & About 6\% of the worker outputs in finance and 9\% in law are rejected for such a quotation and redone \\
Independent reproduction & A second worker derives the answer without seeing the reference analysis, and the two are compared & Without it, the defects it exposes stay in the reference analysis. We also replaced it with a simple inspector (Appendix~\ref{app:direct-review}) & It exposed a defect in the reference analysis of 5.9\% of the finance candidates and 16.2\% of the law candidates (\S\ref{sec:results-prod}). The inspector did not correctly identify 43\% of such defects (Table~\ref{tab:direct-review}) \\
Question written last & The question is written from the verified reference analysis & We wrote the question first (Appendix~\ref{app:findings}) & 5 of 20 questions needed their requirements corrected, against 1 of 20 when the question is written last \\
Validity review & Judges the question from what the solver sees & Without it, a returned task is delivered as it was before the return & 21.2\% of the delivered finance tasks and 5.6\% of the delivered law tasks were returned by it \\
Consistency review & Checks the question, the reference analysis, and the scoring contract against one another & Without it, a returned task is delivered as it was before the return & 44.5\% of the delivered finance tasks and 24.0\% of the delivered law tasks were returned by it \\
Wording polish & Edits only the wording of the question & Without it, the question is delivered with its original wording & 32\% of the polish reviews in finance edited the wording \\
Whole pipeline & All of the above & The model writes a task directly from the same sources (Appendix~\ref{app:direct-generation}) & Defective reference answer in 16 of 50 tasks, against 4 of 50 for \pipename{} \\
\bottomrule
\end{tabular}
\end{table}

\begin{table}[h]
\caption{Delivered tasks that each review returned at least once,
and the average number of returns per delivered task.}
\label{tab:review-returns}
\centering\footnotesize
\begin{tabular}{@{}lrrrr@{}}
\toprule
 & \multicolumn{2}{c}{Finance} & \multicolumn{2}{c}{Law} \\
\cmidrule(lr){2-3}\cmidrule(l){4-5}
 & Returned & Returns per task & Returned & Returns per task \\
\midrule
Validity review    & 21.2\% & 0.24 & 5.6\%  & 0.06 \\
Consistency review & 44.5\% & 0.60 & 24.0\% & 0.32 \\
Either review      & 56.1\% & 0.84 & 27.2\% & 0.37 \\
\bottomrule
\end{tabular}
\end{table}

\begin{table}[h]
\caption{What the wording polish changed in the finance questions
it edited. One edit can make several kinds of change.}
\label{tab:polish-edits}
\centering\footnotesize
\begin{tabular}{@{}lr@{}}
\toprule
What the edit does & Share of the edits \\
\midrule
States the form the answer should take & 34\% \\
Makes explicit a requirement that the question only implied & 34\% \\
Removes a hint to the answer & 29\% \\
Replaces internal vocabulary & 22\% \\
Names a source less specifically & 8\% \\
Removes a request that is not scored & 7\% \\
\bottomrule
\end{tabular}
\end{table}

\section{Extended Related Work}
\label{app:related}

\paragraph{Data Engine with an execution oracle.}
SWE-Gym~\citep{swegym2024} collects real GitHub issues with executable
environments and trains smaller models on teacher trajectories that
pass the tests.
SWE-smith~\citep{swesmith2025} removes the dependence on human-written
issues by synthesizing bugs into existing repositories through
LM-based modification, procedural edits, and PR mirroring, yielding
tens of thousands of instances; R2E-Gym~\citep{r2egym2025} likewise
builds executable environments procedurally and adds execution-free
verifiers for test-time scaling.
SWE-Next~\citep{swenext2026} mines merged pull requests at scale and
keeps only commit pairs whose tests strictly improve without
regressions, so that each task verifies itself.
CLI-Universe~\citep{cliuniverse2026} synthesizes terminal tasks from
technical materials with fail-to-pass tests and hint-assisted teacher
execution.
TaskCraft~\citep{taskcraft2026} composes tool-answerable atomic
operations through depth and width expansion and validates by
rejection sampling; APIGen-MT~\citep{apigenmt2025} generates
multi-turn tool-use data through blueprint-then-execution, with
blueprints validated by a committee of LM reviewers.
In all of these the reference is fixed by executing something; the
training recipe they share, filtered-trajectory
fine-tuning~\citep{agenttuning2023}, is the one we adopt, with the
scoring contract as the filter.

\paragraph{Data Engine without an execution oracle.}
When the deliverable cannot be executed, existing methods obtain
verifiability by restricting it.
FinForge~\citep{finforge2026} pairs an expert-curated financial corpus
with LM-generated multiple-choice questions.
AGORA~\citep{agora2026} builds cross-document questions over workplace
document collections in eight domains, with independent solving,
leakage checking, and answer verification, but the deliverable is a
single value.
SPARK~\citep{spark2026} and WebSailor-V2~\citep{websailor2026} derive
questions from knowledge graphs---over scientific literature and over
web-expanded entities respectively---so that the graph fixes the
answer; SPARK checks answers against graph paths and trains proposer
and solver by self-play, while WebSailor-V2 scores its answers with a
language-model judge.
QUEST~\citep{quest2026} keeps a long-form deliverable and grades it
against rubric trees, but the rubrics are produced by the same
generation process that writes the task.
SWE-QA-Pro~\citep{sweqapro2026} likewise asks for free-form answers,
to repository-level questions seeded from real issues; it drops
questions a direct-answer baseline can solve and trains with
fine-tuning followed by reinforcement learning from AI feedback.
\pipename{} shares the information-asymmetry principle of SPARK and
the source grounding of AGORA, but its construction unit is a complete
professional analysis, and its reference is reproduced independently from the
sources by a second worker rather than checked against a graph or trusted from the generator that wrote it.

\paragraph{Expert-authored professional benchmarks.}
GDPval~\citep{gdpval2025} compares model deliverables with those of
occupational experts in blinded pairwise judgments, and
APEX~\citep{apex2026} grades long-horizon professional tasks against
expert-audited rubrics.
In finance, Finance Agent Benchmark~\citep{fabv1_2025,fabv2_2026}
defines nine task categories and an agentic harness over SEC filings;
BigFinanceBench~\citep{bigfinancebench2026} and
FrontierFinance~\citep{frontierfinance2026} grade open-ended
financial-research tasks with point-weighted and source-attributed
rubrics respectively; Hedge-Bench~\citep{hedgebench2026} derives tasks from the
recorded reasoning of analysts.
In law, LAB~\citep{harvey_lab2026} assigns agents a client matter and
grades the work product against expert-written criteria.
These benchmarks fix the shape of the task and of the rubric that
\pipename{} reproduces without expert authoring; we adopt their
solving environments and categories (\S\ref{sec:setup},
\S\ref{sec:task-composition}) and use their public questions as
transfer targets (\S\ref{sec:results-learn}).

\paragraph{Learning from constructed tasks.}
We fine-tune on solver trajectories that pass the scoring contract,
the rejection-sampling recipe of STaR~\citep{star2022} applied to
agents by AgentTuning~\citep{agenttuning2023}, SWE-Gym~\citep{swegym2024},
and APIGen-MT~\citep{apigenmt2025}, with the contract replacing
execution as the filter.
The neighboring construction methods go further and train by
reinforcement learning on their own tasks: TaskCraft and QUEST on
rubric or answer rewards, WebSailor-V2 in a simulated web
environment, and SPARK by proposer--solver self-play.
The item-level contract of \pipename{} is a verifiable reward of the
same kind, and we leave reinforcement learning on it to future work.

\paragraph{Task form.}
Table~\ref{tab:related} compares how methods construct and verify
their reference answers; Table~\ref{tab:task-form} compares the form of
the tasks themselves: their sources, how they are solved, the form of
the answer, and how it is scored.
Among the task-synthesis methods, only \pipename{} produces tasks that
require multi-turn tool use over professional primary documents and are
answered with a long analysis scored against a rubric, which is the
task form of the expert-written benchmarks in the lower part of the
table.

\begin{table}[h]
\caption{Form of the tasks produced by task-synthesis methods (top)
and of expert-written benchmarks (bottom).}
\label{tab:task-form}
\centering\scriptsize
\setlength{\tabcolsep}{3pt}
\renewcommand{\arraystretch}{1.08}
\newcolumntype{L}[1]{>{\raggedright\arraybackslash}p{#1}}
\begin{tabular}{@{}L{3.3cm}L{1.9cm}L{2.5cm}L{1.75cm}L{1.4cm}L{1.9cm}@{}}
\toprule
 & Domain & Sources & Solving & Answer & Scoring \\
\midrule
\multicolumn{6}{@{}l}{\emph{Task-synthesis methods}} \\
FM2DS~\citep{fm2ds2025} & general & Wikipedia & single-turn QA & short answer & reference-answer matching \\
FinForge~\citep{finforge2026} & finance & financial text & single-turn QA & multiple choice & correct-option matching \\
Autodata~\citep{autodata2026}, legal setting & law & professional primary documents & single-turn QA & long answer & rubric \\
WebSailor-V2~\citep{websailor2026} & general & open web & multi-turn tool use & short answer & reference-answer matching \\
WebShaper~\citep{webshaper2025} & general & open web & multi-turn tool use & short answer & reference-answer matching \\
TaskCraft~\citep{taskcraft2026} & general & open web, financial-report PDFs & multi-turn tool use & short answer & reference-answer matching \\
AGORA~\citep{agora2026} & eight industries, including finance, healthcare, and law & professional primary documents & multi-turn tool use & short answer & reference-answer matching \\
QUEST~\citep{quest2026} & general & open web & multi-turn tool use & short or long answer & rubric \\
\pipename{} (ours) & finance, law, energy, medicine & professional primary documents & multi-turn tool use & long answer & rubric \\
\midrule
\multicolumn{6}{@{}l}{\emph{Expert-written benchmarks}} \\
LAB~\citep{harvey_lab2026} & law & professional primary documents & multi-turn tool use & long answer & rubric \\
FrontierFinance~\citep{frontierfinance2026} & finance & professional primary documents, open web & multi-turn tool use & long answer & rubric \\
BigFinanceBench~\citep{bigfinancebench2026} & finance & professional primary documents & multi-turn tool use & short answer & rubric \\
FAB v1 / v2~\citep{fabv1_2025,fabv2_2026} & finance & professional primary documents, open web & multi-turn tool use & long answer & rubric \\
\bottomrule
\end{tabular}
\end{table}

\section{Human Review Details}
\label{app:human-review}

{  %
\paragraph{Finance.}
Table~\ref{tab:review-agreement} gives the number of tasks in which
each of the two reviewers of \S\ref{sec:human-review} found a
substantive problem, by criterion, and their agreement.
Of the \nreviewed{} tasks, 15 were flagged by at least one reviewer.
One of the reviewers re-examined these 15 tasks one by one against the
sources: 12 were confirmed to have a substantive problem and the other
3 were not, which gives the 93.5\% of \S\ref{sec:human-review}
(174 of \nreviewed{}).

The problems in the 12 confirmed tasks are of three kinds.
In 9 tasks, the scoring contract asks for more than the question
requires: an additional explanation or statement (5 tasks), a
derived value or subtotal (3 tasks), or a specific date (1 task).
In 2 tasks, the information cutoff set during construction is too
early, so that evidence the task depends on was published after it.
In 1 task, the reference analysis contains a misstatement or an
omission.
}

\begin{table}[h]
\caption{Tasks in which each reviewer found a substantive problem,
out of \nreviewed{}, and the agreement between the two reviewers.}
\label{tab:review-agreement}
\centering\footnotesize
\begin{tabular}{@{}lrrrrr@{}}
\toprule
Criterion & Reviewer 1 & Reviewer 2 & Both & Agreement & Cohen's $\kappa$ \\
\midrule
Task realism                  & 0  & 0 & 0 & 100\%  & -- \\
Source fidelity               & 0  & 0 & 0 & 100\%  & -- \\
Correctness of the conclusion & 5  & 2 & 2 & 98.4\% & 0.56 \\
Scoring fairness              & 9  & 3 & 2 & 95.7\% & 0.32 \\
\midrule
Whole task                    & 14 & 5 & 4 & 94.1\% & 0.40 \\
\bottomrule
\end{tabular}
\end{table}

{  %
\paragraph{Law.}
One expert with a legal background reviewed 50 law tasks under the same
four criteria and found a substantive problem in 1 task, on the
correctness of the conclusion and on scoring fairness: the scoring
contract requires the answer to state a specific deadline that has
nothing to do with the regulatory judgment the question asks for.

\paragraph{Review by models.}
We used two models, Claude Fable~5.2 and GPT-6 Astra, to review 50
tasks from each domain under the same four criteria.
Like the human reviewers, each model receives the question, the
reference analysis, the scoring contract, and the source documents, and
can reopen the official sources and recompute.
Table~\ref{tab:model-review} gives the number of tasks in which each
model flagged a substantive problem.
The last row gives how many of the flagged tasks were confirmed by
human review.
In finance and law, the tasks are among those reviewed by the experts.
In energy and medicine, authors with a computer-science background
re-reviewed the flagged tasks against the sources, assisted by a
language model, and confirmed 2 in energy and none in medicine.

\paragraph{Models and experts.}
Table~\ref{tab:model-expert} compares the two models with the experts
on the 50 finance tasks.
Both models found all 5 problems the experts found and flagged other
tasks that were not confirmed.
These false alarms are of three kinds: overlooking a scope that the
question has already fixed (3 tasks for Fable~5.2 and 2 for Astra);
treating equivalent wording, or an ordinary requirement on precision or
format, as unfair (6 and 4); and treating lenient scoring as a problem
(0 and 1).
The models can therefore serve as a first screen whose flags an expert
then checks, not as a replacement for expert review.
}

\begin{table}[h]
\caption{Tasks in which each model flagged a substantive problem, out
of 50 per domain (Fable~5.2 / Astra).}
\label{tab:model-review}
\centering\footnotesize
\begin{tabular}{@{}lcccc@{}}
\toprule
Criterion & Finance & Law & Energy & Medicine \\
\midrule
Task realism                  & 0 / 0   & 0 / 0 & 0 / 1 & 0 / 0 \\
Source fidelity               & 1 / 1   & 0 / 1 & 0 / 0 & 0 / 0 \\
Correctness of the conclusion & 0 / 0   & 0 / 1 & 0 / 1 & 1 / 1 \\
Scoring fairness              & 14 / 12 & 1 / 4 & 2 / 2 & 2 / 2 \\
\midrule
Flagged by the model          & 14 / 12 & 1 / 4 & 2 / 4 & 3 / 3 \\
Confirmed by human review     & 5       & 1     & 2     & 0     \\
\bottomrule
\end{tabular}
\end{table}

\begin{table}[h]
\caption{The two models against the experts on 50 finance tasks.}
\label{tab:model-expert}
\centering\footnotesize
\begin{tabular}{@{}lccccc@{}}
\toprule
Model & \shortstack{Found by experts\\/ flagged} & False alarms & Misses & \shortstack{Agreement with\\the experts} & Cohen's $\kappa$ \\
\midrule
Fable~5.2 & 5 / 14 & 9 & 0 & 82.0\% & 0.44 \\
Astra     & 5 / 12 & 7 & 0 & 86.0\% & 0.52 \\
\bottomrule
\end{tabular}
\end{table}

\end{document}